%% file: main_rebuttal.tex
\documentclass{article} 
\usepackage{iclr/iclr2026_conference,times}

\usepackage[dvipsnames]{xcolor}
\usepackage[utf8]{inputenc} 
\usepackage[T1]{fontenc}    
\usepackage{microtype}      
\usepackage{url}            
\usepackage{graphicx}
\usepackage{subcaption}
\usepackage{booktabs}       
\usepackage{multirow}
\usepackage{makecell}

\usepackage{floatrow}
\usepackage{algorithm}
\usepackage{algorithmic}
\usepackage{hyperref}       
\usepackage{pifont}
\usepackage{enumitem}
\usepackage{nicefrac}       
\usepackage{caption}
\usepackage{subcaption}
\usepackage{tablefootnote}
\usepackage{soul}
\usepackage{wrapfig}
\usepackage{tikz}
\usetikzlibrary{fit}
\usetikzlibrary{arrows.meta, positioning}

\definecolor{color1}{HTML}{D81B60}
\definecolor{color2}{HTML}{1E88E5}
\definecolor{color4}{HTML}{FFC107}
\definecolor{color3}{HTML}{004D40}
\usepackage{lib/researchpack}
\usepackage{lib/macros}

\hypersetup{
   colorlinks=true,
   linkcolor=MidnightBlue,
   citecolor=MidnightBlue,
   urlcolor=MidnightBlue
}

\usepackage[capitalize, nameinlink]{cleveref}
\usepackage{thmtools, thm-restate}
\declaretheorem[style=plain]{definition, theorem, proposition, lemma}

\theoremstyle{plain}

\newtheorem{example}{Example}

\newtheorem{assumption}{Assumption}

\crefname{algorithm}{Algorithm}{Alg.}

\newif\ifshowcomments
\showcommentsfalse

\title{GNN Explanations that do not Explain\\ and How to find Them}

\author{
  Steve Azzolin \footnotemark[2]
  \And
  Stefano Teso\footnotemark[2]
  \And
  Bruno Lepri\footnotemark[3]
  \And
  Andrea Passerini\footnotemark[2]
  \And
  Sagar Malhotra\footnotemark[4]
}

\iclrfinalcopy
\begin{document}

\maketitle

\renewcommand{\thefootnote}{\fnsymbol{footnote}}
\footnotetext[0]{Corresponding author: \texttt{steve.azzolin@unitn.it}}

\footnotetext[0]{$^{\dagger}$ University of Trento, Italy \quad
$^{\ddagger}$ Fondazione Bruno Kessler, Italy \quad
$^{\S}$ TU Wien, Austria}

\renewcommand{\thefootnote}{\arabic{footnote}}
\setcounter{footnote}{0}

\begin{abstract}
Explanations provided by Self-explainable Graph Neural Networks (SE-GNNs) are fundamental for understanding the model's inner workings and for identifying potential misuse of sensitive attributes.
Although recent works have highlighted that these explanations can be suboptimal and potentially misleading, a characterization of their failure cases is unavailable.
In this work, we identify a critical failure of SE-GNN explanations: explanations can be unambiguously unrelated to how the SE-GNNs infer labels.
We show that, on the one hand, many SE-GNNs can achieve optimal true risk while producing these degenerate explanations, and on the other, most faithfulness metrics can fail to identify these failure modes.
Our empirical analysis reveals that degenerate explanations can be maliciously planted (allowing an attacker to hide the use of sensitive attributes) and can also emerge naturally, highlighting the need for reliable auditing.
To address this, we introduce a novel faithfulness metric that reliably marks degenerate explanations as unfaithful, in both malicious and natural settings.
\end{abstract}

\section{Introduction}

Self-Explainable GNNs (\SEGNNs) combine the predictive power of standard GNNs with ante-hoc explainability \citep{miao2022interpretable, chen2024howinterpretable}. They couple an explanation extractor, which identifies explanatory subgraphs, with a classifier that uses these subgraphs to generate predictions.
Being designed with explainability in mind, \SEGNNs promise to be genuinely interpretable and therefore suitable for high-risk use cases, \eg power grid analysis \citep{varbella2024powergraph}, health forecast \citep{hu2024self}, and drug discovery \citep{wong2024discovery}.

Despite their promise, previous work has highlighted weaknesses in \SEGNNs' explanations.
For instance, they can be redundant \citep{tai2025redundancy}, ambiguous \citep{azzolin2025formal}, may not fully reflect the individual importance of each of their sub-components \citep{chen2024howinterpretable}, and can be affected by spurious correlations \citep{wu2022discovering}.
Nonetheless, little research has focused on whether there exist cases in which \SEGNN{s} catastrophically fail to provide meaningful explanations.

\begin{figure}[t]
    \centering
    \includegraphics[width=0.9\linewidth]{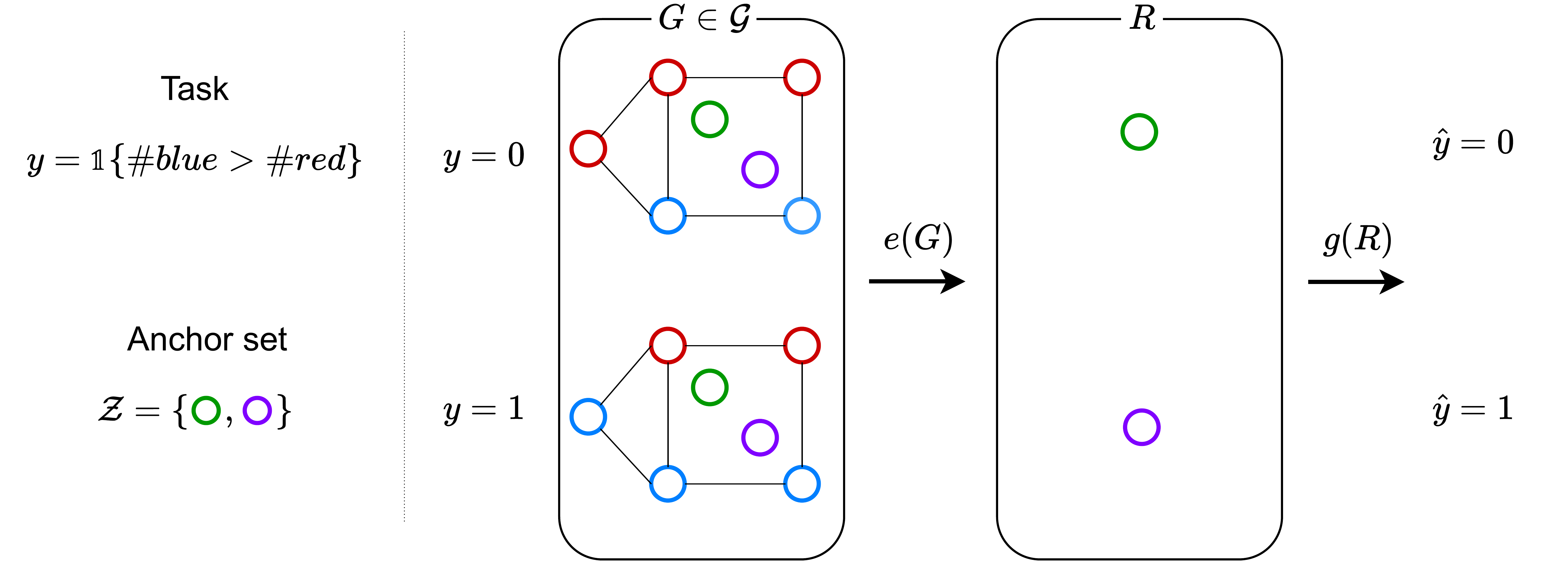}
    \caption{
        \SEGNNs couple an explanation extractor \DET producing an explanation $R$, and a classifier \CLF using $R$ to infer the prediction.
        \textbf{We identify a critical failure case of \SEGNNs, i.e., degenerate explanations -- explanations that encode the label but are unrelated to how the model actually infers it.}        
        In this example, green (\green) and violet (\violet) nodes appear identically in every graph and form an anchor set $\calZ$, hence they have no class-discriminative power by construction (see \cref{sec:failure-cases} and \cref{ex:red-blue} for details).
        Yet the explanation extractor $\DET$ can exploit them to secretly encode the predicted label while hiding the use of red and blue nodes.
        Such explanations are highly unfaithful and mislead users by falsely suggesting that green and violet nodes are relevant for the label.
    }
    \label{fig:example}
\end{figure}

In this work, we identify a critical failure case of \SEGNN's explanations: \textit{explanations can be unrelated to the \SEGNN's inner workings} (see \cref{fig:example} for an example).
Such explanations are unambiguously unfaithful, undermining the original purpose of \SEGNNs, raising the risk of explanations concealing the use of protected attributes, and impairing model debugging and scientific discovery.
First, we theoretically show that, under mild assumptions, several \SEGNNs can output these unfaithful explanations while achieving optimal loss.
We then exploit this result to answer the following questions positively:
 \begin{itemize}[leftmargin=1.25em]
    \item \textbf{RQ1:} Can \SEGNN be manipulated to output maliciously defined unfaithful explanations?

    \item \textbf{RQ2:} Can these unfaithful explanations go undetected by faithfulness metrics?

    \item \textbf{RQ3:} Can these unfaithful explanations also emerge without manipulation?
\end{itemize}
%
To address the undetectability of these unfaithful explanations, we present a \textit{benchmark} for faithfulness metrics and a novel \textit{evaluation metric}, \SUFFCAUSE, that is shown to be more robust.
More generally, our work warns practitioners from blindly trusting \SEGNN explanations, highlights avenues to make these models more reliable, and provides users with more reliable tools to audit their explanations.


\section{Background}
\label{sec:background}

\textbf{Graph Classification}. 
We consider the standard supervised graph classification setting, where the goal is to learn a deterministic graph classifier $\MONO$ mapping from the set of graphs $\calG$ to the set of labels $\calY$, typically implemented as a GNN \citep{bacciu2020gentle}.
A graph $G=(V,E)$ is defined as a set $V$ of nodes connected by edges $E$, and both can have features associated with them.
We write $R \subseteq G$ to denote that $R$ is a subgraph of $G$, and $|G|$ to denote the size of the graph in terms of nodes, edges, or both, as will be clear by the context.
Subgraphs may preserve all, some, or none of the features.
%

\textbf{Self-explainable GNNs.}  
\SEGNNs aim to enhance the interpretability of graph classifiers by generating explanations during inference \citep{miao2022interpretable, miao2022interpretablerandom, chen2024howinterpretable}.
An \SEGNN consists of two components:
(1) an \textbf{explanation extractor} $\DET$ that maps the input graph $G$ to a subgraph $\DET(G) = R \subseteq G$, and
(2) a \textbf{classifier} $\CLF$ that takes $R$ as input and produces the final prediction:
\[
    \MONO(G) = \CLF(\DET(G)).
\]

The explanation extractor \DET can either produce per-edge relevance scores $p_{uv}=\DET(G)_{(u,v)} \in [0,1]$, or per-node relevance scores $p_{u}=\DET(G)_{u} \in [0,1]$, which are thresholded or top-$k$ selected to form the explanatory subgraph $R$ \citep{wu2022discovering, tai2025redundancy}.
We will focus on per-node relevance scores, but our analysis also applies to per-edge scores.
Also, we say the explanation extractor is \textit{hard} if it outputs scores in $\{r,1\}$, where $r$ is either $0$ or an \SEGNN-specific hyperparameter; the precise value will be clear from the context.
We will assume $|R| > 0$ to prove our theoretical result.

\input{tables/taxonomy}

\textbf{Faithfulness metrics.}
At a high level, faithfulness metrics assess whether the predictor's output changes upon perturbing the input \citep{azzolin2025reconsidering}.
Metrics that estimate an explanation's \textit{sufficiency} perturb the complement of the explanation subgraph while keeping the rest unchanged, while metrics that evaluate its \textit{necessity} do the opposite.
%
An explanation is considered highly faithful if perturbing the complement does not induce a large shift, and if perturbing the explanation does.
Different metrics also differ based on the set of perturbations $\calI$ they allow:
\textit{i \& ii)} \textbf{Complement removal} and \textbf{Explanation removal} metrics erase the complement and the explanation altogether, respectively;
\textit{iii)} \textbf{Edge removal} metrics erase edges at random;
\textit{iv)} \textbf{Complement swap} metrics replace the complement of an explanation with that of another graph.
%
%
With a slight abuse of notation, we use $G' \in \calI$ if $G'$ is obtainable from $G$ by applying the perturbations given by $\calI$.
We summarize existing metrics in \cref{tab:faith-metrics-unreliable}.


\section{\SEGNNs's explanations that do not explain}
\label{sec:failure-cases}

In this section, we study \SEGNNs through the lens of their loss functions, and provide a sufficient condition to achieve optimal true risk for several representative \SEGNNs, namely, \GSAT \citep{miao2022interpretable}, \LRI \citep{miao2022interpretablerandom}, \CAL \citep{Sui2022cal}, \GMTLin \citep{chen2024howinterpretable}, and \SMGNN \citep{azzolin2025formal}.
Surprisingly, \textit{this condition is met by degenerate explanations highlighting recurrent patterns with no class-discriminative power}, like background pixels for object recognition or punctuation for text classification.

We formalize a class of such recurrent patterns for graphs as an \textbf{anchor set}: a set of single-node subgraphs $\mathcal{Z} =\{z_i\}_{i=1}^m$ where $m=|\calY|$ and $z_i \subseteq G$, for all instances $G \in \calG$ and for all $z_i \in \calZ$.
Anchor sets can be seen as a collection of prototypes, one for each class, appearing in every graph.
As an example, green (\green) and violet (\violet) nodes form an anchor set $\mathcal{Z}$ in \cref{fig:example}.
Nodes in an anchor set have no class-discriminative power, as they appear in every sample.
Nonetheless, we show that picking them as explanations can be an optimal solution for \SEGNNs.

\begin{restatable}{theorem}{degminimalloss}
\label{thm:deg-suff-condition}
    Let $\calD_{\calG \times \calY}$ be a data distribution with deterministic ground truth labeling function $\phi: \calG \mapsto \calY$,
    %
    \DET be a hard explanation extractor, and $\calZ=\{z_y\}_{y \in \calY}$ be an anchor set.
    Then, there exists an explanation extractor $\DET(G) \defeq z_{\phi(G)}$ and a classifier $\CLF(z_{y}) \defeq y$, such that the SE-GNN $\CLF \circ \DET$ implemented by \GSAT, \LRI, \CAL, \GMTLin or \SMGNN achieves optimal true-risk.
\end{restatable}

The proof is provided in \cref{appx:proofs}.
\cref{thm:deg-suff-condition} shows that the explanation extractor can use any recurrent set of nodes as label-encoding explanations for the classifier.
%
%
%
These explanations are \textbf{unambiguously unfaithful}.  To see this, consider a highly accurate \SEGNN.
If it relied solely on explanations from the anchor set $\calZ$, it would be unable to predict the label at all, as these nodes are constant across graphs. In order to attain high accuracy, it \textit{must} be looking at other parts of the input graph, yet these are not revealed by the explanation.
Hence, \Cref{thm:deg-suff-condition} shows that SE-GNNs with optimal true risk can provide completely unfaithful explanations. 
This gives rise to rather counterintuitive degenerate explanations, as depicted in the following example.

\begin{example}
\label{ex:red-blue}
    Let us consider the binary graph classification dataset \BAColorGV, where graphs are composed of red and blue nodes randomly connected (see \cref{fig:example}).
    Positive graphs are those where the number of blue nodes ($\# blue$) exceeds the number of red nodes ($\# red$).
    Every instance further contains two single isolated nodes, one green ($u_{green}$) and one violet ($u_{violet}$).
    Therefore, green and violet nodes form a valid anchor set $\calZ$, and are completely irrelevant to the task that only depends on red and blue nodes.
    However, the following \SEGNN's explanation extractor \DET and classifier \CLF can achieve perfect task accuracy while providing explanations consisting of green and violet nodes:
    \begin{align}
    \label{eq:ex-red-blue}
        R = \DET(G) = 
        \begin{cases}
            u_{green}  & \text{if } \# red \ge \# blue \\
            u_{violet} & \text{if } \# blue > \# red \\
        \end{cases}
        \quad\quad
        \CLF(R) =
        \begin{cases}
            0   & \text{if } R=u_{green}\\
            1   & \text{if }  R=u_{violet}\\            
            0.5 & \text{otherwise}
        \end{cases}
    \end{align}
\end{example}

Notably, \cref{thm:deg-suff-condition} can be extended beyond sets of nodes appearing in all graphs (see \cref{appx:subgraph-anchor-sets} for a formal analysis):
First, if members of mutually-exclusive sets of nodes appear across graphs, the explanation extractor can partition them into class-specific blocks $\{\{z_0, \dots, z_n\}_y\}_{y \in \calY}$, which the classifier then maps to the respective labels.
Second, the explanation extractor can encode predictions inside subgraphs if no node-level anchor set (or no class-specific partition) exists.
In fact, we discuss in \cref{sec:naturaldeg} an \SEGNN where the optimization avoids single-node degenerate explanations but still highlights task-irrelevant subgraphs.


A consequence of \cref{thm:deg-suff-condition} is that the same explanation can appear for opposite classes for different optimal \SEGNNs, \ie explanations fail to be consistent \citep{dasgupta2022framework}. 
%
%
Specifically, given an optimal \SEGNN as per \cref{thm:deg-suff-condition} and a permutation $\pi: \calY \mapsto \calY$ of class indices, it is possible to construct another \SEGNN, say $\text{\SEGNN}_\pi$, composed of 
$$\DET_\pi(G) \defeq z_{\pi(\phi(G))} \quad \text{and} \quad \CLF_\pi(z_y) \defeq \pi^{-1}(y)$$ 
that also achieves optimal true risk.
This is made intuitive by \cref{ex:red-blue}, where it is possible to swap $u_{green}$ with $u_{violet}$ across classes in \cref{eq:ex-red-blue} without altering the final predictions.

As a final remark, although \cref{thm:deg-suff-condition} covers a broad range of \SEGNNs, alternative formulations outside its scope exist, such as \DIR \citep{wu2022discovering} and \SUNNY \citep{deng2024sunny}.
These models enforce compact explanations through an explanation-selection mechanism retaining a fixed ratio $K$ of nodes, with $K$ a hyperparameter. 
We will empirically show in \cref{sec:naturaldeg} that an inappropriate choice of this hyperparameter can, however, still force the label to be encoded in degenerate explanations.


\textbf{Takeaways.}
\cref{thm:deg-suff-condition} shows that \SEGNN explanation extractor $e$ can output perfect label-encoding subgraphs which, with an appropriate classifier $g$, provide optimal true-risk. 
%
Yet such label-encoding subgraphs can be completely degenerate.
%
%
These explanations are highly unfaithful, as they only need to convey an encoding of the label to the classifier, without giving any insight into how the overall model arrives at the decision.
%
Next, we show that a malicious attacker can exploit this failure case to make \SEGNNs output fabricated explanations concealing the model's inner workings.

\section{RQ1: Can \SEGNN Explanations be Manipulated?}
\label{sec:segnns-can-be-manipulated}

We introduce an attack to \SEGNNs aimed at training accurate models that, however, provide explanations unambiguously unrelated to how \SEGNNs infer their predictions.
This is significant in two distinct ways:
First, it shows how an attacker can manipulate explanations without impairing accuracy, potentially misleading users and concealing reliance on protected features.
Second, it offers a controlled, reproducible setup for evaluating faithfulness metrics, as we will discuss in \cref{sec:faithfulness-metrics}.

\textbf{Setup}
Our attack comprises two main steps:
First, the attacker defines a designated malicious explanation for each class;
Second, they encourage an \SEGNN to output the designated explanation while optimizing for downstream performance.
In practice, it amounts to training the \SEGNN using a regular classification loss $\calL_{cl\!f}$ paired with a binary cross-entropy loss $\calL_{expl}$ designed to push the relevance scores of nodes belonging to the designated explanation $p_u^{y}$ to $1$ and the rest to $0$, that is:
\begin{align}
    \label{eq:loss-attack}
    \min_{\CLF, \DET} \sum_{G, y} \calL_{cl\!f}(\CLF, \DET, G, y) + \calL_{expl}(\DET, G, y),
    \qquad
    \calL_{expl}(\DET, G, y) := \frac{1}{|V|} \sum_{u \in V} \text{BCE} \big (\DET(G)_u, p_u^{y} \big).
\end{align}
%
%
%
To ensure $\calL_{expl}$ does not conflict with other regularization terms, it is preferable to deactivate these.
We provide further details in \cref{appx:details-attack-malicious}.

\textbf{Datasets and designated explanations}  We empirically evaluate this attack on one synthetic dataset -- \BAColorGV, introduced in \cref{ex:red-blue} -- and three real-world datasets: \MNIST \citep{knyazev2019understanding}, \MUTAG \citep{debnath1991structure}, and \SSTP.\footnote{\SSTP is a variant of \SST \citep{yuan2022explainability} where we ensure a single ',' and a single '.' to be present in every graph. To avoid the emergence of unwanted correlations, we append them as isolated nodes and with fixed embeddings $\vec{0}$ (for ',') and $\vec{1}$ (for '.'). More details available in \cref{appx:datasets}.}
To underscore the counterintuitive nature of our results, we will focus on designated explanations belonging to an anchor set, which are therefore \textit{unrelated to the task being solved}.
Specifically, our designated explanations include only nodes that \textit{i)}  are not predictive of the label, and \textit{ii)} can be used by the \SEGNN to encode its prediction.
Examples include green/violet nodes for \BAColorGV, background pixels for \MNIST, and punctuation for \SSTP.
We report in \cref{appx:designated-explanations-details} the full list of the designated explanations for each dataset in use in our experiments.
Under ideal circumstances, a model that always and only relies on these unrelated subgraphs should be weakly predictive of the true label.
Surprisingly, however, \SEGNNs can be trained to achieve high accuracy even when using them as explanations.  As it turns out, they can do so by, in essence, \textit{encoding the predicted label in the (relevance scores of the) explanation itself}, even when the highlighted subgraph alone is uninformative for the label.

\textbf{Evaluation metrics}
We assess the success of the attack by comparing the task accuracy of the attacked model with that of an unmanipulated baseline.
Also, to evaluate how well the predicted explanations align with the designated ones, we compute the $F_1$ score\footnote{We prefer $F_1$ over \AUCROC \citep{luo2020parameterized}, which ignores non-sparse yet accurate relevance scores.} between predicted and designated relevance scores.
Specifically, for each node, we binarize relevance scores -- as $\Ind{p_u > 0.9}$ if $p_u^{y}=1$ and $\Ind{p_u > 0.1}$ if $p_u^{y}=0$ -- to penalize uncertain relevance scores, and then report the average across classes.
The $F_1$ score is computed only on nodes from correctly classified graphs, as designated explanations cannot be learned if the underlying model fails to infer the right label.

\textbf{Results}  \cref{tab:attack-models} lists the results of attacking three representative \SEGNN architectures, namely \GSAT \citep{miao2022interpretable}, \DIR \citep{wu2022discovering}, and \SMGNN \citep{azzolin2025formal}, averaged over five random seeds.
Crucially, the attack always succeeds: all attacked models output explanations similar to the designated ones ($F_1 \ge 92\%$).
The exception is \SMGNN on \SSTP ($F_1 \approx 59\%$).  This is because the \SSTP test set includes Out-Of-Distribution (OOD) graphs \citep{wu2022discovering}, which are inherently more difficult to recognize for models not designed for OOD generalization.  In fact, computing the $F_1$ on the in-distribution validation set yields a much higher value ($F_1 = 96.5 \pm 0.8$), supporting this observation.
We report several example explanations for each dataset in \cref{appx:examples-attacks}.

\input{tables/attack-models-macro}
Yet, attacked models perform comparably to their unmanipulated counterparts.
The exception is \SSTP, where attacked models slightly underperformed: the difference is however within one standard deviation for two models out of three.
Another interesting case is \MNIST: while the manipulated \GSAT performs slightly worse ($-1\%$ Acc), \DIR and \SMGNN perform surprisingly better ($+50\%$ and $+5\%$ Acc, respectively).
This suggests that encoding the predicted label into the explanation can help models improve their own prediction accuracy during training.  We will expand on this in \cref{sec:naturaldeg}.

For transparency, we report in \cref{appx:failures-attacks} some cases in which the attack failed to match the designated explanation exactly.  Nonetheless, even in these cases, we show that it still managed to force the models into concealing the features they are truly relying on.

These results allow us to answer \textbf{RQ1} in the affirmative: in the vast majority of cases, \textit{our attack successfully teaches \SEGNN{s} to output degenerate explanations unrelated to the task at hand, all without degrading (and sometimes improving) predictive performance}.

\textbf{Consequences for evaluating plausibility.}
Our attack can be naturally employed to teach \SEGNNs to output plausible but unfaithful explanations. 
We test this empirically in \cref{app:plausible-but-unfaithful-experiment}, due to space constraints:  in this experiment, the attacked \SEGNNs learn to output highly plausible explanations, \ie explanations very close to human-defined ground-truth explanations, but are shown to also rely on attributes that do \textit{not} appear in the predicted explanations.
This result entails that evaluating the plausibility of explanations \citep{miao2022interpretable, wu2022discovering, chen2024howinterpretable, tai2025redundancy}, \ie how well they match human expectations, does not reflect their trustworthiness.


\section{RQ2: Can Unfaithful Explanations go Undetected?}
\label{sec:faithfulness-metrics}

As we just saw, plausibility metrics cannot detect unfaithful explanations.  The question is whether metrics explicitly designed to measure faithfulness, in fact, can.
This question is currently unanswered:  so far, faithfulness metrics have been validated either by comparing them against ground-truth explanations or by correlating them with rankings induced by progressively randomized explanations \citep{fang2023evaluating, zheng2023robust, zheng2025ffidelity}.  Neither setup explicitly tests whether they can identify known-unfaithful explanations.

We fill this gap by proposing an empirical benchmark (\cref{sec:empirical-benchmark}) that exploits the failure cases outlined in \cref{sec:failure-cases} for evaluating faithfulness metrics in practice.
Overall, our results indicate that existing faithfulness metrics can be surprisingly ineffective.  To amend this issue, we introduce a simple but more reliable metric (\cref{sec:proposing-new-metric}).

\subsection{Benchmark of faithfulness metrics}
\label{sec:empirical-benchmark}

\cref{sec:segnns-can-be-manipulated} shows that one can manipulate \SEGNNs to output degenerate unfaithful explanations.
In this section, we exploit this result to construct a controlled benchmark where faithfulness metrics are evaluated based on how well they can judge known-unfaithful explanations, where these are extracted from the manipulated models of \cref{sec:segnns-can-be-manipulated}.

\textbf{Experimental setup:}
We select a set of representative faithfulness metrics from \cref{tab:faith-metrics-unreliable}, one for each family of perturbation. 
We also include the Counterfactual Fidelity (\COUNTERFID) \citep{chen2024howinterpretable}, a metric that does not readily fit into \cref{tab:faith-metrics-unreliable}, as it perturbs \textit{both} the complement and the explanation based on the raw relevance scores, which may not be available \citep{ma2025gnnstealing}.
Details about metric implementations are provided in \cref{appx:detail-metrics}.

To find a common evaluation setup across metrics, we adopt a similar setting as \citet{bhattacharjee2024ngng} where an auditor is asked to validate a set of explanations given by an external provider, rejecting those that are unfaithful.
In particular, given a set $D=\{(G_i, R_i)\}$ of input graphs and associated explanations, we define the rejection ratio for metrics estimating \textit{sufficiency} as:
\begin{equation}
\label{eq:rejection-ratio}
    \mathsf{RejRatio}_\calI (D) = \frac{1}{|D|} \sum_{(G,R) \in D} \max_{G' \in \calI} \Ind{ \CLF(\DET(G)) \ne \CLF(\DET(G'))}
\end{equation}
$\mathsf{RejRatio}_\calI$ computes the fraction of instances deemed unfaithful by each metric, where $\calI$ is the set of perturbations allowed by the metric in use (cf. \cref{tab:faith-metrics-unreliable}).
For \textit{necessity} metrics, the inner condition is replaced with $\Ind{ \CLF(\DET(G)) = \CLF(\DET(G'))}$.
To make the computation tractable, \cref{eq:rejection-ratio} is evaluated on a budget of perturbations per sample, fixed to $50$ across metrics.
Details are available in \cref{appx:details-faithfulness-metrics}.

\input{tables/rejections}

\textbf{Results:}
\cref{tab:rejection-attack-models} reports the $\mathsf{RejRatio}_\calI$ values computed for the attacked models of \cref{sec:segnns-can-be-manipulated} and, for completeness, the raw metrics themselves in \cref{tab:rejection-attack-models-metric}.
Since those models are trained to output unfaithful explanations intentionally, \textit{we aim to observe as large rejection ratios as possible}.
%
However, \cref{tab:rejection-attack-models} shows that some metrics can catastrophically fail to reject these explanations, as shown by the ratios around zero for \SUF, \RFIDM, and \COUNTERFID in \BAColorGV, for \FIDM in \SSTP, and \FIDP in \MUTAG.

In conclusion, these results positively answer \textbf{RQ2}: \textit{previous faithfulness metrics can fail to detect the known-unfaithful explanations extracted in \cref{sec:segnns-can-be-manipulated}}.
To solve the undetectability of unfaithful explanations, we introduce an evaluation metric that is shown to be more robust.

\subsection{Spotting the failure cases: A reliable faithfulness metric}
\label{sec:proposing-new-metric}

We observe that existing metrics can fail because each of them focuses on a restricted set of perturbations, such as edge or complement removals, and for each of them, it is possible to construct unfaithful explanations evading detection under the chosen perturbations.
For instance, \textit{sufficiency} metrics that only perturb edges cannot probe models that rely primarily on nodes for their decisions.\footnote{Although GNNs ideally depend on both topology and node features \citep{maya2025position}, a reliable faithfulness metric should remain robust when only one of these is truly relevant.}
%
%
This is best seen in \cref{ex:red-blue}, where the explanations $u_{green}$ and $u_{violet}$ will go undetected by any metric perturbing only edges, and $u_{green}$ will also be undetected by any metric testing only complement removals, 
as feeding \cref{eq:ex-red-blue} with $G'=u_{green}$ preserves the prediction.
This observation generalizes to the case where \textit{sufficiency} and \textit{necessity} are evaluated jointly, as discussed in \cref{appx:suff-nec-jointly}.

Building on these observations, we propose the \textit{Extension Sufficiency Test} (\SUFFCAUSE) metric:

\begin{definition}[\SUFFCAUSE]
\label{def:suff-cause}
    Let $G$ be an input graph with prediction $\CLF(\DET(G))$ and associated explanation $R=\DET(G)$, and $d(\cdot)$ a suitable distance measure. Then, \SUFFCAUSE estimates the \textit{sufficiency} of $R$ as follows:
    \begin{equation}
    \label{eq:suffcause-metric}
          \SUFFCAUSE(R, G) = \max_{R \subseteq G' \subseteq G} d \big ( \CLF(\DET(G)), \CLF(\DET(G')) \big).
    \end{equation}
\end{definition}


\SUFFCAUSE is a simple but robust faithfulness metric that avoids the previously underlined issues by holistically considering \textit{all} supergraphs $G'$ of $R$ contained in $G$, \textit{regardless of what perturbations are necessary for constructing them from $R$}.
%
%
%
%
%
%
%
Note that \SUFFCAUSE, similarly to previous metrics \citep{hase2021oodxaiproblem, zheng2023robust}, may create an OOD supergraph $G'$ whose prediction differs from that of $G$ just because it is OOD, leading to a large (i.e., worse) \SUFFCAUSE value regardless of whether $R$ was faithful. 
However, without access to the model or to the data distribution, it is impossible to know whether $G'$ is OOD or if it was obtained by perturbing the features the model is relying on.
For this reason, we prioritize a more conservative metric that, although it may flag faithful explanations as unfaithful, it reliably marks the unfaithful ones as such, minimizing the risks of users being deceived.
We provide more details and an algorithm implementing \SUFFCAUSE in \cref{appx:detail-metrics} and \cref{alg:suffcause}.

To avoid the computationally prohibitive cost of enumerating all possible supergraphs of $R$ within $G$ in \cref{def:suff-cause}, we fix a budget of supergraphs to test in practice.
Also, note that \SUFFCAUSE estimates only the \textit{sufficiency} of explanations; since the classifier \CLF may be sensitive to changes in the explanation even when unfaithful, \textit{necessity} metrics are unsuitable for identifying the failure cases in \cref{sec:failure-cases} (see \cref{sec:are-nec-metrics-nec}).
We additionally leverage existing literature on \SEGNNs to theoretically characterize which family of explanations is deemed unfaithful by \SUFFCAUSE, by relating \cref{def:suff-cause} with increasingly weaker formal notions of explanations \citep{azzolin2025formal}.
%
This analysis, reported in \cref{appx:characterize-formal-notions}, shows that \SUFFCAUSE provably labels the class of explanations including $u_{green}$ and $u_{violet}$ in \cref{ex:red-blue} as unfaithful.
We now proceed to analyze how \SUFFCAUSE fares at the benchmark introduced in \cref{sec:empirical-benchmark}.

\textbf{Results.}
The last column of \cref{tab:rejection-attack-models} lists the results for \SUFFCAUSE, which is the only metric consistently achieving the highest -- or close to thereof -- rejection ratios across every configuration while keeping a relatively low standard deviation.
In particular, in all the cases where previous metrics score a rejection ratio around zero, \SUFFCAUSE rejects at least $\approx50\%$ of degenerate explanations.
%
%
%
This result can be easily further improved by increasing the budget of allowed perturbations. In fact, we show in the appendix (\cref{fig:budget-ablation}) that the rejection ratio of \SUFFCAUSE monotonically increases with the budget, reaching $\approx65\%$. 
Other metrics, instead, either do not benefit from larger budgets or flatten around $\approx17\%$.
In addition, \SUFFCAUSE also correctly recognizes non-degenerate explanations that include all relevant information, which are instead erroneously rejected by \SUF and \COUNTERFID, as shown in \cref{appx:rejecting-faithful}.

We proceed to analyze how \SUFFCAUSE behaves in natural settings, and to verify whether the failure cases of \cref{sec:failure-cases} can in fact emerge even without malicious attacks.


\section{RQ3: Can degenerate explanations emerge naturally?}
\label{sec:naturaldeg}

Finally, we show that \SEGNNs can output degenerate explanations -- similar to those presented in \cref{sec:segnns-can-be-manipulated} -- without any explicit manipulation, and that \SUFFCAUSE is effective in finding them.


In practice, we compute \SUFFCAUSE's rejection ratios for each \SEGNN across the dataset and complement this with a qualitative analysis of explanation examples, reported in the appendix.
This analysis aligns with \SUFFCAUSE's results: when the rejection ratio is high, explanations highlight little-class-discriminative subgraphs; when rejections are near zero, explanations emphasize class-discriminative subgraphs, often aligning with human expectations.

\input{tables/natural-degeneracy}

\textbf{Setup:}
We train the \SEGNN architectures in \cref{tab:attack-models} by following their original protocol and natural losses, tuning their hyperparameters to encourage explanation sparsity.
For \DIR, the original implementation selects explanations based on a relatively large top-$K$; we train the model using a much smaller $K$ instead, as indicated in the results table.  This is useful to show that \DIR can still encode label-relevant information inside a small subgraph even when the subgraph has no discriminative power by itself.
For each model, we report the accuracy, the \AUCROC with the ground truth explanations when these are available, and compare the rejection ratios of \SUFFCAUSE with those of \FIDM and \RFIDM (as baselines).
Further details on the experimental setup are available in \cref{appx:details-naturaldeg}.

\textbf{Results:}
As shown in \cref{tab:natural-degeneracy}, the results are largely consistent with those for naturally trained models in \cref{tab:attack-models}, with few exceptions.
On \BAColorGV and \MNIST, \SMGNN achieves a modest decrease in accuracy ($\approx 2\%$).
For \DIR, instead, the small $K$ value hinders learning on \MUTAG, and the accuracy on \MNIST ($\approx 20\%$) confirms \DIR's unsuitability for this task, in line with the accuracy reported in \cref{tab:attack-models} ($\approx 40\pm20\%$).
In \BAColorGV and \SSTP, reducing \DIR's $K$ substantially (from $50\%$ to $1\%$ and from $60\%$ to $10\%$, respectively) results in only a $\approx 2\%$ drop in accuracy.
This is remarkable, as \DIR does not fit in \cref{thm:deg-suff-condition}, and yet can encode label-relevant information inside degenerate explanations, as illustrated in \cref{fig:naturaldeg-BAColorGV-DIR} and \cref{fig:naturaldeg-SSTP-DIR}.
%
Interestingly, the behavior of \SUFFCAUSE is contrasting: rejection ratios are high for certain models but almost zero for others, calling for a more detailed analysis, reported below.
A case-by-case discussion for some selected examples is reported in the figures listed in \cref{tab:natural-degeneracy}.
Our major findings are as follows:

\textbf{i) \SEGNNs can output degenerate explanations:}
We provide examples of naturally occurring degenerate explanations in \cref{fig:naturaldeg-BAColorGV-DIR}, \cref{fig:naturaldeg-BAColorGV-SMGNN}, and \cref{fig:naturaldeg-MNIST-SMGNN}, where \SEGNNs highlight green and violet nodes for \BAColorGV, and background pixels for \MNIST, respectively.
%
%
Similar examples are also found for \DIR on \SSTP, where explanations highlight punctuation and stop words (\cref{fig:naturaldeg-SSTP-DIR}), and for \GSAT and \SMGNN on \MUTAG, where explanations emphasize individual weakly class-discriminative atoms, rather than meaningful functional groups (\cref{fig:naturaldeg-MUTAG-GSAT}, \cref{fig:naturaldeg-MUTAG-SMGNN}).
In all cases, \SEGNNs retain high predictive performance, and \SUFFCAUSE rejects a substantial fraction of explanations ($\ge 50\%$)

\textbf{ii) \SEGNNs can also output non-degenerate explanations:}
\SEGNNs do not always fail, and the low rejection ratios ($\approx2\%$) of \GSAT on \MNIST and \SSTP suggest the model indeed outputs \textit{sufficient} explanations.
In fact, \GSAT correctly extracts the digit-subgraph in \MNIST, as indicated by the high \AUCROC (cf. \cref{fig:naturaldeg-MNIST-GSAT}). In \SSTP, however, it simply outputs the entire graph as explanation, see \cref{fig:naturaldeg-SSTP-GSAT}.
Furthermore, on \SSTP, \SMGNN consistently highlights emotion-laden words and achieves around zero rejection ratios (see \cref{fig:naturaldeg-SSTP-SMGNN}).
%
Note that these results do not contradict \cref{thm:deg-suff-condition}, as it provides existential conditions under which \SEGNNs \textit{can} extract degenerate unfaithful explanations. 
In practice, stochastic optimization can still steer the model to select other, non-degenerate explanations.
Notwithstanding, this behavior remains beyond the practitioner’s control, motivating further investigation into more robust \SEGNNs in future work.

\textbf{iii) Faithfulness metrics can fail in the wild:}
\FIDM and \RFIDM achieve high rejection ratios in \BAColorGV, but exhibit substantial seed-to-seed variability, reducing their reliability.
For instance, the examples reported in \cref{fig:naturaldeg-BAColorGV-DIR} and \cref{fig:naturaldeg-BAColorGV-SMGNN} highlight severe failure cases where these metrics mark most of the explanations composed of green and violet nodes as faithful.
Similarly, \cref{fig:naturaldeg-MNIST-SMGNN} and \cref{fig:naturaldeg-SSTP-DIR} show that \FIDM and \RFIDM can fail to reject explanations highlighting background pixels or punctuation for \MNIST and \SSTP, respectively.
In contrast, \SUFFCAUSE achieves more consistent results, and it proved helpful in spotting explanations containing only irrelevant information.

In conclusion, the answer to \textbf{RQ3} is positive: \textit{naturally trained \SEGNNs can output degenerate explanations, and popular faithfulness metrics can fail to mark them as such.}
Conversely, \SUFFCAUSE proved effective in distinguishing explanations containing or omitting key relevant features.


\section{Related work}
\label{sec:related-work}

Popular faithfulness metrics (see \cref{sec:background} and \cref{tab:faith-metrics-unreliable} for a taxonomy) are typically evaluated by their correlation with \textit{i)} human-defined ground-truth explanations or \textit{ii)} rankings from progressively randomized explanations \citep{fang2023evaluating, christiansen2023faithful, zheng2023robust, zheng2025ffidelity}, but are rarely tested to flag known-unfaithful explanations as such. To address this, in \cref{sec:empirical-benchmark} we introduce a controlled benchmark where \SEGNNs are trained to output known-unfaithful explanations, and metrics are assessed by the fraction of explanations they correctly reject.
%

Prior work has shown that explanations can be manipulated in ways that evade detection \citep{dombrowski2019explanations, heo2019fooling, slack2020fooling, slack2021counterfactual}, with most studies focusing on post-hoc methods for tabular or image data.
Furthermore, similar failure cases to the ones presented in \cref{sec:failure-cases} can be found in rationalization methods for text \citep{yu2019rethinking}, Neuro-Symbolic architectures \citep{marconato2023not}, and Concept Bottleneck Models \citep{bortolotti2025shortcuts}.
\citet{tai2025redundancy} has highlighted that when not regularized for sparsity, \SEGNN{s} extract redundant and potentially unfaithful explanations.  This work is complementary to ours, in that we show that when \SEGNNs are strongly regularized for sparsity, their explanations can also be unfaithful.
We provide an extended discussion on some of this related work in \cref{appx:related-work} due to space constraints.

\section{Conclusion}
\label{sec:conclusions}

We outlined a critical failure case of \SEGNNs whereby they output explanations completely unfaithful to their actual inner workings.
%
We provided a sufficient condition under which several \SEGNNs achieve optimal true risk and demonstrated that this condition is met by explanations with no class-discriminative power, which the model could not have used to achieve high accuracy (\cref{sec:failure-cases}).
We then showed that a malicious attacker can exploit this fallacy to deliberately conceal the features the model relies on (\cref{sec:segnns-can-be-manipulated}), potentially hiding the use of protected attributes.
We also observed that these degenerate explanations can emerge naturally (\cref{sec:naturaldeg}), underscoring the need for reliable auditing of explanations.
Motivated by this, we proposed a benchmark for faithfulness where faithfulness metrics are tested based on how many known-unfaithful explanations they reject, and showed that popular metrics perform poorly (\cref{sec:empirical-benchmark}).
Finally, to address this shortcoming, we introduced a new metric that is shown to be more effective (\cref{sec:proposing-new-metric}).

\textbf{Limitations.}
To highlight the counterintuitive aspect of our analysis, \cref{thm:deg-suff-condition} is provided for a restricted but representative setting: explanations composed solely of nodes from an anchor set, which the model cannot use in isolation to achieve high accuracy.
Nonetheless, it shows that \SEGNNs can output explanations that do not faithfully represent what the model is actually using for inferring predictions, opening the door to several unfaithful behaviors beyond explanations composed of anchor sets.
To improve on this, we provide in \cref{appx:subgraph-anchor-sets} an extended analysis where we generalize the notion of anchor set to the case of subgraphs, showing how popular \SEGNNs can still prefer uncorrelated subgraphs over other, more informative explanations.
In both analysis, we assume the explanation extractor to be \textit{hard}: This is a natural and commonly found desideratum for extracting explanations \citep{yu2020graph, azzolin2025formal}, as it represents an extractor maximally confident of what is relevant and what is irrelevant; This desideratum is already included in the design of practical \SEGNNs, which employ TopK masking \citep{wu2022discovering, deng2024sunny}, Gumbel-softmax trick \citep{miao2022interpretable, miao2022interpretablerandom}, or entropy regularization \citep{lin2020graph, azzolin2025formal}, to push the relevance scores to extremal values.
While requiring access to the \SEGNN's training process is a strong assumption for our attack in \cref{sec:segnns-can-be-manipulated}, it naturally aligns with several practical Machine Learning-as-a-Service scenarios, where the service provider has full control of the model.
Furthermore, despite \SUFFCAUSE effectively detecting unfaithful explanations omitting key relevant features, it cannot identify unfaithful behaviors arising from redundancy, i.e., explanations containing both truly relevant elements and irrelevant ones.
Finally, although our focus is on graph classification, our findings carry over directly to node classification. For example, consider a variant of \BAColorGV in which each node must be classified based on whether it has more red or blue neighbors; In this case, \SEGNNs can still encode the overall count of colored neighbors within green and violet nodes, similarly to \cref{ex:red-blue}.

\subsection*{Acknowledgement}

Funded by the European Union. Views and opinions expressed are however those of the author(s) only and do not necessarily reflect those of the European Union or the European Health and Digital Executive Agency (HaDEA). Neither the European Union nor the granting authority can be held responsible for them. Grant Agreement no. 101120763 - TANGO and Grant Agreement no. 101120237 - ELIAS.
SM acknowledges the support of FWF and ANR project NanOX-ML (6728).

\subsection*{Ethics Statement}

All authors have read and approved the ICLR Code of Ethics.  As for societal consequences, the aim of this work is to shed light on potential misuse of interpretable-by-design neural models, warning users from blindly trusting explanations, and providing a more reliable tool to audit explanations. It can therefore contribute to the development of more trustworthy models for graph-based data and more reliable certifications of explanation validity.

\subsection*{Reproducibility Statement}

The proofs of our theoretical results in \cref{sec:failure-cases} are provided in \cref{appx:proofs}.
Details about the experiments, along with details about datasets and metric implementation, are available in \cref{appx:impl-details}.
Our code is available at \url{https://github.com/steveazzolin/gnn_deg_expl}.

\bibliography{ref/explanatory-supervision, ref/references}
\bibliographystyle{iclr/iclr2026_conference}

\appendix

\section{Use of LLMs}

LLMs were used to polish the writing, to rephrase sentences, and to debug the code.
Our manuscript and our code was first human-generated, and then possibly enhanced by LLMs.

\section{Proofs}
\label{appx:proofs}

\subsection{Proof of \cref{thm:deg-suff-condition}}
\degminimalloss*
\begin{proof}
    Recall that an \SEGNN $\CLF \circ \DET$ is composed of an explanation extractor $\DET$ and a classifier $\CLF$.
    Also, recall that $\calG$ is the set of graphs and $\calY$ the set of labels with deterministic ground truth labeling function $\phi: \calG \mapsto \calY$, and that $\calZ$ is an anchor set, i.e., a set of single-node subgraphs $\mathcal{Z} =\{z_i\}_{i \in \calY}$ such that $z_i \subseteq G$, for all instances $G \in \calG$ and for all $z_i \in \calZ$.
    Then, we show that selecting as the explanation extractor $\DET(G) \defeq z_{\phi(G)}$ and as the classifier $\CLF(z_{y}) \defeq y$ achieves optimal true-risk for \GSAT, \LRI, \CAL, \GMTLin, and \SMGNN.
    At a high level, this shows that whenever $\DET$ can encode the predicted label inside a single node, then the classifier $\CLF$ implementing a simple mapping from these nodes to labels achieves optimal true risk for several \SEGNNs, even when the selected nodes are uncorrelated with the ground truth task.

    %
    We restate below two assumptions relevant to our proof:

    \begin{assumption}
    \label{assump:discrete-extractor}
        \DET is a hard explanation extractor, i.e., it can only output relevance scores saturated in $\{0,1\}$ for \SMGNN and \CAL, and $\{r,1\}$ for \GSAT, \LRI, and \GMTLin.
    \end{assumption}
    \begin{assumption}
    \label{assump:nonempty-expl} 
        The explanation $R$ cannot be empty, \ie $|R|>0$. In other words, and following \cref{assump:discrete-extractor}, $\exists u \in V: p_u=1$.
    \end{assumption}
    %
    %
    Note that $r$ in \cref{assump:discrete-extractor} is the parameter controlling the degree of stochasticity in the relevance scores of each node, and it therefore acts as an uninformative baseline value.
    For example, a value of $p_u=1$ indicates no stochasticity for node $u$, i.e., maximally relevant, while $p_u=r$ indicates maximum stochasticity, i.e., low relevance \citep{miao2022interpretable}.

    Without any loss of generality, we will consider a binary classification setting, i.e., $|\calY|=2$, and two distinct nodes, $z_0, z_1$ belonging to the anchor set, that is $z_0, z_1 \in \calZ$.
    We proceed to show that, for each of the following \SEGNN, an explanation extractor selecting $z_0$ for $\phi(G)=0$ and $z_1$ for $\phi(G)=1$, paired with a suitable classifier $\CLF$, will yield optimal risk.
    We will analyze each \SEGNN separately, as follows.

    \textbf{\GSAT \& \LRI \citep{miao2022interpretable, miao2022interpretablerandom}:}
    The learning objective of both \GSAT and \LRI is:
    \begin{equation}
    \label{eq:gsat-objective}
        \calL_{cl\!f}(\CLF, \DET, G, y) + \lambda_1 \sum_{u \in V} p_{u}\log(\frac{p_{u}}{r}) + (1-p_{u})\log(\frac{1-p_{u}}{1-r})
    \end{equation}
    where $V$ is the set of nodes for input graph $G$, $p_{u}$ is the explanation relevance score for node $u$, $r \in [0,1]$ is the hyperparameter controlling the uninformative prior below which a node is considered to be non-relevant \citep{miao2022interpretable}, and $\lambda_1$ is an hyperparameter controlling the relative strength of the regularizer over the cross entropy loss $\calL_{cl\!f}$.

    We now provide a suitable pair of explanation extractor and classifier achieving an optimal value for \cref{eq:gsat-objective}.
    For exposition convenience, instead of considering $\DET$ as a mapping from graph to subgraphs as described in \cref{sec:background}, we will unpack it as a function from graph to individual node relevance scores, as follows:
    \begin{align}
    \label{eq:gsat-proof-det-clf}
        p_u \defeq \DET(G)_u = 
        \begin{cases}
            1 & \text{if } u = z_0 \land \phi(G) = 0 \\
            1 & \text{if } u = z_1 \land \phi(G) = 1 \\
            r & \text{otherwise }
        \end{cases}
        \quad\quad
        \CLF(R) =
        \begin{cases}
            0 & \text{if } p_{z_0} > p_{z_1}\\
            1 & \text{if } p_{z_1} > p_{z_0}\\            
            0.5 & \text{otherwise}
        \end{cases}
    \end{align}

    Since \CLF classifies every sample correctly with maximum confidence, $\calL(\CLF, \DET, G, y)$ equals zero. 
    Also, since each $p_u$ equal to $r$ yields zero loss and \DET highlights only a single node between $z_0$ and $z_1$ per graph, the second term in \cref{eq:gsat-objective} boils down to $-\lambda_1 \log(r)$.
    By \cref{assump:discrete-extractor} and \cref{assump:nonempty-expl}, it is not possible to find an assignment of explanation scores with more terms equal to $r$ than the explanation provided by \cref{eq:gsat-proof-det-clf}, hence $-\lambda_1 \log(r)$ is the minimum risk attainable.

    As a remark, note that the classifier in \cref{eq:gsat-proof-det-clf} can indeed be learned by a linear layer with sigmoid output, receiving as input an embedding of the form $\vh = [p_{z_1} \Ind{z_1 \in R}, p_{z_0} \Ind{z_0 \in R}]$, with weights $\vw = [\alpha, -\alpha]$ and $b=0$, for a sufficiently large $\alpha$.
    Note that $\vh$ can be recovered by a classifier implementing a GNN with weighted message passing and sum global readout by simply mapping $z_0$ and $z_1$ to two different one-hot encodings, and multiplying this value by the weight given by the relevance scores. Hence, the classifier expressed in \cref{eq:gsat-proof-det-clf} is easily realizable by practical GNNs.

    \textbf{\GMTLin \citep{chen2024howinterpretable}}
    We observe that \GMTLin differs from \GSAT just in the number of weighted message passing layers (see Appendix E.1 in \citet{chen2024howinterpretable}). The proof above applies since the underlying training objective remains the same.

    \textbf{\SMGNN \citep{azzolin2025formal}} 
    The sparsity-based learning objective of \SMGNN is:
    \begin{equation}
    \label{eq:smgnn-objective}
        \calL_{cl\!f}(\CLF, \DET, G, y) + \frac{\lambda_1}{|V|}\sum_{u \in V} p_{u} + \frac{\lambda_2}{|V|}\sum_{u \in V} p_{u}\log(p_{u}) + (1-p_{u})\log(1-p_{u})
    \end{equation}
    where $\calL_{cl\!f}$, $V$, $\lambda_1$, $\lambda_2$, and $p_u$ are specular to those for \GSAT and \LRI.

    We now provide a suitable pair of explanation extractor and classifier achieving an optimal value for \cref{eq:smgnn-objective}.
    \begin{align}
    \label{eq:smgnn-deg-proof-loss}
        p_u \defeq \DET(G) =
        \begin{cases}
            1 & \text{if } u = z_0 \land \phi(G)=0 \\
            1 & \text{if } u = z_1 \land \phi(G)=1 \\
            0 & \text{otherwise }
        \end{cases}
        \quad\quad
        \CLF(R) =
        \begin{cases}
            0 & \text{if } z_0 \in R \land z_1 \not \in R\\
            1 & \text{if } z_1 \in R \land z_0 \not \in R\\            
            0.5 & \text{otherwise}
        \end{cases}
    \end{align}

    Then, by plugging \cref{eq:smgnn-deg-proof-loss} inside the training objective, we get that $\calL_{cl\!f}(\CLF, \DET, G, y) = 0$ since the model achieves correct highly-confident predictions, the middle term becomes $\lambda_1 / |V|$ as only one between $z_0$ and $z_1$ gets a score above zero, and the last term becomes zero as relevance scores are binary in $\{0,1\}$.
    Hence, \cref{eq:smgnn-objective} boils down to $\lambda_1 / |V|$.
    By \cref{assump:discrete-extractor}, the only possible smaller assignment of explanation scores would be giving zero score to every node, which, however, violates \cref{assump:nonempty-expl}. Hence, the explanation given by \cref{eq:smgnn-deg-proof-loss} is the smallest possible, and therefore $\lambda_1 / |V|$ is the smallest risk achievable.

    As a remark, note that the classifier in \cref{eq:smgnn-deg-proof-loss} can indeed be learned by a linear layer with sigmoid output, receiving as input an embedding of the form $\vh = [\Ind{z_0 \in R}, \Ind{z_1 \in R}]$ similar to the one-hot encoding defined above for \GSAT, with weights $\vw = [-\alpha, \alpha]$ and $b=0$, for a sufficiently large $\alpha$. Hence, the classifier expressed in \cref{eq:smgnn-deg-proof-loss} is easily realizable by practical GNNs.

    \textbf{\CAL \citep{Sui2022cal}:}
    \CAL aims at learning an explanation $R = \DET(G)$ containing the causal features for the prediction, while delegating spurious features to the complement $C = G \setminus R$, by applying causal interventions.

    For doing that, it uses a shared explanation extractor $\DET$ that predicts node and edge explanation relevance scores, which can be stacked into a node-wise attention matrix $M_X$ and an edge-wise attention matrix $M_A$.
    The original graph $G=(A,X)$ is therefore separated into $R=(A \odot M_A, X \odot M_X)$ and $C=(A \odot 1-M_A, X \odot 1-M_X)$.
    Then, \CAL instantiates three separate classifiers: $\CLF_R$ to predict the label from the causal explanation; $\CLF_C$ to predict the label from the spurious complement; and $\CLF$ to predict the label from the \textit{implicit intervened graph} defined below.
    Note that, in this case, $\CLF$ is a classifier mapping from the global graph readout to labels, i.e., $\CLF: \mathbb{R}^d \mapsto \calY$, with $d$ the size of the embedding.
    Let $\vh_R$ be the graph embedding computed by $\CLF_R$ before the classification head, and $\vh_C$ be the same graph embedding computed by $\CLF_C$.
    Then, \CAL generates \textit{implicit intervened graph} by simulating causal interventions.
    In practice, this is achieved by summing the embedding of $\vh_{R_i}$ for graph $G_i$ with the embedding $\vh_{C_j}$ of graph $G_j$ chosen at random, and by predicting the original label of sample $G_i$ (refer to Section 3.4.3 of \citet{Sui2022cal} for a detailed description).
    The overall objective amounts to:
    \begin{equation}
    \label{eq:cal-objective}
        \calL_{cl\!f}(\CLF_R, \DET, G_i, y_i)
        + \lambda_1 \KL \big (\calU \{1, |\calY|\}, \CLF_C(C_i) \big )
        + \lambda_2 \calL_{interv}(\CLF, \vh_{R_i} + \vh_{C_j}, y_i) 
    \end{equation}
    where $\calL_{interv}$ is the standard cross entropy loss defined for implicit intervened graphs with $\vh_{R_i} + \vh_{C_j}$ the input to $\CLF$, $\lambda_1, \lambda_2$ are hyperparameters controlling the relative strength of the regularizers, and $\calU \{1, |\calY|\}$ is a discrete uniform distribution over class labels.

    Consider now the same construction as for \SMGNN:
    \begin{align}
    \label{eq:cal-deg-proof-loss}
        p_u \defeq \DET(G) =
        \begin{cases}
            1 & \text{if } u = z_0 \land \phi(G)=0 \\
            1 & \text{if } u = z_1 \land \phi(G)=1 \\
            0 & \text{otherwise }
        \end{cases}
        \quad\quad
        \CLF_R(R) =
        \begin{cases}
            0 & \text{if } z_0 \in R \land z_1 \not \in R\\
            1 & \text{if } z_1 \in R \land z_0 \not \in R\\            
            0.5 & \text{otherwise}
        \end{cases}
    \end{align}
    Also, since $\CLF_C$ is trained to output maximally uncertain predictions for any input (second term in \cref{eq:cal-objective}), a possible trivial solution would be having $\vh_{C} = \vec{0}$ for any $C$, with an associated constant classifier outputting $1 / |\calY|$ for each class, i.e., $\CLF_C(C) \defeq 1 / |\calY|$.
    Note that any $\CLF_C$ trained with weight decay is encouraged to converge to this solution, as no term in \cref{eq:cal-objective} is pushing $\vh_{C}$ to learn a meaningful representation.
    Together, these observations imply that both the first and second terms in \cref{eq:cal-objective} are zero.
    Furthermore, since we fixed $\vh_{C} = \vec{0}$, then the last term in \cref{eq:cal-objective} boils down to:
    \begin{equation}
        \lambda_2 \calL_{interv}(\CLF, \vh_{R_i}, y_i)
    \end{equation}
    Therefore, the solution where the classification head $\CLF$ equals that of $\CLF_R$ implies that all three terms are zero.

    \textbf{Final remarks:}
    Note that our examples are provided assuming the classification head of \CLF is a linear layer.
    Nonetheless, our examples generalize to the case where multi-layer perceptions are used, as commonly found in practical implementations of \SEGNNs.
\end{proof}

\subsection{Extended analysis of the Anchor Set}
\label{appx:subgraph-anchor-sets}

{
We provide a generalization of the theoretical analysis in \cref{sec:failure-cases} to the case where the anchor set is defined in terms of subgraphs rather than single nodes, and where each of these subgraphs is not required to appear in \textit{all} graphs.
%
%
With this new setting, we will show that the risk attained by several popular \SEGNNs when selecting uninformative subgraphs can be lower than the risk attained by selecting other, more informative explanations $R_y^*$.
To keep the discussion intuitive -- and in line with previous work studying \SEGNNs from a causal lens \citep{miao2022interpretable, chen2022learning, azzolin2025reconsidering} -- we will discuss the specific case where $R_y^*$ is the ground truth explanation for class $y$.\footnote{Although $R_y^*$ may depend on the specific graph instance, we consider a fixed subgraph for the ease of the argument. This remains general enough to capture most \SEGNN use-cases commonly found in the literature, like motif-based tasks \citep{miao2022interpretable, chen2022learning, azzolin2025reconsidering}.}
Nonetheless, as we will discuss, our insights will equally apply to any other notion of explanation.
Let us now extend the anchor set definition provided in \cref{sec:failure-cases} to the case of subgraphs. 
\begin{definition}[Subgraph anchor set]
\label{def:subgraph-anchor-set}
    Let $\mathcal{D}_{\mathcal{G}\times\mathcal{Y}}$ be the data distribution, and $\calA=\{ \barG_i \}_{i=1}^m$ a set of generic subgraphs $\barG$ that can appear in any graph $G \in \calG$, with $m \ge |\calY|$.
    The partition of $\calA$ into $\calZ'= \{ \{ \barG_i^y\}_{i=1}^{m_y} \}_{y \in \calY}$ with $\sum_y m_y=m$ is called a \textbf{subgraph anchor set} if:
    
    \begin{enumerate}
        \item (\textbf{Per-label coverage})
        For every $(G,y)$ in the support of $\mathcal{D}_{\mathcal{G}\times\mathcal{Y}}$ there exists at least one subgraph $\barG_i^y \in G$.
    
        \item (\textbf{Disjointness of partitions}) The partitions are pairwise disjoint: $\{ \barG_i^y\}_i \cap \{ \barG_i^{y'}\}_i = \emptyset$ for all $y \neq y'$.
    \end{enumerate}    
\end{definition}

%
Note that $\calA$ constitutes an arbitrary subset of possible subgraphs appearing in the dataset, with no additional assumptions on the structure of the graphs. We will use $\calA$ to represent the possible subgraph explanations that are provided by a generic \DET learned on $\mathcal{D}_{\mathcal{G}\times\mathcal{Y}}$.
\cref{def:subgraph-anchor-set} allows for $\calA$ containing subgraphs $\barG_i$ carrying no information about class labels, meaning they are neither ground truth nor spurious correlation.
In fact, anchor sets (as originally defined in \cref{sec:failure-cases}) are a specific case of subgraph anchor sets, where $m=|\calY|$, and where each $\barG_i$ is a single-node subgraph appearing in all graphs.

Importantly, \cref{def:subgraph-anchor-set} generalizes the original definition of node-wise anchor set (c.f. \cref{sec:failure-cases}) in the following ways:
First, it considers subgraphs instead of single nodes.
Second, it relaxes the condition that anchor nodes must be present in all graphs to the case where different subgraphs can appear in different graphs (see \textit{Per-label coverage} in \cref{def:subgraph-anchor-set}).

We now proceed to show that many popular \SEGNNs can achieve a lower risk by picking explanations from the subgraph anchor set rather than the ground truth $R^*_y$.
%
Crucially, this holds even when $\calA$ contains only subgraphs with no class-discriminative power, showing how the degenerate explanations discussed in \cref{sec:failure-cases} can emerge also in the presence of subgraph explanations.

\begin{restatable}{theorem}{degminimalloss2}
\label{thm:deg-suff-condition2}
    Let $\calD_{\calG \times \calY}$ be a data distribution with deterministic ground truth labeling function $\phi: \calG \mapsto \calY$ and ground truth explanation $R_y^*$,
    \DET be a hard explanation extractor, and $\calZ'= \{ \{ \barG_i^y\}_{i} \}_{y \in \calY}$ be a subgraph anchor set.
    If $|R_y^*| \ge \max_{i} |\barG_i^y|$ for all labels $y$, then there exists a \SEGNN $\CLF\circ\DET$ such that the true risk of \GSAT, \LRI, \CAL, \GMTLin and \SMGNN will be lower or equal for
    $$\DET(G) \defeq \underset{\barG_i^{\phi(G)} \in G}{\argmin} |\barG_i^{\phi(G)}| \qquad \text{and} \qquad \CLF(\barG_i^{y}) \defeq y$$
    than
    $$\DET^*(G) \defeq R^*_y \qquad \text{and} \qquad \CLF^*(R^*_y) \defeq y.$$
\end{restatable} 
\begin{proof}
    To keep the proof simple, we will use the same setting of \cref{thm:deg-suff-condition} for a boolean classification task, and reuse some intermediate results.

    \textbf{\GSAT, \GMTLin, \& \LRI:}
    Let us consider the following pair of explanation extractor and classifier:
    %
    %
    \begin{align}
    \label{eq:gsat-proof-det-clf2}
        R \defeq \DET(G) = 
        \begin{cases}
            \underset{\barG_i^{0} \in G}{\argmin} |\barG_i^{0}| & \text{if } \phi(G) = 0 \\
            \underset{\barG_i^{1} \in G}{\argmin} |\barG_i^{1}| & \text{if } \phi(G) = 1 \\
        \end{cases}
        \quad\quad
        \CLF(R) =
        \begin{cases}
            0 & \text{if } R \in \{ \barG_i^0\}_{i}\\
            1 & \text{if } R \in \{ \barG_i^1\}_{i}\\
            0.5 & \text{otherwise}
        \end{cases}
    \end{align}
    In the equation above, $\argmin_{\barG_i^{0} \in G} |\barG_i^{0}|$ means that $\DET$ will select the smallest subgraph $i$ belonging to partition $y=0$ that is present in the subgraph anchor set for the input sample $G$.
    The classifier $\CLF$, instead, is just left with inferring to which partition the explanation belongs, and with outputting the corresponding partition index.
    In addition, to preserve the semantics of \citet{miao2022interpretable} and in accordance with \cref{assump:discrete-extractor}, we assume that each node in $R$ is given a score of $1$, while the rest is given a score of $r$.

    Since \CLF replicates the predictions of the ground truth labeling function $\phi$ -- hence it classifies every sample correctly with maximum confidence by construction -- $\calL(\CLF, \DET, G, y)$ in \cref{eq:gsat-objective} equals zero. 
    Also, since each node with score equal to $r$ yields zero loss, and each node with score equal to $1$ yields a loss value of $\lambda_1 \log(r)^{-1}$ (see proof of \cref{thm:deg-suff-condition}), the true risk is bounded above as follows:
    \begin{equation}
    \label{eq:GSAT-truerisk-up}
        \bbE_{(G,y)\sim \calD_{\calG\times\calY}} 
            \big [ 
                    \lambda_1 |\DET(G)| \log(r)^{-1}
            \big ]
    \le 
        \bbE_{(G,y)\sim \calD_{\calG\times\calY}} 
            \big [ 
                   \max_i \lambda_1 |\barG_i^y| \log(r)^{-1}
            \big ]
    \end{equation}
    By repeating the same analysis for the following ground truth explanation extractor $\DET^*$ and ground truth classifier $\CLF^*$:
    \begin{align}
    \label{eq:gsat-proof-det-clf2}
        R \defeq \DET^*(G) = R^*_y
        \quad\quad
        \CLF^*(R) =
        \begin{cases}
            0 & \text{if } R = R^*_0\\
            1 & \text{if } R = R^*_1\\
            0.5 & \text{otherwise}
        \end{cases},
    \end{align}
    we get that the true risk when selecting the ground truth explanation $R^*_y$ equals:
    \begin{equation}
    \label{eq:GSAT-truerisk-gt}
        \bbE_{(G,y)\sim \calD_{\calG\times\calY}} 
            \big [ 
                    \lambda_1 |R^*_y| \log(r)^{-1}
            \big ]
    \end{equation}
    It follows that, whenever $|R_y^*| \ge \max_{i} |\barG_i^y|$ for all labels $y$, the true risk of \cref{eq:GSAT-truerisk-up} will be smaller or equal to that of \cref{eq:GSAT-truerisk-gt}.

    \textbf{\SMGNN:}
    The proof works similarly to above. Let us consider the following \SEGNN:
    \begin{align}
    \label{eq:smgnn-proof-det-clf2}
        R \defeq \DET(G) = 
        \begin{cases}
            \underset{\barG_i^{0} \in G}{\argmin} |\barG_i^{0}| & \text{if } \phi(G) = 0 \\
            \underset{\barG_i^{1} \in G}{\argmin} |\barG_i^{1}| & \text{if } \phi(G) = 1 \\
        \end{cases}
        \quad\quad
        \CLF(R) =
        \begin{cases}
            0 & \text{if } R \in \{ \barG_i^0\}_{i}\\
            1 & \text{if } R \in \{ \barG_i^1\}_{i}\\
            0.5 & \text{otherwise}
        \end{cases}
    \end{align}

    $\calL(\CLF, \DET, G, y)$ in \cref{eq:smgnn-objective} is again zero for the reasons above, and since \cref{eq:smgnn-objective} penalizes the loss by a factor of $\lambda_1 / |V|$ for each node in $R$, it reduces to $\lambda_1 |\DET(G)| / |V|$ (see proof of \cref{thm:deg-suff-condition}). 
    The true risk is then upper bounded as follows:
    \begin{equation}
    \label{eq:SMGNN-truerisk-up}
        \bbE_{(G,y)\sim \calD_{\calG\times\calY}} 
            \big [ 
                    \lambda_1 \frac{|\DET(G)|}{|V|}
            \big ]
        \le 
        \bbE_{(G,y)\sim \calD_{\calG\times\calY}} 
            \big [ 
                   \max_i \lambda_1 \frac{|\barG_i^y|}{|V|}
            \big ]
    \end{equation}

    By repeating the same analysis for the same ground truth explanation extractor $\DET^*$ and ground truth classifier $\CLF^*$ of \cref{eq:gsat-proof-det-clf2}, the true risk equals:
    \begin{equation}
    \label{eq:SMGNN-truerisk-gt}
        \bbE_{(G,y)\sim \calD_{\calG\times\calY}} 
            \big [ 
                    \lambda_1 \frac{|R^*_y|}{|V|}
            \big ]
    \end{equation}
    It follows that, whenever $|R_y^*| \ge \max_{i} |\barG_i^y|$ for all labels $y$, the true risk of \cref{eq:SMGNN-truerisk-up} will be smaller or equal to that of \cref{eq:SMGNN-truerisk-gt}.

    \textbf{\CAL:}
    Exactly as in the original proof of \cref{thm:deg-suff-condition}, we can set $\vh_C \equiv 0$ and train $g_R$ and $g$ to predict the right label from the explanation only, achieving zero classification and intervention loss in \cref{eq:cal-objective} for both choices of explanation extractor (\DET or $\DET^*$).
    Moreover, we can choose $g_C$ to output the uniform distribution on every input, making the KL term zero as well as in the proof of \cref{thm:deg-suff-condition}. 
    Hence, both \SEGNNs attain the same value of the \CAL objective, independently of $|R|$.
\end{proof}

As a remark, note that \cref{thm:deg-suff-condition2} does not depend on the specific form of $R_y^*$, but relies only on its cardinality $|R_y^*|$.
Therefore, the same result applies to any other notion of explanation $R_y^*$ beyond ground truth explanations, like explanations highlighting spurious correlation \citep{wu2022discovering}, invariant subgraphs \citep{chen2022learning}, or minimal label-preserving subgraphs \citep{azzolin2025formal}.
That is, \SEGNNs in \cref{thm:deg-suff-condition2} can prefer to highlight smaller uninformative explanations over other reasonable notions of explanations.
}

\section{Extended Related Work}
\label{appx:related-work}
\input{extras/extended-related-work}

\section{Implementation details}
\label{appx:impl-details}

We relied on the codebase provided by \citet{gui2022good}, which implements \GSAT and \DIR.
The implementation of \SMGNN is derived from \citet{azzolin2025formal}.
Every model uses \ACR \citep{barcelo2020logical} as backbone, where each operation is modified to incorporate the conditioning on explanation relevance scores as follows:
\begin{equation}
    h_u^i = COM^i \big (
        Upd^i(p_uh_u^{i-1}),
        Aggr(  \{\{ p_vh_v^{i-1} \mid v \in \calN_G(u) \}\}  ),
        Read( \{\{ p_vh_v^{i-1} \mid v \in V \}\} )
    \big )
\end{equation}
where $Read$ and $Aggr$ are defined as the sum, $Upd^i$ and $COM^i$ as a 3-layer MLP for \MNIST and \MUTAG, and as a 2-layer MLP for other datasets.
This choice of model backbone was driven by the fact that \ACR has a precise mapping to the range of tasks it can solve \citep{barcelo2020logical}, and therefore guides the range of tasks for which the explanation extractor can meet the conditions highlighted in \cref{thm:deg-suff-condition}, which nonetheless remains valid for any backbone.
In the \SEGNN's classifier, each MLP has the bias term deactivated, discouraging the emergence of \textit{default} predictions \citep{faber2021comparing}.
Also, the final graph global readout implements \textit{Explanation Readout} \citep{azzolin2025reconsidering}, i.e., it weights the final node embeddings based based on their relevance score.
Further details about hyperparameters and the dataset are provided below.

\subsection{Implementation details of \cref{sec:segnns-can-be-manipulated}}
\label{appx:details-attack}

\subsubsection{Natural training}
\label{appx:details-attack-natural}
When training models without malicious attacks, we stick to the original training procedure of each \SEGNN.
In particular, we fix the hyperparameters as follows:

\textbf{\BAColorGV:}
Each model is trained with a batch size of $64$ and for a number of epochs set to $200$ with a learning rate of $0.0001$ and without weight decay or dropout.
Also, we fix the global readout to sum, and use 2 layers for both the explanation extractor and classifier. The embedding dimension is set to $100$.
For \GSAT, we set $r=0.3$ and $\lambda=0.1$.
For \SMGNN, we set $\lambda_{spars}=0.4$, $\lambda_{entr}=0.1$.
For \DIR, we set K to $0.5$ and $\lambda=10$.

\textbf{\MNIST:}
Each model is trained with a batch size of $256$ and for a number of epochs set to $200$ with a learning rate of $0.001$ and without weight decay or dropout.
Also, we fix the global readout to sum, and use 2 layers for both the explanation extractor and classifier. The embedding dimension is set to $300$.
For \GSAT, we set $r=0.7$ and $\lambda=0.1$.
For \SMGNN, we set $\lambda_{spars}=0.01$, $\lambda_{entr}=0.015$.
For \DIR, we set K to $0.8$ and $\lambda=0.0001$.
\SMGNN and \DIR further adopt Batch Normalization across GNN layers \citep{ioffe2015batch}.

\textbf{\MUTAG:}
Each model is trained with a batch size of $64$ and for a number of epochs set to $100$ with a learning rate of $0.001$ and without weight decay or dropout.
Also, we fix the global readout to sum, and use 2 layers for both the explanation extractor and classifier. The embedding dimension is set to $64$.
For \GSAT, we set $r=0.7$ and $\lambda=1$.
For \SMGNN, we set $\lambda_{spars}=0.001$, $\lambda_{entr}=0.008$.
For \DIR, we set K to $0.5$ and $\lambda=0.0001$.

\textbf{\SSTP:}
Each model is trained with a batch size of $256$ and for a number of epochs set to $100$ with a learning rate of $0.001$ and without weight decay. The dropout ratio is set to $0.3$.
Also, we fix the global readout to mean, and use 2 layers for both the explanation extractor and classifier. The embedding dimension is set to $64$.
For \GSAT, we set $r=0.7$ and $\lambda=1$.
For \SMGNN, we set $\lambda_{spars}=0.1$, $\lambda_{entr}=0.1$.
For \DIR, we set K equals $0.6$ and $\lambda=10$. In this particular case, we slightly changed the learning rate to $0.0002$ and the training batch size to $128$, which we found helped with convergence.
\SMGNN further employs Batch Normalization across GNN layers \citep{ioffe2015batch}.

\subsubsection{Malicious training}
\label{appx:details-attack-malicious}

Since we aim to maliciously attack the model, we assume full control over the model's training.
While this is a limitation of our attack, it fits the scenario in which an external service provider want to conceal the use of protected attributes, highlight a risk in the trustworthiness of the explanations they provide.
Regardless, we show in \cref{sec:naturaldeg} that explanations similar to those extracted in \cref{sec:segnns-can-be-manipulated} also appear in natural scenarios, further motivating the study of this weakeness.
When training models under the attack presented in \cref{sec:segnns-can-be-manipulated}, the objective becomes the one illustrated in \cref{eq:loss-attack}.
Other explanation-regularization losses are deactivated to avoid conflict with the explanation-supervision loss of \cref{eq:loss-attack}.
The other hyperparameters, including the training batch size, remain unchanged with respect to the ones reported above.

Since the supervision provided by \cref{eq:loss-attack} can be highly class-unbalanced, with many more negatives than positives, we manually reweigh by $10$ the contribution from positive nodes for \BAColorGV, and by $100$ for other datasets.
Then, \SEGNNs are trained for a maximum of $1500$ epochs, and we define the stopping criterion as a condition on the \SEGNN's loss and accuracy on the training set, as follows:

For \DIR, we set the threshold on minimum accuracy as $98$ for \MNIST and \CPatchMNIST, $95$ for \MUTAG, and $99$ for \SSTP and \BAColorGV. The threshold on minimum classification loss is instead set as $0.01$ \MNIST, \CPatchMNIST, and 
\BAColorGV, $0.08$ for \MUTAG, and 0.015 for \SSTP.
For \GSAT and \SMGNN, instead, the threshold on minimum accuracy is set to $95$ for \MNIST, \CPatchMNIST, and \MUTAG, $96$ for \SSTP, and $99$ for \BAColorGV. The threshold on minimum classification loss is instead set to $0.08$ \MNIST, \CPatchMNIST, and \MUTAG, $0.015$ for \SSTP, and to $0.01$ for \BAColorGV.

Those values were manually defined after inspecting explanations on the validation set.
The final model is picked as the last checkpoint when training is stopped.


\subsection{Implementation details of \cref{sec:faithfulness-metrics}}
\label{appx:details-faithfulness-metrics}

The rejection ratio computed in \cref{eq:rejection-ratio} provides a more interpretable way to judge the faithfulness of explanations.
Typical metrics, in fact, rely on the average change in class probability, which yields scores whose upper-bound depends on the underlying distance function $d$ used, like Total variation or Kullback–Leibler.
Some metrics further non-linearly normalize those values to squash them in the fixed $[0,1]$ range, introducing further sources of variation \citep{azzolin2025reconsidering}.
This makes it harder to find a precise rejection criterion separating \textit{good} from \textit{bad} metric values, as it requires defining a threshold on the expected change.
To avoid this issue, we opt for the max aggregation in \cref{eq:rejection-ratio}, providing a best-case evaluation.

At a high level, \cref{eq:rejection-ratio} computes the fraction of input graphs whose prediction can be changed after applying the perturbations induced by each faithfulness metric, i.e., according to $\calI$.
To allow for some slack, we consider a perturbation as prediction-changing if it makes the classifier’s output highly uncertain, \eg if the predicted probability lies in $[0.4,0.6]$ for binary classification.

\subsection{Implementation details of \cref{sec:naturaldeg}}
\label{appx:details-naturaldeg}

In \cref{sec:naturaldeg} we provide a similar natural training of \SEGNNs as \cref{sec:segnns-can-be-manipulated}, with the only difference that hyperparameters are optimized to increase the sparsity of explanations, while making sure that the final accuracy is not significantly penalized.
All the other aspects of model training are kept unaltered.
We detail below the chosen hyperparameters for this analysis, reporting only those that differ from \cref{appx:details-attack-natural}.

\textbf{\BAColorGV:}
\SMGNN is trained with $\lambda_{spars}=0.4$ and $\lambda_{entr}=1.0$, and with Batch Normalization across GNN layers.
\GSAT is kept the same as in \cref{appx:details-attack-natural}, while \DIR uses a topK selection with K equals $1\%$ as discussed in \cref{sec:naturaldeg}.
This particularly low value is set on purpose and to verify that \DIR can learn to squash all the relevant information inside a small, potentially unrelated, subgraph.

\textbf{\MNIST:}
\SMGNN is trained with $\lambda_{spars}=1$ and $\lambda_{entr}=1.5$.
\GSAT is kept the same as in \cref{appx:details-attack-natural}, while \DIR uses a topK selection with K equals $10\%$.
We note that \DIR reportedly performs poorly on the \MNIST dataset \citep{wu2022discovering}, and even using larger K does not yield good results.

\textbf{\MUTAG:}
\SMGNN is trained with $\lambda_{spars}=0.1$, $\lambda_{entr}=0.8$.
\GSAT is kept the same as in \cref{appx:details-attack-natural}, while \DIR uses a topK selection with K equals $10\%$.
This is the only case where setting a significantly lower K alters performance compared to a larger K.

\textbf{\SSTP:}
\GSAT and \SMGNN are kept the same as in \cref{appx:details-attack-natural}, while \DIR uses a topK selection with K equals $10\%$.
We noticed that \SMGNN trained for random seed number $5$ yields all scores squashed to $0$, and we removed it from our analysis.
Similarly, random seed $3$ for \DIR yielded significantly lower accuracy, and we also removed it.

\subsection{Datasets}
\label{appx:datasets}

We considered the following datasets in our analysis:
\begin{itemize}[leftmargin=1.25em]

    \item \BAColorGV (ours).
    Nodes are colored with a one-hot encoding of either red, blue, green, or violet.
    The task to predict is whether the number of blue nodes is larger than the number of red ones.
    The topology is randomly generated from a Barab{\'a}si-Albert distribution \cite{barabasi1999emergence}, and we ensure that green and violet nodes are always disconnected. 
    Each graph contains a total number of red and blue in the range $[0, 100]$, plus a single green and a single violet node.
    The total dataset size is $5000$ graphs, divided with an 80/10/10 split into train, validation, and test sets, respectively.

    \item \MUTAG \citep{debnath1991structure}, abbreviation for Mutagenicity, is a molecular property prediction dataset, where each molecule is annotated based on its mutagenic effect. The nodes represent atoms and the edges represent chemical bonds. The dataset is composed of 4337 graphs, where negative labels indicate mutagenic molecules and positive non-mutagenic ones.
    
    \item \MNIST \citep{knyazev2019understanding} converts the popular MNIST image-based digit recognition dataset \citep{Lecun1998mnist} into graphs by applying a super pixelation algorithm. Nodes are then composed of superpixels, while edges follow the spatial connectivity of those superpixels.

    \item \CPatchMNIST (ours) is an extension of \MNIST where we color the top-left and bottom-right pixels with a specific color that is indicative of the final label, as detailed in \cref{tab:cpatchmnist-encoding}. This experiment is detailed in \cref{app:plausible-but-unfaithful-experiment}. Visual examples are shown in \cref{fig:cpatchmnist-swap-colors}.
    
    \item \SST is a sentiment analysis dataset, from \citet{yuan2022explainability}. The task requires predicting the sentiment polarity of tweets, which are either labeled as negative or positive. Node features are contextual embeddings from a pretrained language model \citep{yuan2022explainability}. 

    \item \SSTP (ours) is an extension of \SSTP, where we force each input sample to contain a single "," and a single "." in the input, such that they are not informative of the final label.
    To make sure that "," and "." are not correlated with the label, we add them as isolated nodes in the graph.
    If the graph originally contained such a punctuation, we first remove it.
    To further avoid the contextual embedding from carrying over some information regarding emotion-laden words, we use fixed embeddings for  "," and ".", respectively set to $\vec{1}$ and $\vec{0}$.
\end{itemize}

\subsubsection{Designated explanations for \cref{sec:segnns-can-be-manipulated}}
\label{appx:designated-explanations-details}

In our experiments, we define the following designated explanations $p_u^{y}$:
For \MNIST, these explanations highlight background pixels, similarly to \citet{jethani2021have};
For \MUTAG, these highlight hydrogen (H) and carbon (C) atoms, and plot in \cref{fig:hist-MUTAG} the relative frequencies of these atoms, showing they are not informative of the label;
For \SSTP, the designated explanations highlight ",'' and ".'', similarly to \citet{yu2019rethinking};
For \BAColorGV, instead, these highlight green and violet nodes, which are clearly uninformative of the underlying task to predict.
We report in \cref{tab:induced-explanations} the mapping between classes and designated explanations.

\begin{table}[!h]
  \centering
  \caption{Designated explanations are defined as nodes that are unrelated to the task being solved. Yet, in \cref{sec:segnns-can-be-manipulated} we show that \SEGNNs can extract these explanations while accurately solving the task. "H'' and "C'' stand for hydrogen and carbon atoms, respectively.}
  \label{tab:induced-explanations}
  \scalebox{.95}{
      \begin{tabular}{lll}
        \toprule
            \textbf{Dataset} & \makecell[c]{\textbf{Task} \textbf{to predict}} & \makecell[c]{\textbf{Designated Expl. $p_u^{y}$}}\\
        \midrule
            \BAColorGV & \makecell[l]{If blue $>$ red\\nodes} & \makecell[l]{Green node for $y=1$,\\ violet node for $y=0$}\\
            \MNIST & \makecell[l]{Digit number} & \makecell[l]{Top-left and bottom-\\right $y$-th pixel}\\
            \MUTAG & \makecell[l]{Mutagenicity\\of molecules} & \makecell[l]{H atoms for $y=0$, C\\atoms for $y=1$}\\
            \SSTP & \makecell[l]{Tweet sentiment\\polarity} & \makecell[l]{',' for $y=0$, '.' for $y=1$}\\
        \bottomrule
      \end{tabular}
  }
\end{table}

\subsection{Thresholding explanations}
\label{appx:thresholding-explanations}

As discussed in \cref{sec:background}, \SEGNNs output relevance scores in the interval $[0,1]$.
%
%
Then, to extract a discrete subgraph that users can consume, we need to define a selection criterion to decide which nodes are to be considered part of the explanation, and which are not.
Following \citet{tai2025redundancy}, and in accordance with the semantics each \SEGNN gives to explanation relevance scores it predicts, we define some threshold values for \SMGNN and \GSAT, as follows.

\textbf{For naturally trained models}, we fix the threshold to $0.5$ for \SMGNN as \citet{tai2025redundancy}, whereas for \GSAT we set it to $0.9$. This is done because \GSAT's relevance scores are associated with the probability of sampling each node/edge, hence we consider as relevant only those that can be consistently sampled with high probability.
The only exception is \GSAT trained for \BAColorGV, where the threshold is again set to $0.5$. This is because for \BAColorGV, \GSAT is trained to push the sampling probability around $0.3$ (by setting $r=0.3$), and therefore a probability of $0.5$ is already consistently higher than $0.3$.
Occasionally, \SMGNN's relevance scores all collapse to values around $0$, or to values such that no node has a value greater than $0.5$. To avoid manually defining case-by-case thresholds when this happens, we apply an instance-wise min-max normalization while sticking to the $0.5$ threshold. In particular, this happened for \MNIST and \MUTAG.
For \DIR, instead, we keep the topK selection used during training.

\textbf{For attacked models}, the threshold is fixed to $0.5$ as \SEGNNs are explicitly trained to extract binary relevance scores.
This also applies to \DIR, for which we remove nodes inside the topK subgraph having a score less than the threshold.
This is because nodes with a score close to $0$ should still be considered irrelevant, even if in the topK.

\subsection{Faithfulness metrics}
\label{appx:detail-metrics}

Each faithfulness metric defines a set of allowed perturbations indicated with $\calI$.
Perturbations are either applied to the complement -- for \textit{sufficiency} -- or to the explanation -- for \textit{necessity} metrics.
Below, we detail the implementation details for the metrics we empirically tested in \cref{sec:faithfulness-metrics}.
In all cases, the explanation selection mechanism is fixed and detailed in \cref{appx:thresholding-explanations}.

\paragraph{\FIDM:}
$\calI$ is simply defined as the perturbation removing the complement of the explanation from the graph, and feeding the explanation alone back to the model.

\paragraph{\FIDP:}
$\calI$ is simply defined as the perturbation removing the explanation from the graph, and feeding the complement of the explanation alone back to the model.

\paragraph{\RFIDM:}
$\calI$ is defined as random edge removals from the complement, where each edge is removed independently from the others and with a fixed probability. Following \citet{zheng2023robust}, we fix the probability to $0.9$.
Nodes are left untouched.

\paragraph{\RFIDP:}
$\calI$ is defined as random edge removals from the explanation, where each edge is removed independently from the others and with a fixed probability. Following \citet{zheng2023robust}, we fix the probability to $0.1$. Nodes are left untouched.

\paragraph{\SUF:}
$\calI$ is defined as complement swaps. In particular, given the reference graph for which we compute the metric, we pick a random sample from the same split and with the same label, and we swap their complements. The explanation is randomly attached to the new complement while preserving the total number of edges of the original graph.

\paragraph{\NEC:}
$\calI$ is defined as random edge removals from the explanation, where each edge is removed independently from the others, and the overall budget of removal is defined dataset-wise and kept fixed for each sample. Following \citet{azzolin2025reconsidering}, we fix the budget as $10\%$ of the average graph size of the split of data where the metric is computed. Nodes are left untouched.

\paragraph{\COUNTERFID:}
$\calI$ perturbs only the relevance scores, where the perturbation is randomly sampled from a Gaussian distribution with mean and variance estimated instance-wise, and applied to scores inside and outside of the explanation. Therefore, the topology of the perturbed graph is the same as the original one, with perturbed relevance scores. The perturbed graph is then fed to the \SEGNN's classifier \CLF only. The expected semantics is that the classifier should be highly responsive to perturbations. Hence, large changes in the output should be expected for faithful explanations.

\paragraph{\SUFFCAUSE:}
$\calI$ is defined as any possible supergraph of the explanation inside the input graph. Equivalently, given the input graph and its explanations, $\calI$ randomly samples perturbations removing nodes and edges, jointly. This can be seen as a more general version of \RFIDM, where nodes can also be removed.

\begin{algorithm}[!h]
\caption{Uniform Sufficiency Test (\SUFFCAUSE, \cref{def:suff-cause})}
\label{alg:suffcause}
    \begingroup
        \begin{algorithmic}[1]
        \REQUIRE Input graph $G$, explanation $R$, budget of perturbations $b$, distance function $d$

        $\mathcal{S} = []$
        \FOR{$i = 1$ to $b$}
            
            \STATE \texttt{// Subsample nodes}
            
            \STATE Sample random node weights $\mathbf{w} \sim \mathrm{Unif}(0,1)^{|V(G)|}$
            \STATE Force $\mathbf{w}[v] \gets 1$ for all $v \in V(R)$
            \STATE Define node set $V_i = \{ v \in V(G) \mid \mathbf{w}[v] \geq 0.5 \}$
            \STATE Construct node-induced subgraph $G_i = G[V_i]$

            \STATE
            \STATE \texttt{// Subsample edges}
            
            \STATE Sample random edge weights $\mathbf{w} \sim \mathrm{Unif}(0,1)^{|E(G_i)|}$
            \STATE Force $\mathbf{w}[e] \gets 1$ for all $e \in E(R)$
            \STATE Define edge set $E_i = \{ e \in E(G_i) \mid \mathbf{w}[e] \geq 0.5 \}$
            \STATE Construct edge-induced subgraph $G'_i = G_i[E_i]$
            
            \STATE Append $G'_i$ to $\mathcal{S}$
        \ENDFOR
        \STATE \textbf{return} $\max_{G' \in \mathcal{S}} d\!\left(\MONO(G), \MONO(G')\right)$
        \end{algorithmic}
    \endgroup
\end{algorithm}

\textbf{Remark:}
Contrary to previous faithfulness metrics, which typically average the contribution from multiple perturbations, \SUFFCAUSE adopts a worst-case evaluation by taking the maximum change in prediction across perturbations.
This stops frequent but irrelevant perturbations from diluting rare but impactful perturbations.

\section{Additional Experiments}

\input{tables/rejections-metrics}

\begin{figure}[t]
    \centering
    \includegraphics[width=0.95\linewidth]{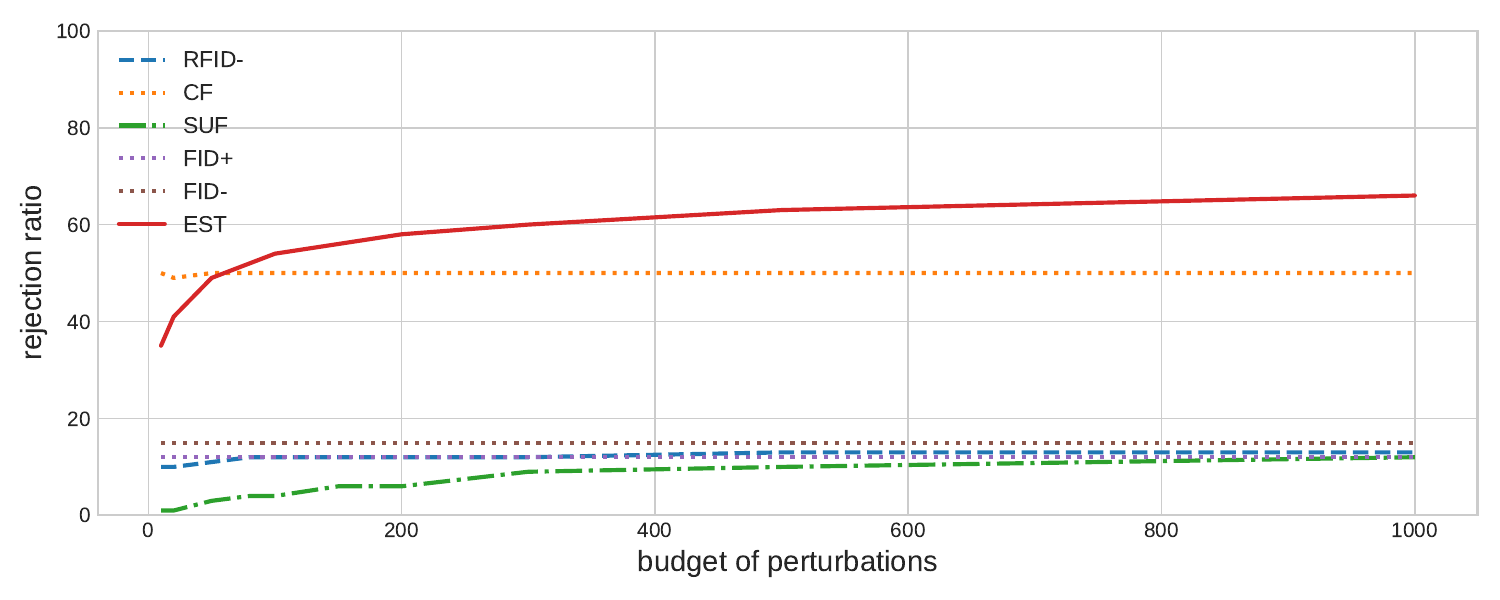}
    \caption{
        Ablation study on the budget of perturbations to estimate \cref{eq:rejection-ratio} for \GSAT on \BAColorGV (cf. \cref{sec:proposing-new-metric}).
        The plot shows that \SUFFCAUSE steadily rejects more unfaithful explanations as the budget increases, while other metrics do not substantially improve.
        \FIDM (\FIDP) have constant values, a they do not apply any sampling but just erase the complement (the explanation) once.
        Note that \SUFFCAUSE samples subgraphs of the complement at random, and thus, the probability of sampling duplicates increases as the budget increases.
        Future work can investigate more complex sampling strategies to avoid such duplicates.
    }
    \label{fig:budget-ablation}
\end{figure}

\subsection{Plausible but unfaithful explanations}
\label{app:plausible-but-unfaithful-experiment}

\begin{table}[!h]
  \centering
  \scalebox{.75}{
      \begin{tabular}{cc}
            \toprule
             \textbf{Color} & \textbf{Class label}\\
            \midrule
             \textcolor{cpatchmnistzero}{\circledrgb{4pt}}   & 0 \\
             \textcolor{cpatchmnistone}{\circledrgb{4pt}}   & 1 \\
             \textcolor{cpatchmnisttwo}{\circledrgb{4pt}}   & 2 \\
             \textcolor{cpatchmnistthree}{\circledrgb{4pt}}  & 3 \\
             \textcolor{cpatchmnistfour}{\circledrgb{4pt}}  & 4 \\
             \textcolor{cpatchmnistfive}{\circledrgb{4pt}}  & 5 \\
             \textcolor{cpatchmnistsix}{\circledrgb{4pt}}  & 6 \\
             \textcolor{cpatchmnistseven}{\circledrgb{4pt}}  & 7 \\
             \textcolor{cpatchmnisteight}{\circledrgb{4pt}}  & 8 \\
             \textcolor{cpatchmnistnine}{\circledrgb{4pt}}  & 9 \\
            \bottomrule
        \end{tabular}
  }
  \caption{Mapping between colors applied to \CPatchMNIST and class label.}
  \label{tab:cpatchmnist-encoding}
\end{table}

In this experiment, we aim to show that it is possible to induce \SEGNNs to output highly plausible but unfaithful explanations.
We achieve this by attacking the model following the setting outlined in \cref{sec:segnns-can-be-manipulated}, i.e., adding an explicit supervisory signal to control the explanations provided by the explanation extractor.
In this case, we supervise the \SEGNN to output only nodes belonging to the digit together with its 1-hop neighborhood, which is typically considered to be the ground truth explanation for \MNIST \citep{miao2022interpretable}.
To simulate the presence of protected attributes, we color the top-left and bottom-right pixels with a specific color that is indicative of the final label, as detailed in \cref{tab:cpatchmnist-encoding}.
The resulting dataset will be referred to as \CPatchMNIST.
Therefore, for the attack to be successful, we expect the explanation to highlight only nodes pertaining to the digit itself while concealing the use of the protected attribute.

\begin{figure}[!h]
    \centering
    \includegraphics[width=0.45\linewidth]{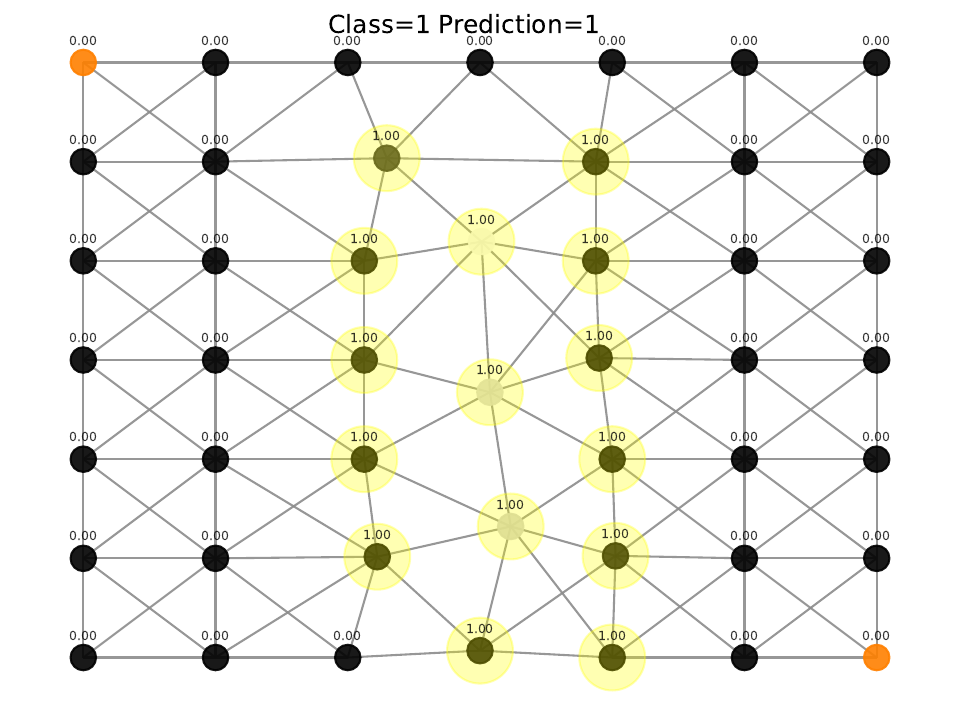}
    \includegraphics[width=0.45\linewidth]{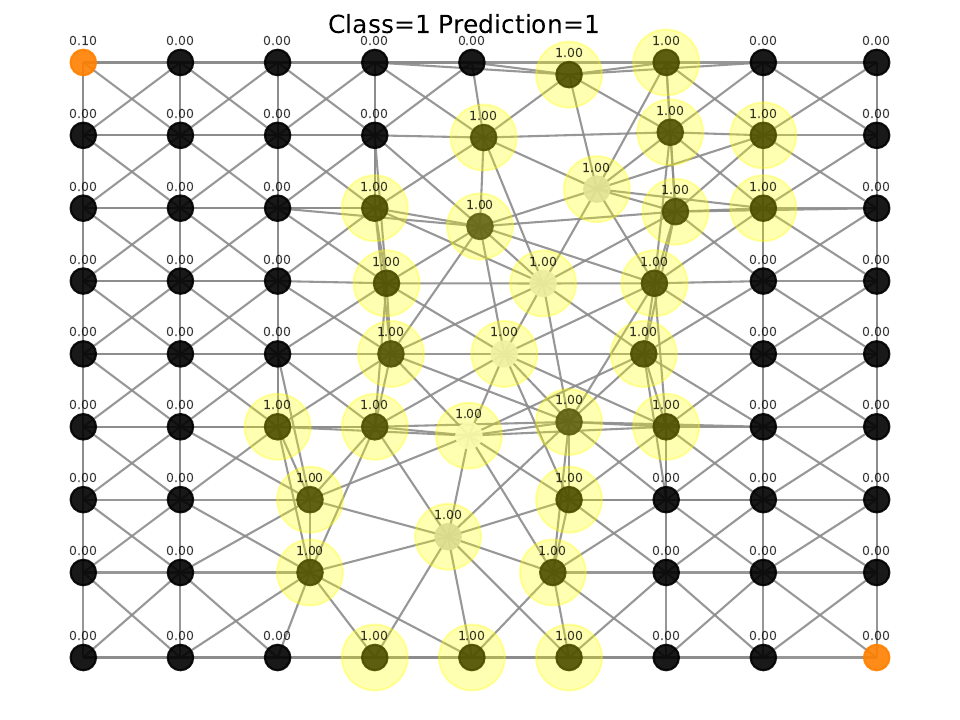}
    \caption{
    \textbf{\SEGNNs can be manipulated to output unfaithful but highly plausible explanations.}
    We train \GSAT (left) and \SMGNN (right) to maximize the downstream accuracy while also optimizing for plausibility, that is, outputting the digit and its 1-hop neighbor as an explanation.
    Nonetheless, both models are heavily relying on the color information, as shown by the severe drop in accuracy when the color information is altered (shown in \cref{tab:cpatchmnist-black-out}), and by the arbitrary manipulation of the predicted label by swapping colors (shown in \cref{fig:cpatchmnist-swap-colors}).
    The underlying dataset is an extension of \MNIST, called \CPatchMNIST, in which we color the upper-right and bottom-left superpixels with class-discriminative colors that are treated as protected attributes. Details about the dataset and the color coding are in \cref{app:plausible-but-unfaithful-experiment}.
    The plot shows two graph examples of class $1$, where nodes belonging to the explanation are selected based on a threshold value on the relevance score at $0.5$. 
    Numbers above each node represent raw explanation relevance scores.
    Better seen in digital format.
    }
    \label{fig:cpatchmnist}
\end{figure}

\begin{figure}[!h]
    \centering
    \includegraphics[width=0.4\linewidth]{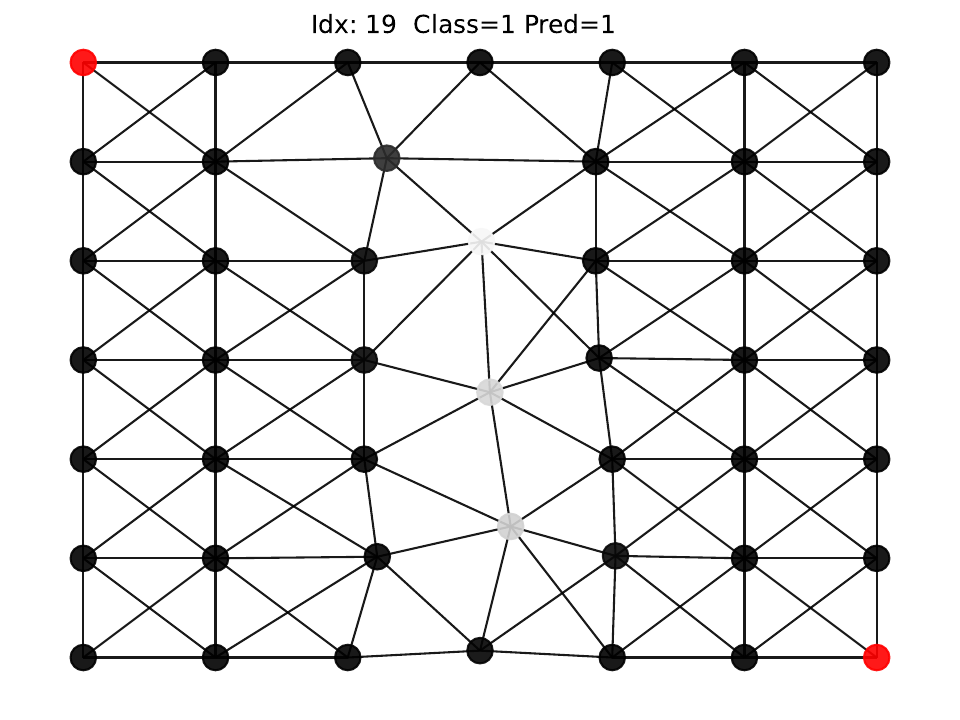}
    \includegraphics[width=0.4\linewidth]{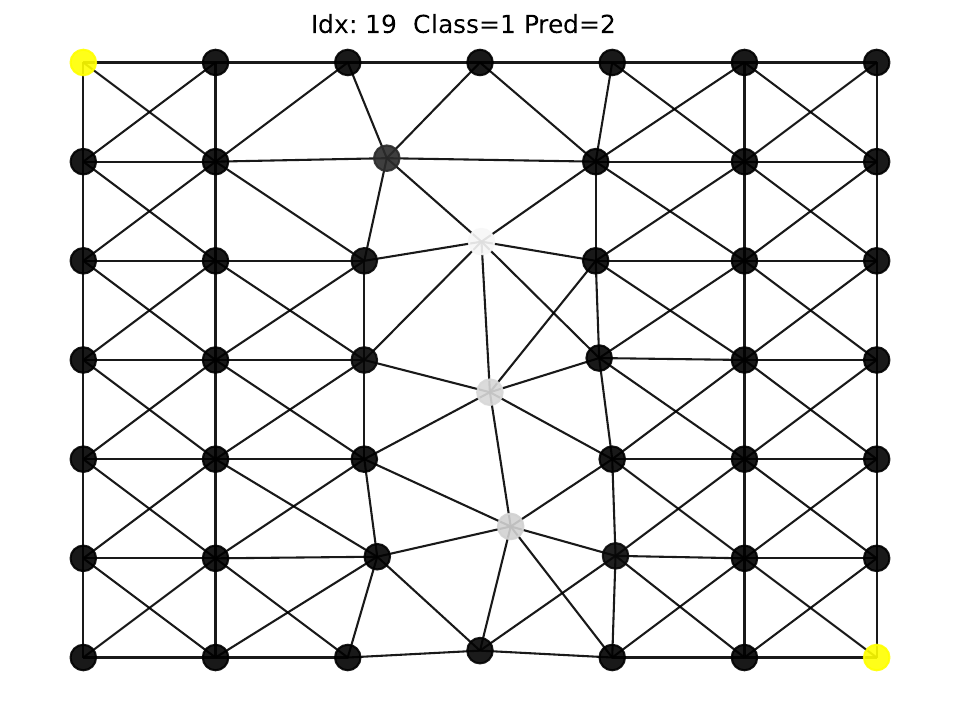}
    \includegraphics[width=0.4\linewidth]{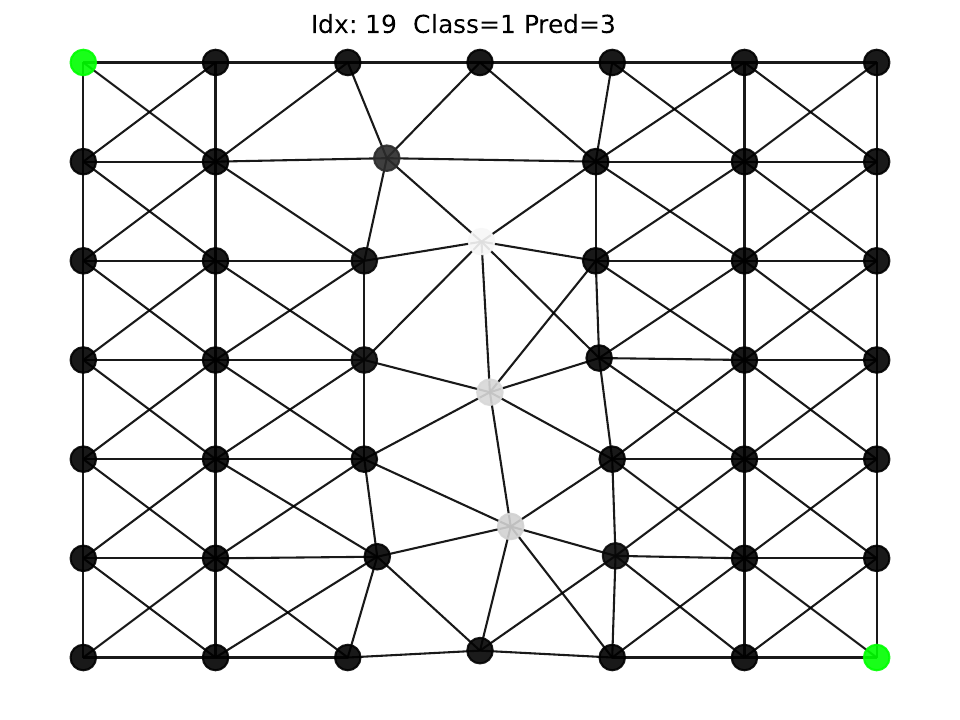}
    \includegraphics[width=0.4\linewidth]{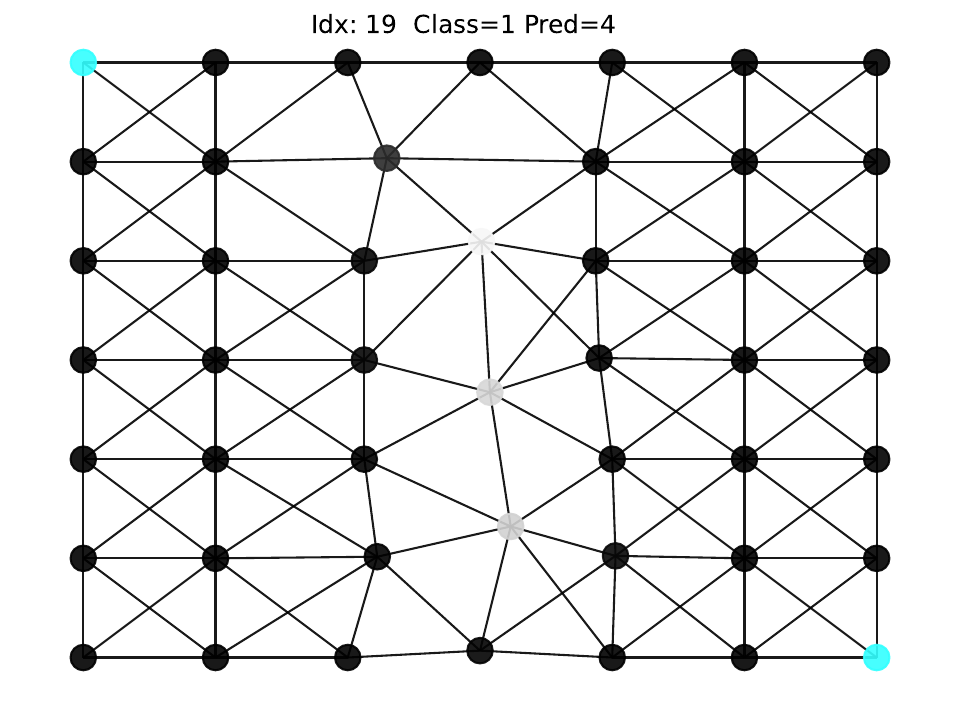}
    \includegraphics[width=0.4\linewidth]{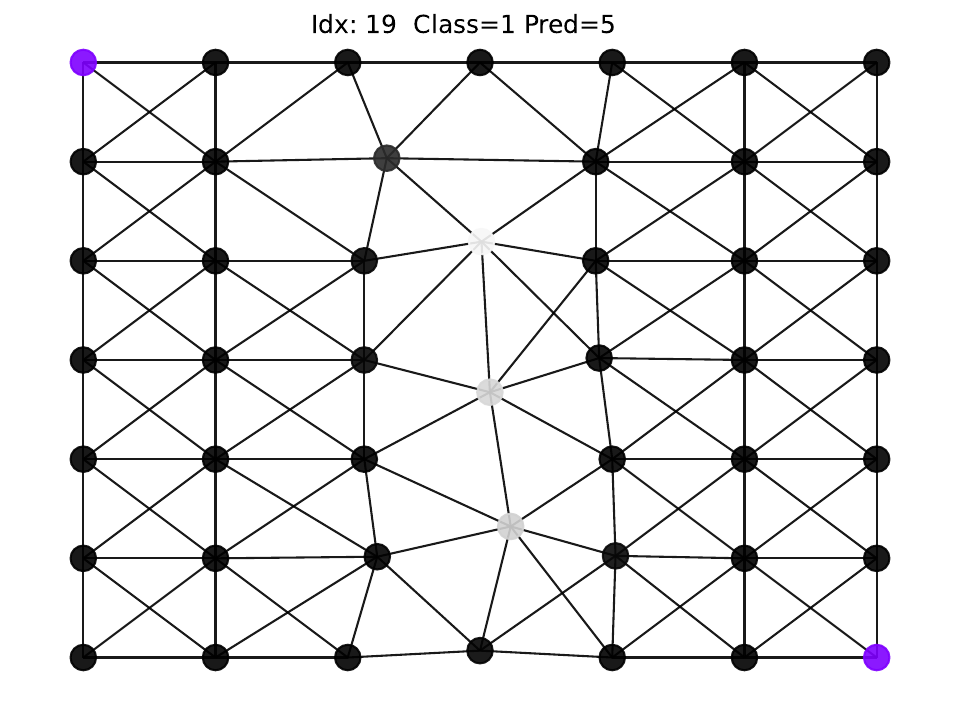}
    \includegraphics[width=0.4\linewidth]{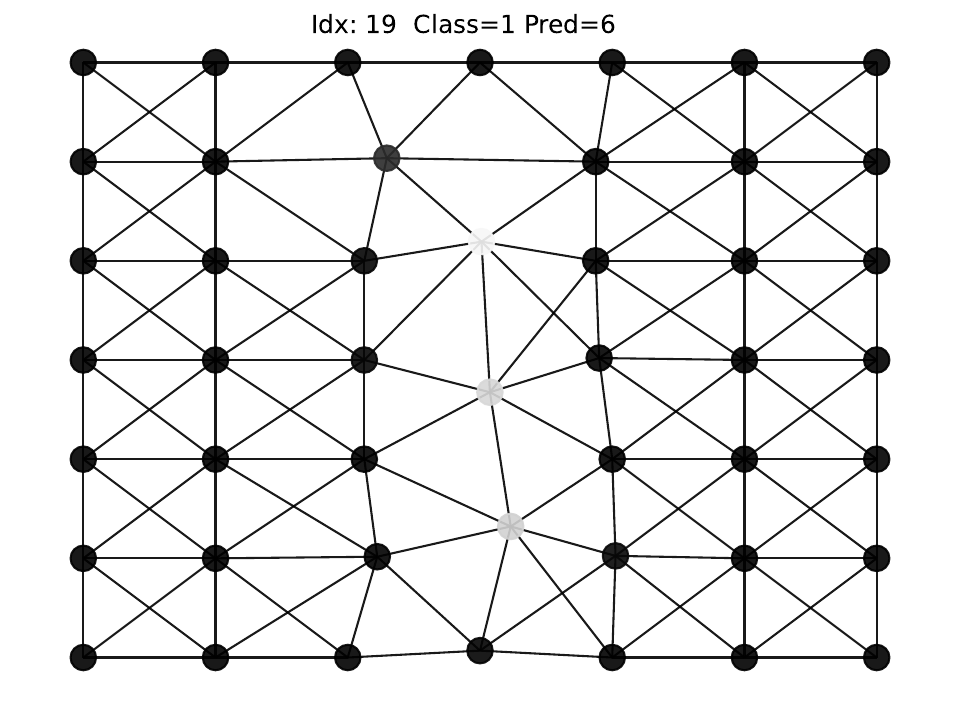}
    \includegraphics[width=0.4\linewidth]{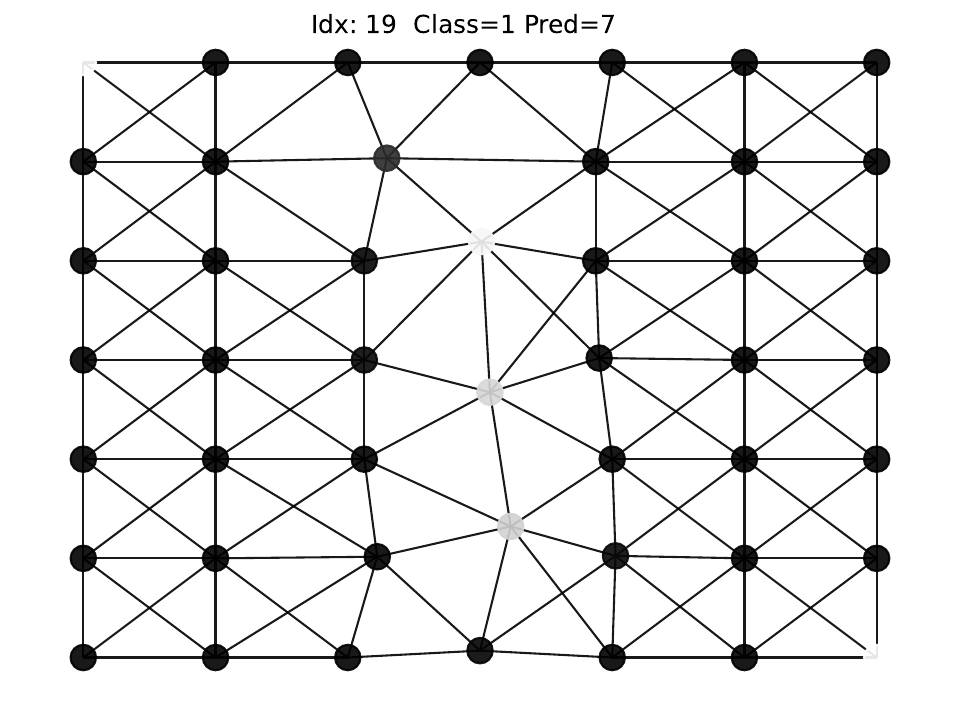}
    \includegraphics[width=0.4\linewidth]{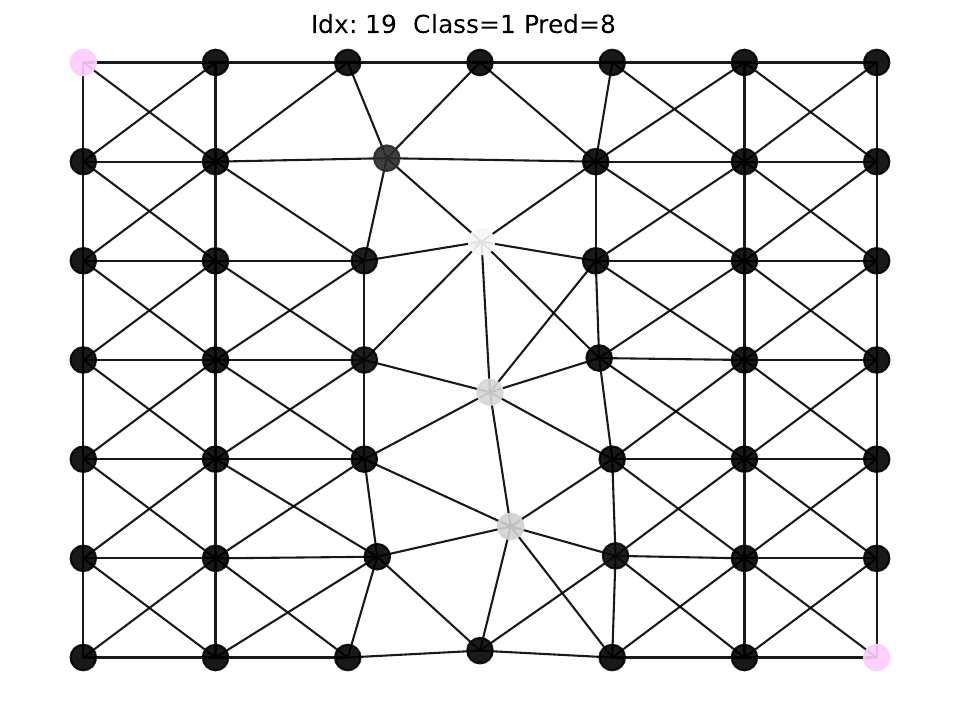}
    \includegraphics[width=0.4\linewidth]{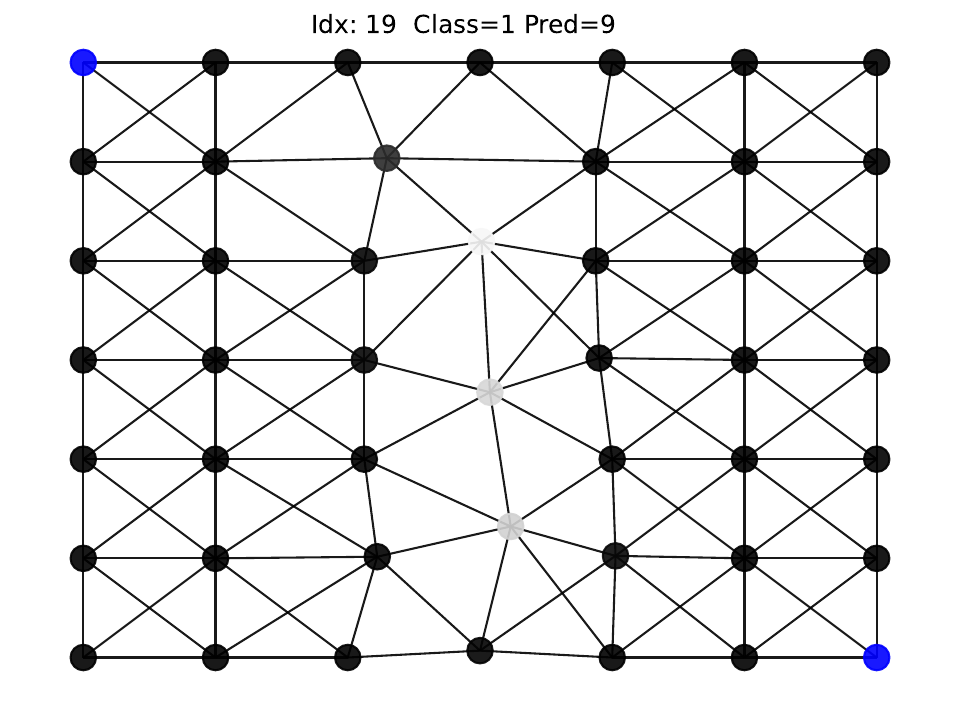}
    \caption{
        Despite \GSAT's explanation declaring that only the digit information is used for the graph in \cref{fig:cpatchmnist}, swapping the color to that of other classes, following the schema of \cref{tab:cpatchmnist-encoding}, results in arbitrary manipulation of the final prediction.
    }
    \label{fig:cpatchmnist-swap-colors}
\end{figure}

We report in \cref{tab:cpatchmnist-black-out} the results of the attack, which confirm that it is possible to train \SEGNNs to output highly human-desirable explanations -- in particular those matching the human expectations -- whilst the model is also relying crucially on other -- potentially protected -- features.
We provide some visual examples of the attack in \cref{fig:cpatchmnist}, and the predicted label for graphs with swapped colors in \cref{fig:cpatchmnist-swap-colors}.
\begin{table}[!h]
    \centering
    \caption{Accuracy and test accuracy after either swapping the color of the colored pixels with that of other classes (see \cref{fig:cpatchmnist-swap-colors}), or blacking out the digit, for the experiment in \cref{app:plausible-but-unfaithful-experiment}.
    Results show that both models are highly dependent on both the digit-related superpixels and the color ones, despite only the former being declared in the explanation.}
    \label{tab:cpatchmnist-black-out}
    \scalebox{0.90}{
        \begin{tabular}{ccccc}
            \toprule
             \textbf{Model} & \textbf{Val Acc} & \textbf{Val $F_1$} & \textbf{Acc color swap} & \textbf{Acc no digit}\\
            \midrule
             \SMGNN  & \nentry{100.0}{0.0} & \nentry{99.3}{0.3} & \nentry{0.1}{0.2} & \nentry{9.6}{0.0}\\
             \GSAT   & \nentry{99.9}{0.1}  & \nentry{99.1}{0.1}  & \nentry{3.4}{4.2} & \nentry{8.7}{1.9}\\
            \bottomrule
        \end{tabular}
    }
\end{table}

\textbf{Can faithfulness metrics spot plausible but unfaithful explanations?}
Similar to the analysis presented in \cref{sec:empirical-benchmark}, we investigate whether popular faithfulness metrics can mark these explanations as unfaithful.
We adopt the same setting as \cref{sec:empirical-benchmark}, and compute rejection ratios for our proposed \SUFFCAUSE, and for \FIDM, and \RFIDM as baselines.
The results in \cref{tab:rejection-cpatchmnist} show that \RFIDM can catastrophically fail to mark these explanations as unfaithful.
Conversely, \FIDM and \SUFFCAUSE can robustly reject a large number of explanations, albeit with consistent variations across random seeds.
Overall, \SUFFCAUSE achieves the highest ratios in both experiments.
\input{tables/rejections-capatchmnist}

\subsection{Examples of Attacked \SEGNNs}
\label{appx:examples-attacks}

\begin{figure}[!h]
    \centering
    \includegraphics[width=0.45\linewidth]{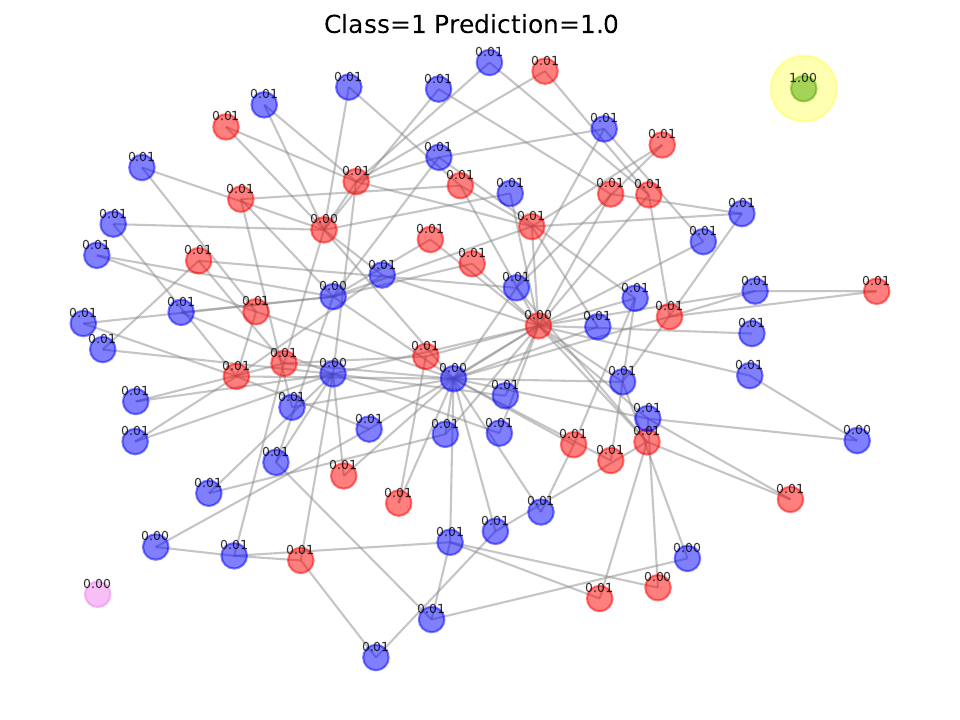}
    \includegraphics[width=0.45\linewidth]{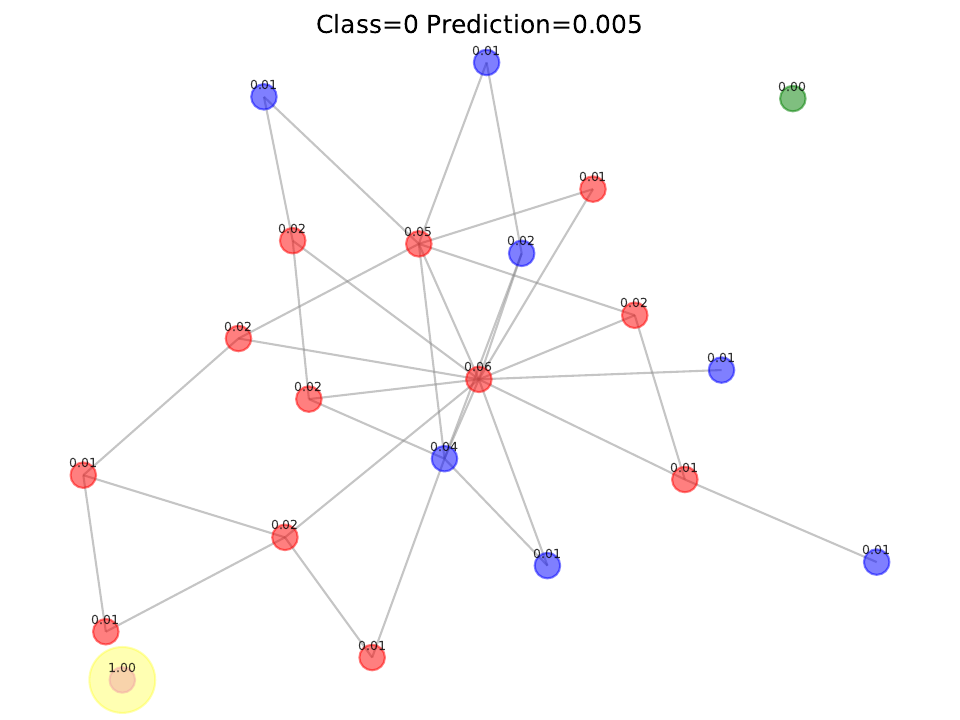}
    \includegraphics[width=0.45\linewidth]{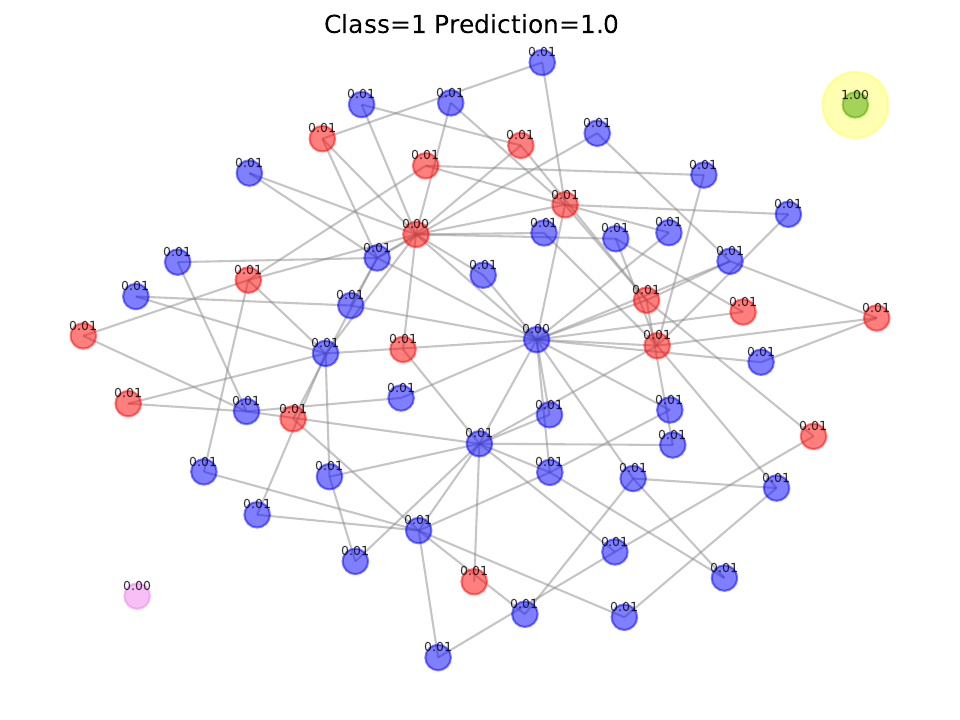}
    \includegraphics[width=0.45\linewidth]{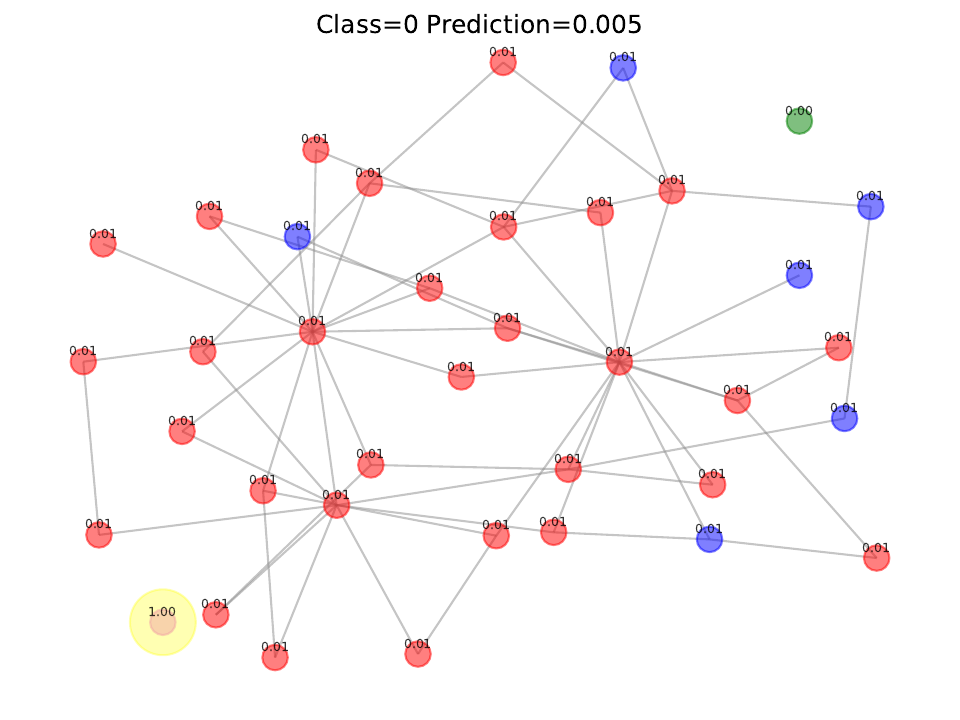}
    \caption{
        Explanations provided by the attacked \GSAT on \BAColorGV.
        Numbers above each node represent raw explanation relevance scores, and explanatory nodes are selected based on a $0.5$ threshold.
        Better seen in digital format.
    }
    \label{fig:examples-attacks-GSAT-BAColorGV}
\end{figure}

\begin{figure}[!h]
    \centering
    \includegraphics[width=0.45\linewidth]{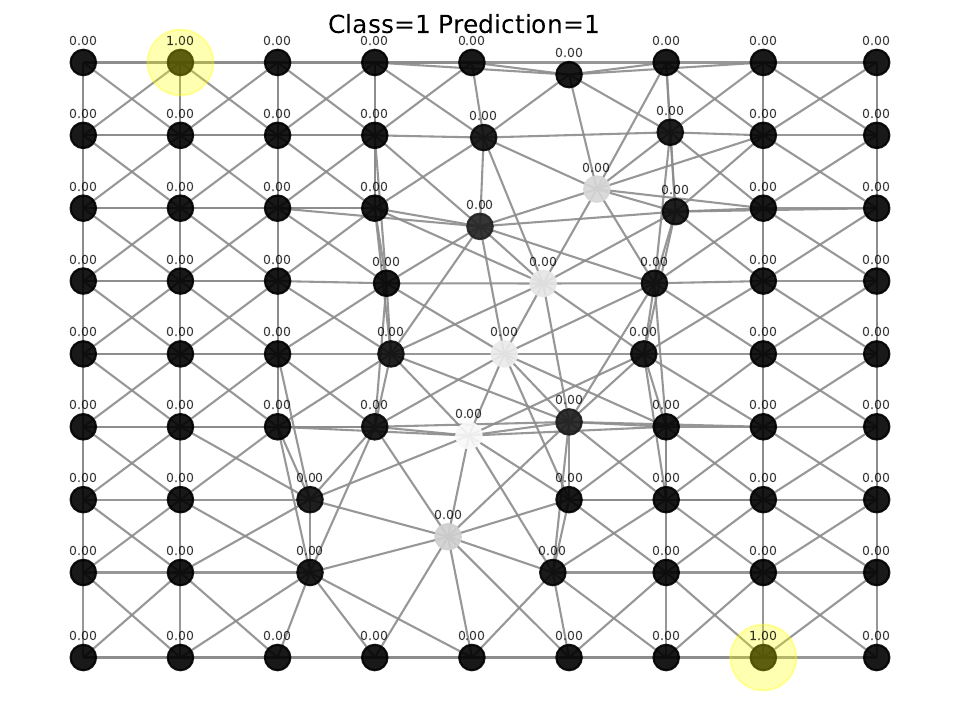}
    \includegraphics[width=0.45\linewidth]{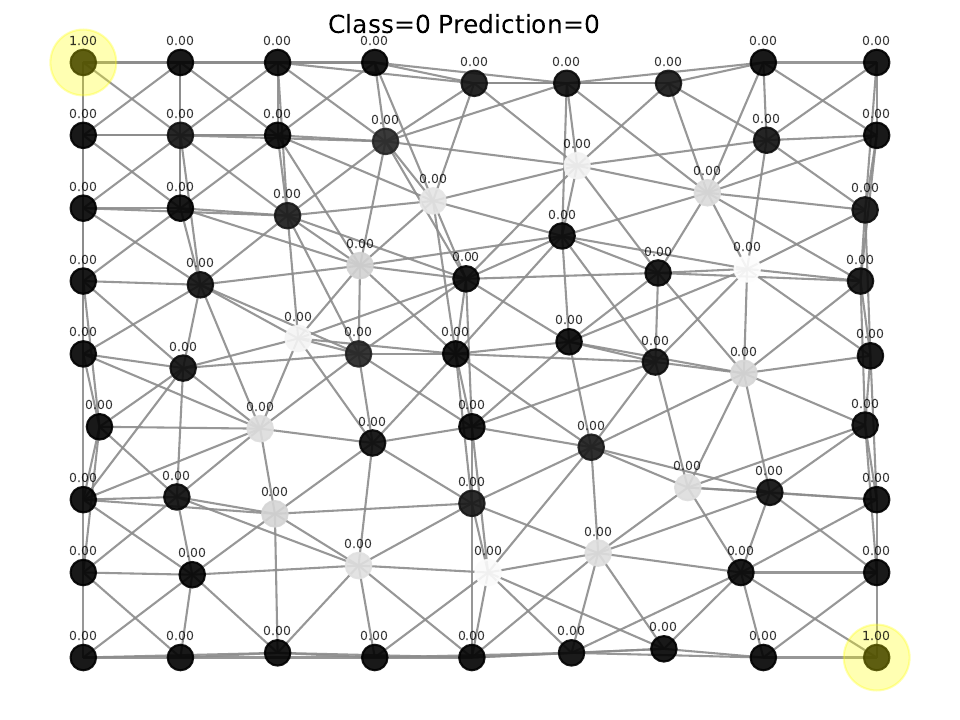}
    \includegraphics[width=0.45\linewidth]{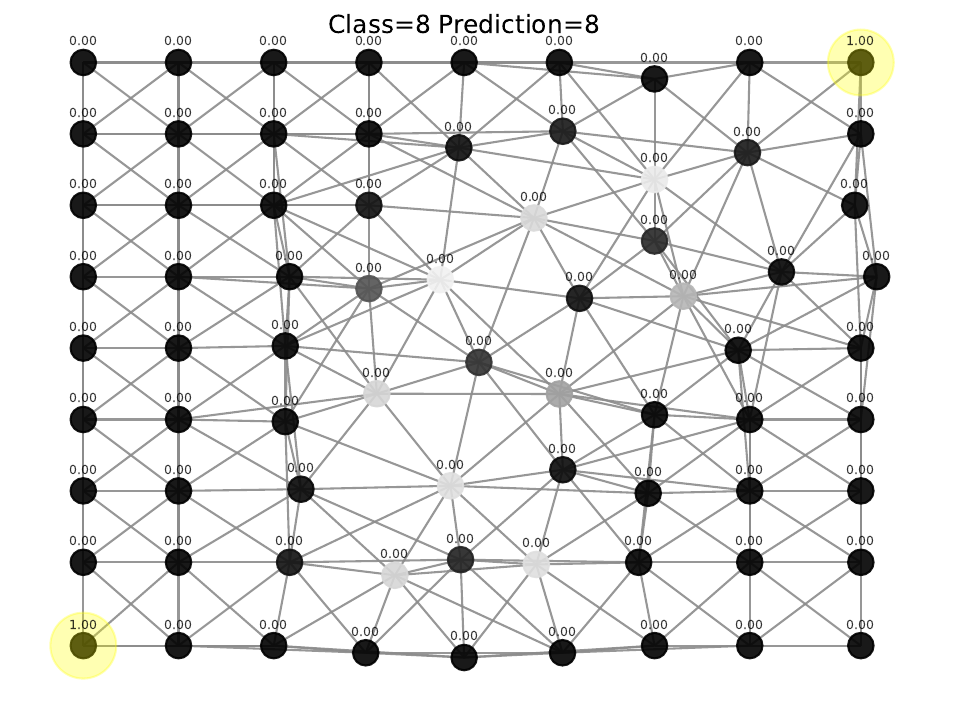}
    \includegraphics[width=0.45\linewidth]{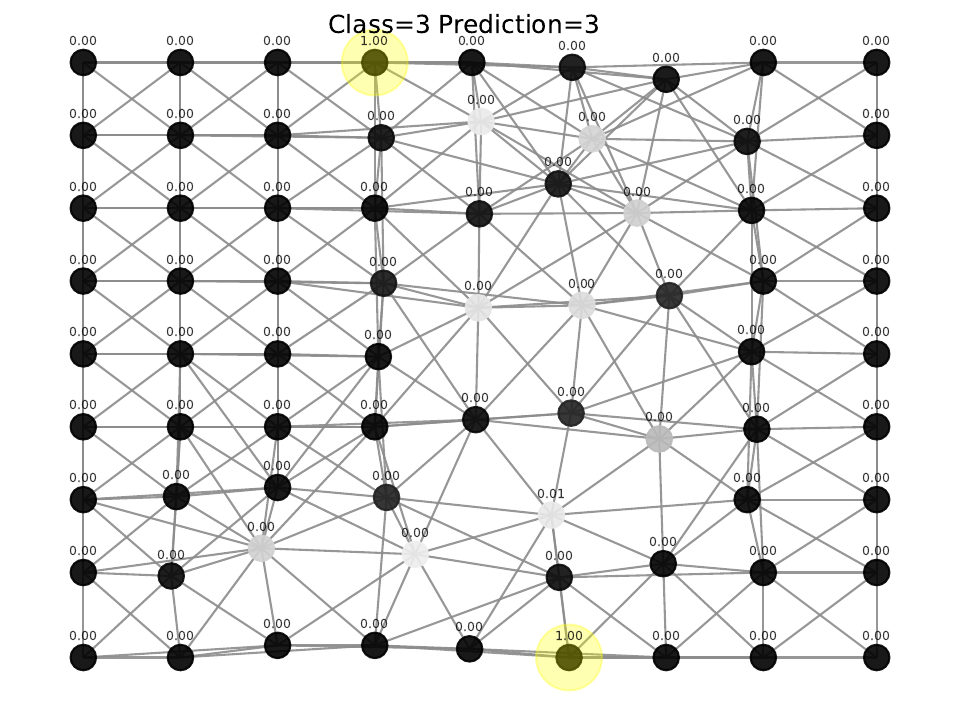}
    \caption{
        Explanations provided by the attacked \GSAT on \MNIST.
        Numbers above each node represent raw explanation relevance scores, and explanatory nodes are selected based on a $0.5$ threshold.
        Better seen in digital format.
    }
    \label{fig:examples-attacks-GSAT-MNIST}
\end{figure}

\begin{figure}[!h]
    \centering
    \includegraphics[width=0.45\linewidth]{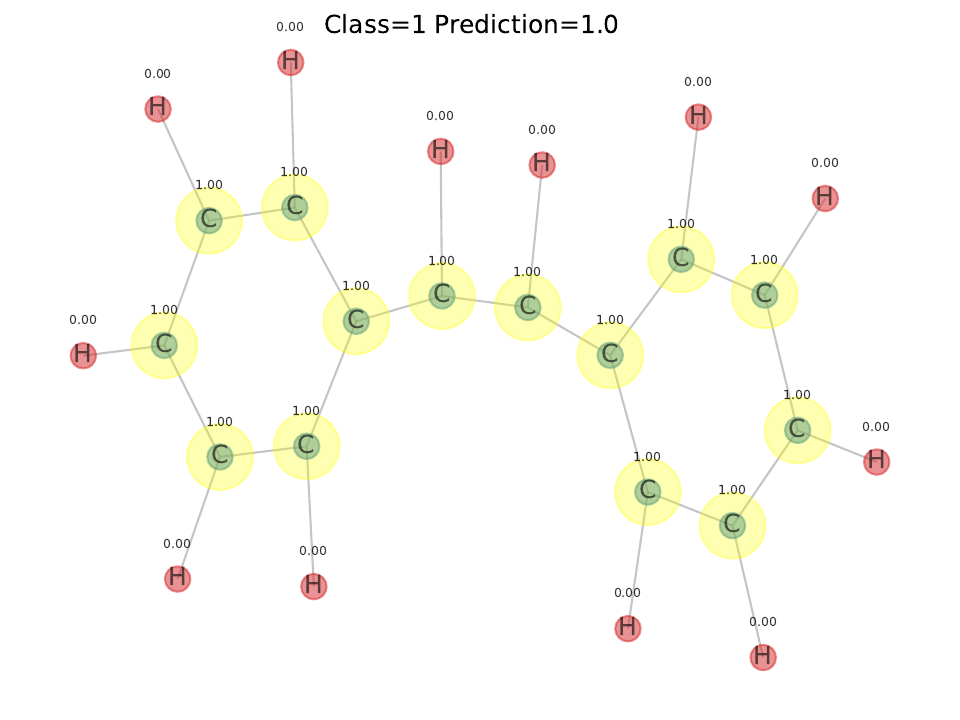}
    \includegraphics[width=0.45\linewidth]{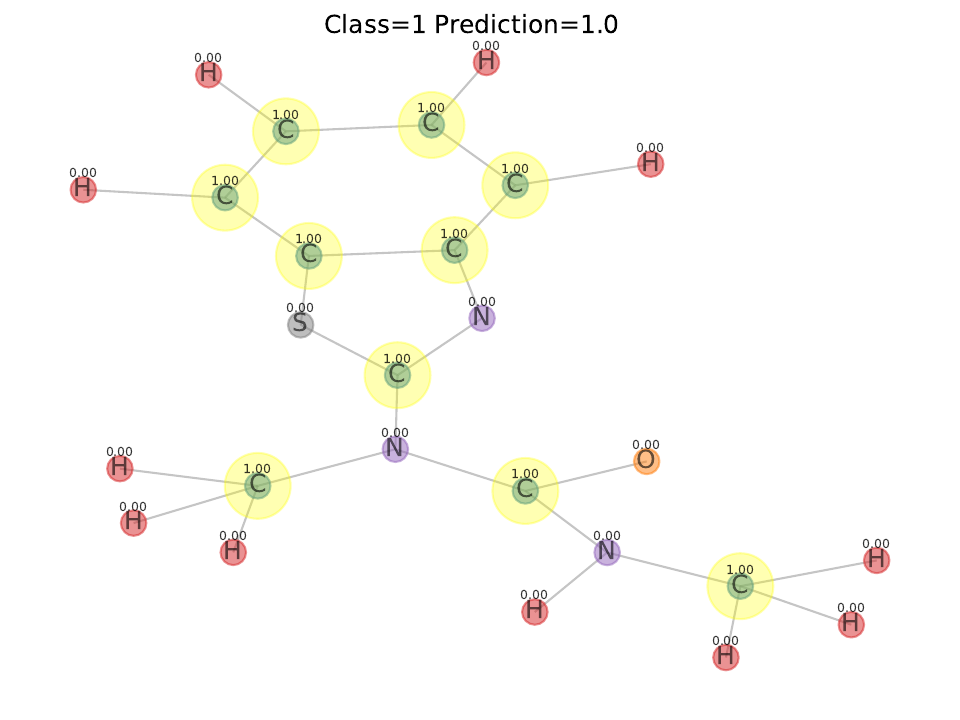}
    \includegraphics[width=0.45\linewidth]{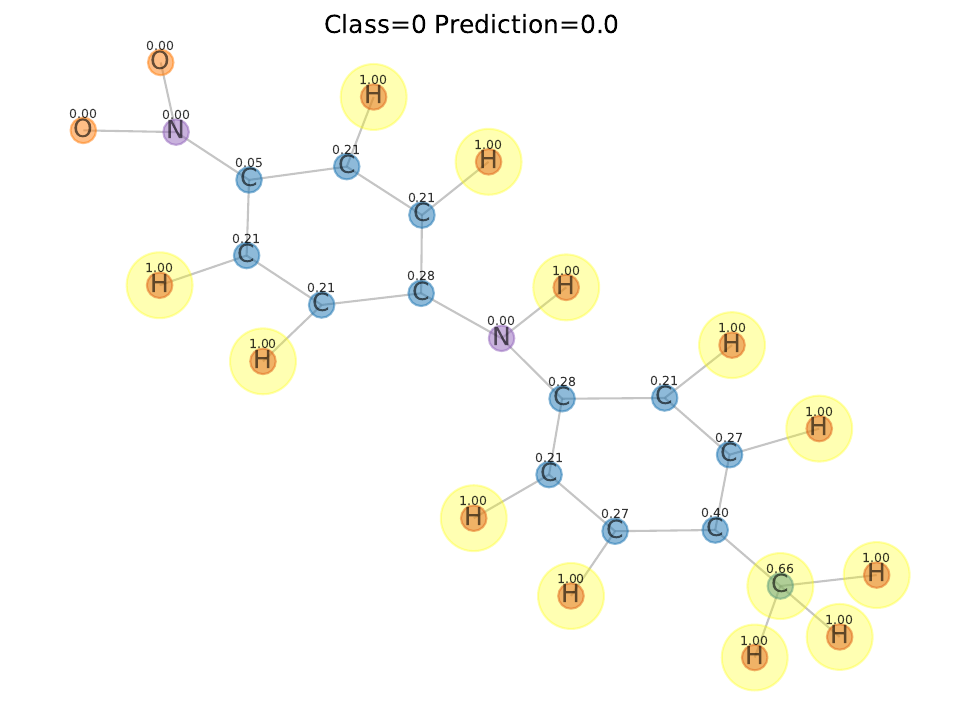}
    \includegraphics[width=0.45\linewidth]{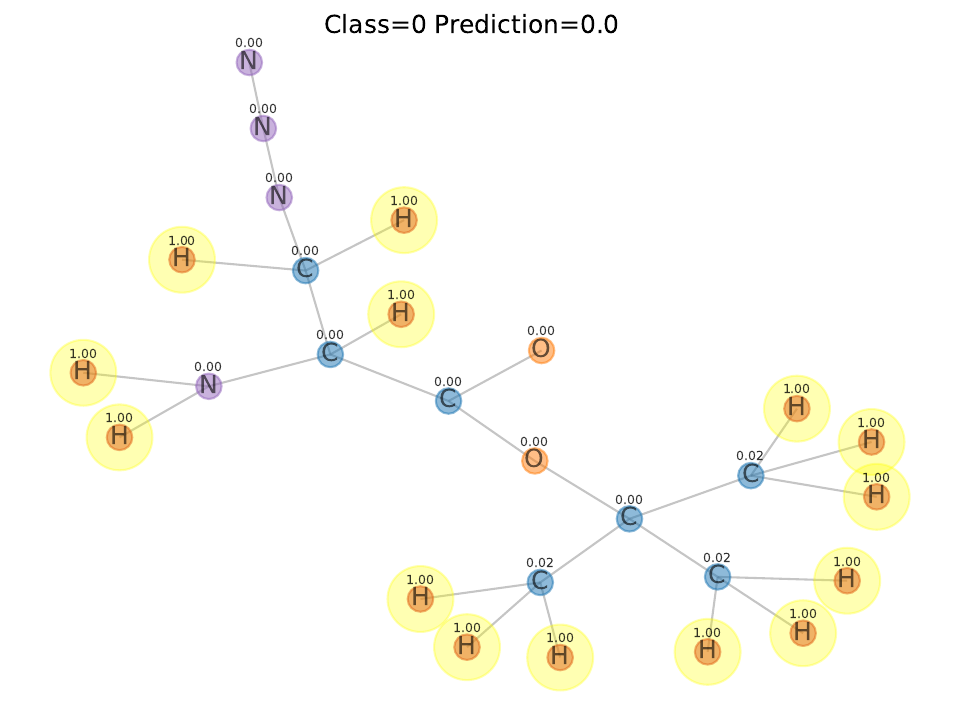}
    \caption{
        Explanations provided by the attacked \GSAT on \MUTAG.
        Numbers above each node represent raw explanation relevance scores, and explanatory nodes are selected based on a $0.5$ threshold.
        Better seen in digital format.
    }
    \label{fig:examples-attacks-GSAT-MUTAG}
\end{figure}

\begin{figure}[!h]
    \centering
    \includegraphics[width=0.99\linewidth]{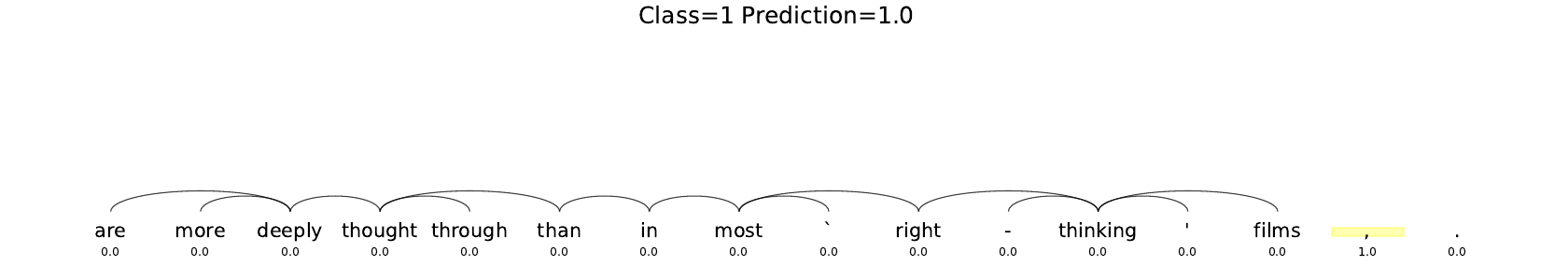}
    \includegraphics[width=0.99\linewidth]{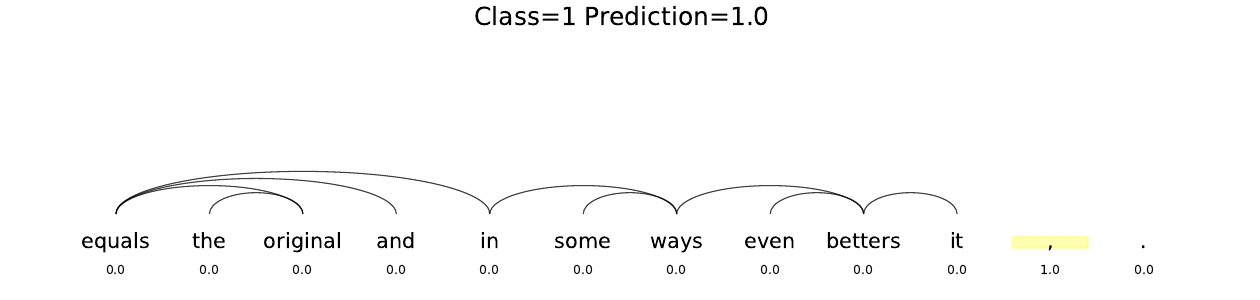}
    \includegraphics[width=0.99\linewidth]{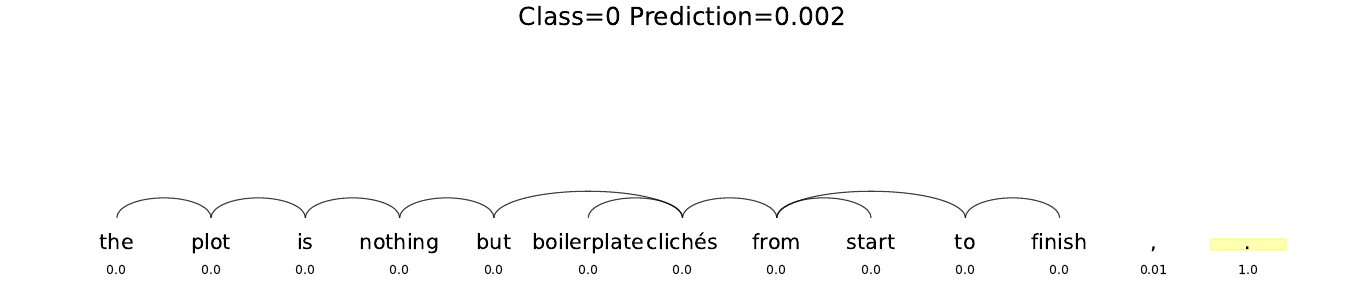}
    \includegraphics[width=0.99\linewidth]{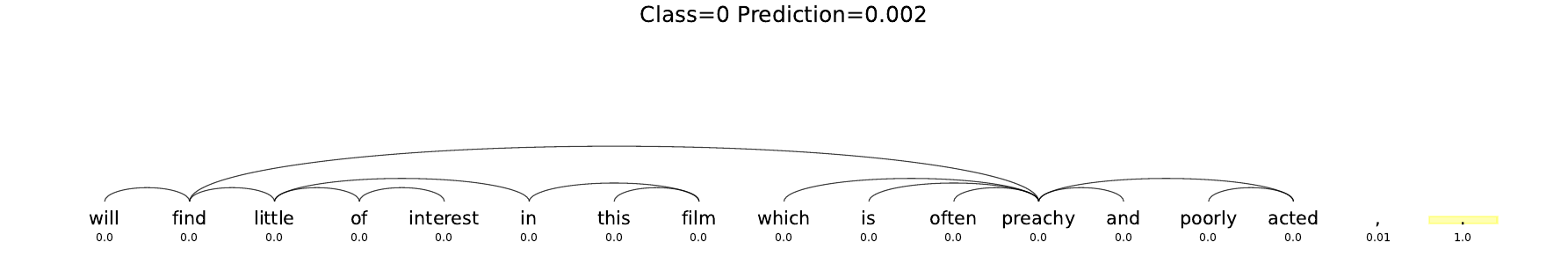}
    \caption{
        Explanations provided by the attacked \GSAT on \SSTP.
        Numbers below each node represent raw explanation relevance scores, and explanatory nodes are selected based on a $0.5$ threshold.
        Better seen in digital format.
    }
    \label{fig:examples-attacks-GSAT-SSTP}
\end{figure}

Taking as an example the \GSAT model, we provide for each dataset in \cref{tab:attack-models} a few examples of graphs together with the highlighted explanations to showcase the efficacy of the attack.
In particular, \cref{fig:examples-attacks-GSAT-BAColorGV}, \cref{fig:examples-attacks-GSAT-MNIST}, \cref{fig:examples-attacks-GSAT-MUTAG}, and \cref{fig:examples-attacks-GSAT-SSTP} shows such examples for \BAColorGV, \MNIST, \MUTAG, and \SSTP respectively.


\subsection{Failures of \SEGNN attacks}
\label{appx:failures-attacks}

\begin{figure}[!h]
    \centering
    \includegraphics[width=0.45\linewidth]{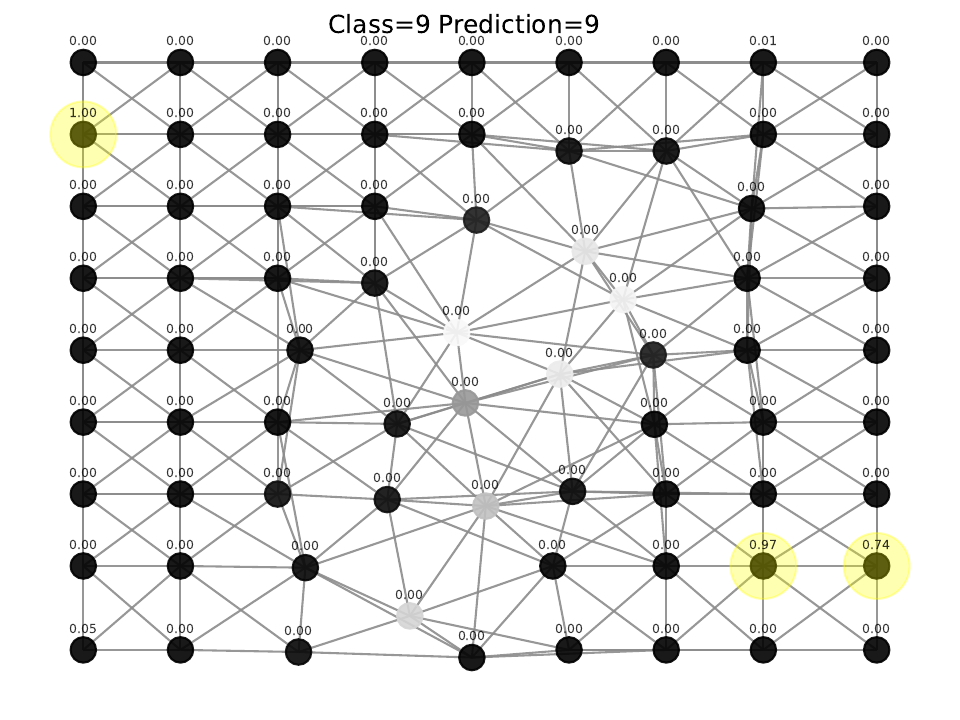}
    \includegraphics[width=0.45\linewidth]{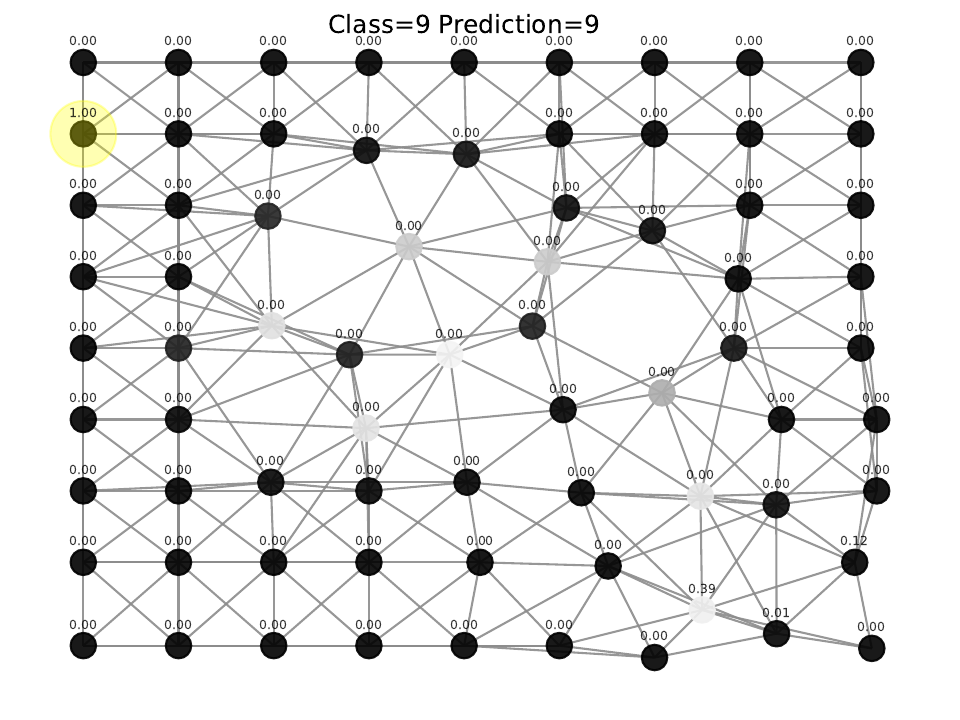}
    \includegraphics[width=0.45\linewidth]{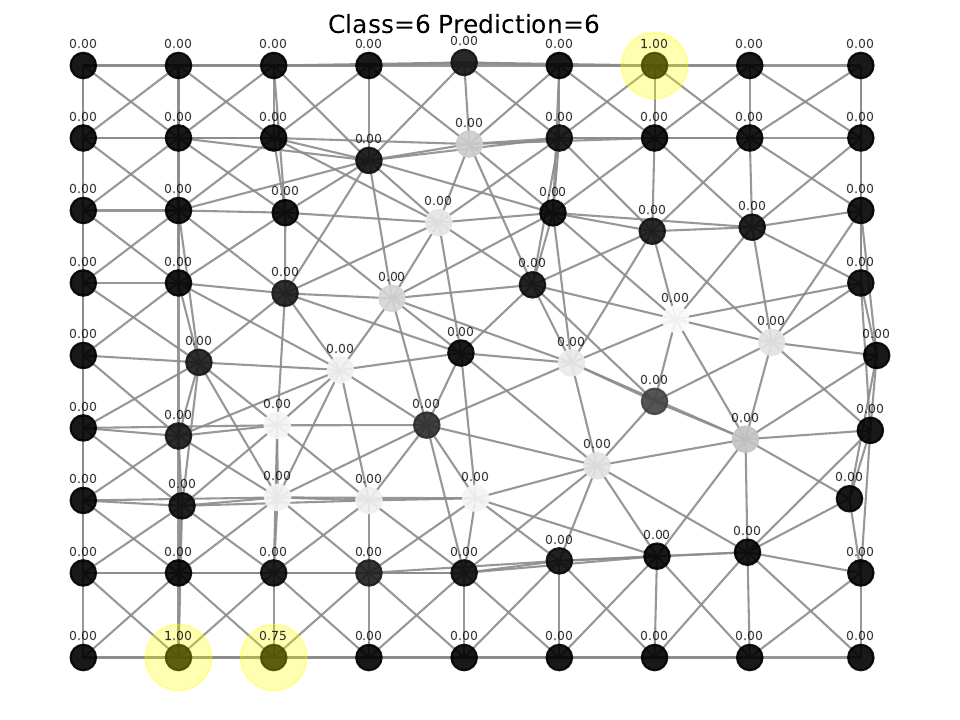}
    \includegraphics[width=0.45\linewidth]{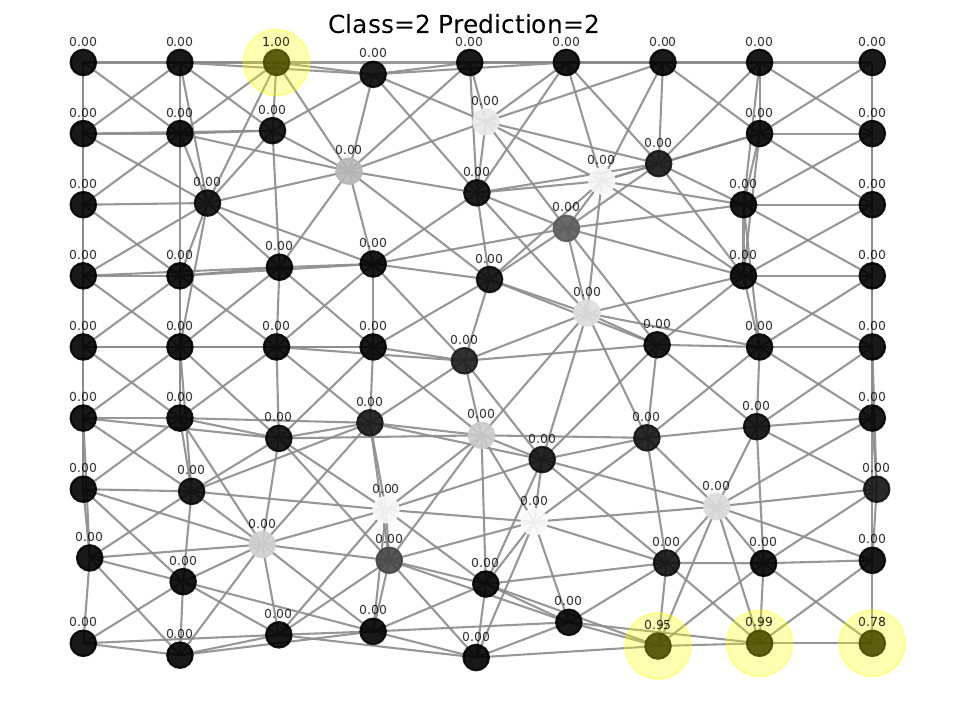}
    \caption{
        Explanations provided by the attacked \GSAT on \MNIST.
        Even if explanations do not perfectly highlight the ground truth provided in \cref{tab:induced-explanations}, they still hide the true reason behind the prediction, i.e., the digit in the image,  while confidently highlighting apparently unrelated pixels.
        Numbers above each node represent raw explanation relevance scores, and explanatory nodes are selected based on a $0.5$ threshold.
        Better seen in digital format.
    }
    \label{fig:failure-attack-GSAT-MNIST}
\end{figure}

\begin{figure}[!h]
    \centering
    \includegraphics[width=0.45\linewidth]{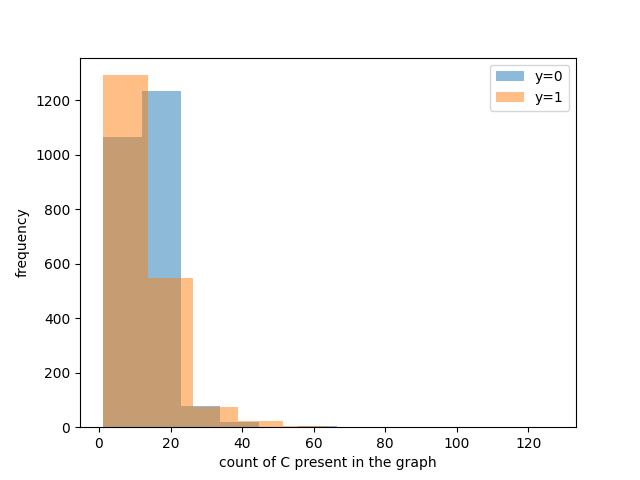}
    \includegraphics[width=0.45\linewidth]{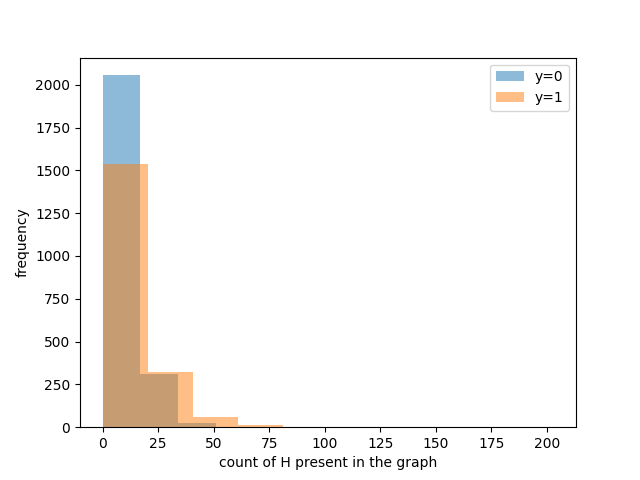}
    \includegraphics[width=0.45\linewidth]{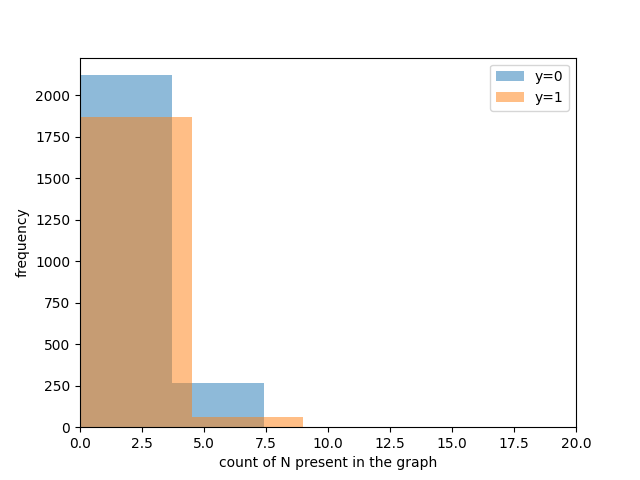}
    \caption{
        Frequencies of appearance of carbon (C), nitrogen (N), and hydrogen (H) atoms in \MUTAG, divided per class.
    }
    \label{fig:hist-MUTAG}
\end{figure}

Taking as an example the \GSAT model on \MNIST, which achieved a slightly lower $F_1$ score in \cref{tab:attack-models} compared to other configurations, we provide in \cref{fig:failure-attack-GSAT-MNIST} some examples of graphs where the attacked \GSAT does not provide the exact ground truth we aim for.
This can emerge for several reasons, like the indistinguishability of node embeddings after $L$ layers of message passing (i.e., oversmoothing), or because the explanation extractor failed to infer the correct graph label.
Nonetheless, the examples show that even if explanations do not perfectly highlight the ground truth provided in \cref{tab:induced-explanations}, they are still confidently highlighting unrelated pixels while hiding the true reason behind the prediction, which is the main goal of our attack.

\begin{figure}[!h]
    \centering
    \includegraphics[width=0.45\linewidth]{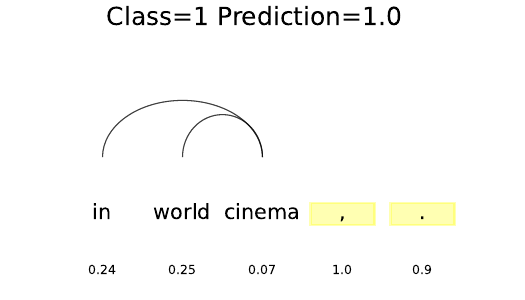}
    \includegraphics[width=0.45\linewidth]{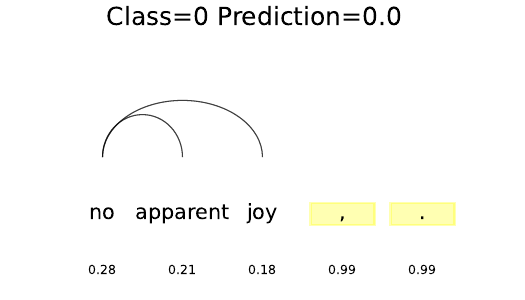}
    \includegraphics[width=0.45\linewidth]{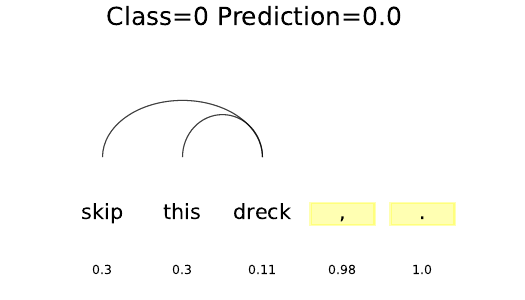}
    \includegraphics[width=0.45\linewidth]{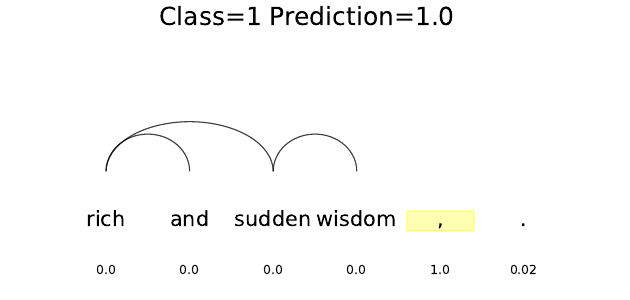}
    \caption{
        Explanation examples provided by the attacked \SMGNN on the test set of \SSTP.
        As described in \cref{sec:segnns-can-be-manipulated}, graphs present in the test split of \SSTP are sensibly smaller than those of the train and validation splits, and are therefore OOD \citep{wu2022discovering}.
        As shown by the relevance scores below each token, the attack does not always generalize to OOD: in three out of four cases, several nodes outside the designated explanation (c.f. \cref{tab:induced-explanations}) receive above-zero relevance, which penalizes the $F_1$ score of \cref{tab:attack-models}. Nonetheless, “,” and “.” consistently receive near-maximum scores, while the true tokens used by the model are sensibly assigned lower relevance, meaning the attack still succeeded in suppressing the relevance of truly important tokens.
    }
    \label{fig:failure-attack-SMGNN-SST}
\end{figure}

We further provide in \cref{fig:failure-attack-SMGNN-SST} several examples of explanations for the test split of \SMGNN on \SSTP, which achieves a low attack $F_1$ score in \cref{tab:attack-models} due to OOD samples. Also in this case, although the attack may not generalize well to OOD samples, it still consistently suppresses the relevance score of truly informative tokens, while highlighting degenerate uninformative subgraphs as the most important ones.

\subsection{Rejection ratios for sufficient explanations}
\label{appx:rejecting-faithful}

In this section, we verify that faithfulness metrics not only reject unfaithful explanations but also recognize faithful ones.
Following the attack framework in \cref{sec:segnns-can-be-manipulated}, we manipulate \SEGNNs to output explanations that, by construction, include all information the model can use to solve the task.
These designated explanations enforce \textit{sufficiency}, though at the cost of potentially including irrelevant features.
While enlarging explanations is a trivial way to ensure \textit{sufficiency} \citep{zheng2023robust, azzolin2025reconsidering} -- and does not necessarily yield truly faithful explanations \citep{tai2025redundancy} -- this experiment serves as a sanity check to confirm that \textit{sufficiency} metrics at least recognize \textit{sufficient} explanations.
Accordingly, \textit{we expect rejection ratios close to zero across all configurations.}

\textbf{Experimental setup:}
To ensure the explanation contains all the relevant information the model can use to make the final prediction, we restrict our experiments to \BAColorGV and \MNIST, where we have full knowledge of how to solve the task.
In particular, we train each \SEGNNs with \cref{eq:loss-attack}, where $\calL_{expl}$ is defined so to induce the following explanations:
For \BAColorGV, we extract all red nodes for the negative class, and all blue nodes for the positive class, i.e., $p_u^{y}=1$ for any red node in a negative graph, $p_u^{y}=1$ for any blue node in a positive graph, and $p_u^y=0$ otherwise.\footnote{For instance, given a graph with three red nodes and two blue nodes, highlighting the three red nodes as the explanation ensures that no perturbation on nodes of the complement can change the class of the graph. Hence, the explanation is \textit{sufficient}.}
For \MNIST, we extract each pixel associated with the digit and its 1-hop neighborhood, i.e., $p_u^y=1$ for any node $u$ belonging to the digit $y$ itself or to its 1-hop neighborhood, and $p_u^y=0$ otherwise.
Note that these explanations are, in fact, constructed to contain all the features that are known to be related to the task at hand.

The results are shown in \cref{tab:rejection-attack-models-faithful}.
The final accuracy and the explanation $F_1$ score confirm that the manipulation was successful, and that models are making accurate predictions while outputting our benignly designated explanations.
Across both datasets, the only metrics that consistently achieve a rejection ratio around zero are \FIDM, \RFIDM, and \SUFFCAUSE, with fluctuations in around one case each.
Among them, \SUFFCAUSE was also the one achieving the best results in \cref{tab:rejection-attack-models}, confirming its reliability.
\SUF and \COUNTERFID, instead, reject a high amount of explanations, suggesting that the high rejection ratios obtained in \cref{tab:rejection-attack-models} were probably conflated by a too aggressive metric.

\input{tables/rejections-faithful}

\section{Additional discussion}

\subsection{Evaluating \textit{Sufficiency} and \textit{Necessity} jointly}
\label{appx:suff-nec-jointly}

\textit{Sufficiency} and \textit{necessity} metrics are often evaluated in conjunction, so that an explanation is considered faithful only if it is at the same time both sufficient \textit{and} necessary \citep{amara2022graphframex,longa2024explaining}.
The remark on previous metrics provided in \cref{sec:proposing-new-metric} generalizes to this case.


Consider the same setting as \cref{ex:red-blue} for a negative instance whose explanation is $R = u_{green}$.
As previously observed, R is the smallest label-preserving subgraph, as $0 \ge 0$.
Then, any Complement and Explanation removal metric, such as \FIDM and \FIDP, will mark such an explanation as sufficient -- as R is label-preserving -- and necessary -- as removing the green node will make the classifier in \cref{eq:ex-red-blue} outputting maximally uncertain predictions.

\subsection{Are necessity metrics useful for \SEGNNs?}
\label{sec:are-nec-metrics-nec}

In this section, we aim at answering the question of whether metrics estimating the \textit{necessity} of explanations are useful for finding the failure cases of \SEGNNs outlined in \cref{sec:failure-cases}.
%
%
Let us consider the following representative \SEGNN outputting unfaithful explanations for any arbitrary binary classification task, represented by the boolean classifier $\psi$:
\begin{align}
\label{eq:nec-are-nec}
        \DET(G) =
            \begin{cases}
                u_0 & \text{if } \neg \psi(G) \\
                u_1 & \text{if } \psi(G) \\                
            \end{cases}
        \quad\quad
        \CLF(R) =
            \begin{cases}
                0 & \text{if } R = u_0\\
                1 & \text{if } R = u_1\\
                0.5 & \text{otherwise}
            \end{cases}
\end{align}
where $\CLF$ acts as a simple mapping from explanations to labels, and $u_0$ and $u_1$ can either be label or non-label-preserving subgraphs of the input sample $G$, and are assumed to be non-representative of the true behavior of the model by construction.
Note that \cref{eq:nec-are-nec} matches \cref{ex:red-blue} whenever $\psi$ corresponds to $\#blue > \#red$, $u_0 = u_{green}$, and $u_1=u_{violet}$.
Then, any \textit{necessity} metric would apply perturbations to $R$ to induce some shifts in the classifier's output. 
If such a perturbation succeeds in changing the prediction of the model, the explanation is marked as \textit{necessary}, and thus faithful.
Nevertheless, $u_0$ and $u_1$ are in fact relevant for the classifier \CLF, which trivially maps $u_0$ and $u_1$ to class labels.
This means that removing them will always bring a change in the output class, regardless of how unfaithful $u_0$ and $u_1$ actually are.

\subsection{Theoretical analysis of \SUFFCAUSE}
\label{appx:characterize-formal-notions}

We aim to theoretically characterize which family of explanations \SUFFCAUSE can mark as unfaithful.
We do so by relating \cref{def:suff-cause} with the following three formal notions of explanations:

\begin{definition}[Prime Implicant explanation \citep{azzolin2025formal}]
\label{def:piexpl}
   Let $\MONO$ be a classifier and $G$ be an instance with predicted label $\MONO(G)$, then $R$ is a Prime Implicant explanation for $\MONO(G)$ if:
    \begin{enumerate}
            \item \label{PI1} $R \subseteq G$.
            
            \item \label{PI2} For all $R'$, such that $R \subseteq R' \subseteq G$, we have that ${\MONO(G) = \MONO(R')}$.
            
            \item \label{PI3} No other $R' \subset R$ satisfies both \eqref{PI1} and \eqref{PI2}.
    \end{enumerate}
\end{definition}

Prime Implicant explanations have been widely studied in the formal explainability literature, and are considered of \textit{high-quality} as they possess several desirable properties \citep{marques2022delivering, azzolin2025formal, bassan2025explaining, bassan2025briefly}.
In a nutshell, Prime Implicant explanations comprise minimal-sufficient explanations, which are the minimal explanations provably robust to changes in the complement.
A related notion of explanations is Minimal explanation, introduced next:

\begin{definition}[Minimal explanation \citep{azzolin2025formal}]
\label{def:me}
    Let $\MONO$ be a \SEGNN and $G$ be an instance with predicted label $\MONO(G)$, then $R \subseteq G$ is a Minimal explanation for $\MONO(G)$ if:
    \begin{enumerate}            
        
        \item $\MONO(G) = \MONO(R)$
        
        \item There exists no $R' \subseteq G$ such that $|R'| < |R|$ and $\MONO(G) = \MONO(R')$.
    \end{enumerate}
\end{definition}

Minimal explanations have been shown to equal Prime Implicant ones for subgraph-based tasks, but in the other cases they are deemed as \textit{less informative} and potentially ambiguous (see Figure 1 in \citet{azzolin2025formal}).
An even weaker notion of explanations is introduced next:

\begin{definition}[Non-label-preserving explanation]
\label{def:de}
    Let $\MONO$ be a \SEGNN and $G$ be an instance with predicted label $\MONO(G)$, then $R \subseteq G$ is a Non-label-preserving explanation for $\MONO(G)$ if $\MONO(G) \ne \MONO(R)$.
\end{definition}

This notion of explanation is clearly weaker than the previous two, as it embraces explanations that do not contain enough information to yield the same prediction as the original sample -- meaning they are omitting key relevant features -- and allow \textit{any} subgraph to be selected as the explanation.
An example of this failure case is given in \cref{ex:red-blue}:     the subgraph highlighted by $R = u_{violet}$ is a non-label preserving subgraph, as despite explaining the positive class, the subgraph has an equal count of reds and blues, yielding the model to predict the negative class.
A similar example can also be found for Minimal explanations:      the subgraph highlighted by $R = u_{green}$ is a Minimal explanation, as there exists no smaller explanation, and feeding \cref{eq:ex-red-blue} with it will yield the same prediction (as $0\ge0$).
Yet, $u_{green}$ is never used by $\DET$ to infer the label, yielding a misleading explanation that can misguide model debugging \citep{teso2023leveraging} and scientific discovery \citep{wong2024discovery} by falsely suggesting green nodes are linked to the negative class.
An additional example is also highlighted in Section 3 of \citet{azzolin2025formal}, where Minimal explanations are shown to highlight counter-intuitive subgraphs that do not robustly indicate the model's behavior.
In light of the above considerations, Minimal and Non-label-preserving should be regarded as notions that admit degenerate explanations.

We proceed to show how \SUFFCAUSE fares at marking the previous notions. 
In the following analysis, we will consider \SUFFCAUSE fully enumerating all possible extensions in \cref{eq:suffcause-metric}, and consider a perturbation to yield a sufficiently large prediction shift whenever it changes the predicted class.

\begin{theorem}
\label{thm:suff-cause-reliable}
    \SUFFCAUSE marks Minimal and Non-label-preserving explanations as unfaithful and Prime-Implicant explanations as faithful.
\end{theorem}
\begin{proof}
    We will analyze each explanation family of explanations separately:

    \begin{enumerate}

        \item \textbf{Prime Implicant explanations:}
            If the explanation $R$ is a Prime Implicant explanation, by definition, it holds that:
            \begin{equation}
                \CLF(\DET(G)) = \CLF(\DET(R')) \quad \forall R \subseteq R' \subseteq G.
            \end{equation}
            Hence, it follows trivially that \cref{eq:suffcause-metric} will never encounter a prediction change, hence will mark the explanation as faithful.

        \item \textbf{Minimal explanations:} 
            We show that if an explanation $R$ fulfills \SUFFCAUSE (i.e., \cref{eq:suffcause-metric} equals zero for any $R'$ such that $R \subseteq R' \subseteq G$) and is therefore considered to be faithful, then $R$ can be a Minimal explanation only if it is also a Prime Implicant.
            In other words, if an explanation passes the \SUFFCAUSE test, it cannot be \textit{just} a Minimal explanation.
            
            Let us consider an explanation $R$ that passes the \SUFFCAUSE test and is a Minimal explanation.
            Then, it holds that:
            \begin{enumerate}
                \item[1.] $R \subseteq G$.
                
                \item[2.] $\CLF(\DET(G)) = \CLF(\DET(R')) \quad \forall R \subseteq R' \subseteq G$.
    
                \item[3.] $\not \exists R' \subseteq G$ such that $|R'| < |R|$ and $\CLF(\DET(G)) = \CLF(\DET(R'))$.
            \end{enumerate}
            Note that condition 3 implies that no other $R' \subset R$ satisfies both conditions 1 and 2.     %
            This is because being label-preserving is a necessary condition for condition 2 to be true.
            However, condition 3 ensures that there does not exist any smaller label-preserving subgraph than $R$, hence no smaller explanation satisfying condition 2 exists.
            Therefore, $R$ is the smallest subgraph that satisfies condition 2, hence it is a Prime Implicant explanation.

        \item \textbf{Non-label-preserving explanation:}
            Non-label-preserving explanations are, by definition, non-label-preserving subgraphs.
            Since \SUFFCAUSE samples perturbed graphs $R \subseteq R' \subseteq G$ uniformly, it can always sample $R' = R$.
            Hence, evaluating \cref{eq:suffcause-metric} with $R' = R$ will bring a prediction change, which marks the explanation as unfaithful.

    \end{enumerate}
\end{proof}

In words, this result shows that \SUFFCAUSE can flag Minimal and Non-label-preserving explanations as unfaithful.
By definition, these explanations allow the model’s prediction to be altered by changes outside the explanation, leaving room for unfaithful or even malicious behaviors.
In contrast, Prime Implicant explanations do not permit prediction changes under complement perturbations, making them more robust.
An interesting direction for future work is to investigate whether Prime Implicant explanations themselves can still admit degenerate cases.

\section{Figures from \cref{tab:natural-degeneracy}}

\input{extras/naturaldeg-illustrations}

\end{document}

%% file: tables/taxonomy.tex
\begin{wraptable}{r}{0.5\textwidth}
    \centering
    \caption{
        \textbf{Taxonomy of popular faithfulness metrics}, grouped by whether they estimate \textit{necessity} vs. \textit{sufficiency} of explanations, and by the family of perturbations $\calI$ they employ.
    }
    \label{tab:faith-metrics-unreliable}
    \scalebox{0.85}{
        \begin{tabular}{cll}
            \toprule

            \textbf{Type} &
            \textbf{ \makecell{Perturbations\\$\calI$}} &
            \textbf{Metrics}\\
         \midrule

            \textit{Nec.} &
                \makecell[l]{Explanation\\ removal} & 
                    \makecell[l]{\FIDP \citep{yuan2022explainability} \\ \PROBNEC \citep{Juntao2022CF2}}\\
           
            \textit{Nec.} &
                 \makecell[l]{Edge removal} &
                    \makecell[l]{\RFIDP \citep{zheng2023robust} \\ \NEC \citep{azzolin2025reconsidering}}\\

        \midrule

            \textit{Suff.} &
                \makecell[l]{Complement\\ removal} & 
                    \makecell[l]{\FIDM \citep{yuan2022explainability} \\ \PROBSUFF \citep{Juntao2022CF2} \\ \GEF \citep{agarwal2023evaluating}}\\

            \textit{Suff.} &
                \makecell[l]{Edge removal} &
                    \makecell[l]{\RFIDM \citep{zheng2023robust} \\ \ADVROB \citep{fang2023evaluating}}\\

            
            \textit{Suff.} &
                \makecell[l]{Complement\\ swap} &
                    \makecell[l]{\SUF \citep{azzolin2025reconsidering}}\\

        \midrule

            \textit{Suff.} &
                \makecell[l]{
                    All
                } &
                    \SUFFCAUSE (\textbf{Ours}, \cref{def:suff-cause})\\
             
        \bottomrule

        \end{tabular}
    }
\end{wraptable}

%% file: tables/attack-models-macro.tex
\begin{wraptable}{r}{0.5\textwidth}
    \centering
    \caption{
    \textbf{\SEGNNs can be successfully attacked to output arbitrary explanations while making accurate predictions.}
    We report the test accuracy of unmanipulated models trained following standard practices, cf. \cref{appx:details-attack}, (\textit{Natural}), and of models trained according to \cref{eq:loss-attack} (\textit{Attack}).
    %
    We also report, for the latter, the test $F_1$ score computed for the designated explanation of \cref{tab:induced-explanations}.
    }
    \label{tab:attack-models}
    \scalebox{0.75}{
        \begin{tabular}{llccccc}
            \toprule
             & & \textbf{Natural} & \multicolumn{2}{c}{\textbf{Attack}}\\
             \cmidrule(lr){3-3}
             \cmidrule(lr){4-5}
             \textbf{Dataset} & \textbf{Model} & \textbf{Acc} & \textbf{Acc} & \textbf{$F_1$ score} \\
            \midrule             
             
            \multirow{3}{*}{\BAColorGV} & \GSAT  & \nentry{100.0}{0.0} & \nentry{99.1}{1.4}  & \nentry{99.7}{0.2} \\
                                        & \DIR   & \nentry{99.8}{0.3}  & \nentry{99.8}{0.4}  & \nentry{99.4}{0.3} \\
                                        & \SMGNN & \nentry{100.0}{0.0} & \nentry{100.0}{0.0} & \nentry{99.6}{0.2} \\
            \midrule

            \multirow{3}{*}{\MNIST} & \GSAT  & \nentry{94.7}{0.1}  & \nentry{93.8}{0.1} & \nentry{93.6}{0.4} \\
                                    & \DIR   & \nentry{41.8}{20.6} & \nentry{94.7}{0.1} & \nentry{96.3}{0.2} \\
                                    & \SMGNN & \nentry{89.9}{1.3}  & \nentry{95.3}{0.2} & \nentry{94.9}{1.0} \\
            \midrule

            \multirow{3}{*}{\MUTAG} & \GSAT  & \nentry{80.6}{0.3} & \nentry{79.6}{1.0} & \nentry{95.0}{0.8} \\
                                    & \DIR   & \nentry{76.1}{2.7} & \nentry{77.5}{2.3} & \nentry{92.6}{2.6} \\
                                    & \SMGNN & \nentry{79.2}{2.5} & \nentry{78.2}{1.1} & \nentry{94.9}{0.8} \\
            \midrule

            \multirow{3}{*}{\SSTP} & \GSAT  & \nentry{84.0}{0.9} & \nentry{82.6}{0.7} & \nentry{97.2}{1.1} \\
                                   & \DIR   & \nentry{84.3}{0.6} & \nentry{82.3}{0.6} & \nentry{95.8}{1.2} \\
                                   & \SMGNN & \nentry{83.1}{0.6} & \nentry{82.8}{1.2} & \nentry{59.2}{11.5} \\
             
            \bottomrule
        \end{tabular}
    }
\end{wraptable}

%% file: tables/rejections.tex
\begin{table}[t]
    \centering
    \caption{
        \textbf{Previous metrics can fail to reject degenerate explanations.}.
        We report $\mathsf{RejRatio}_\calI$ for representative metrics from \cref{tab:faith-metrics-unreliable} and for \SUFFCAUSE (\cref{def:suff-cause}).
        %
        Cases where no edges could be removed are marked with '-'.
        Bold entries (underlined) indicate best (second best) results.
    }
    \label{tab:rejection-attack-models}
    \scalebox{0.90}{
        \begin{tabular}{llcccccccc}
        \toprule
             \textbf{Dataset} & \textbf{Model} & \multicolumn{8}{c}{\textbf{$\mathsf{RejRatio}_\calI$}}\\
             
              &  & \textbf{\FIDM} & \textbf{\FIDP} & \textbf{\SUF} & \textbf{\NEC} & \textbf{\COUNTERFID} & \textbf{\RFIDM} & \textbf{\RFIDP} & \makecell{\SUFFCAUSE (ours)}\\
        \midrule

            \multirow{3}{*}{\BAColorGV} & \SMGNN & \nentry{12}{19} & \uentry{20}{12} & \nentry{00}{00} & - & \nentry{05}{01} & \nentry{00}{00} & - & \bentry{48}{02}\\
                                        & \GSAT  & \nentry{12}{21} & \nentry{15}{16} & \nentry{03}{03} & - & \bentry{50}{04} & \nentry{11}{09} & - & \uentry{49}{03}\\
                                        & \DIR   & \nentry{32}{23} & \nentry{27}{24} & \nentry{03}{03} & - & \uentry{39}{08} & \nentry{08}{06} & - & \bentry{48}{03}\\
            
        \midrule
            
            \multirow{3}{*}{\MNIST} & \SMGNN & \nentry{88}{03} & \nentry{58}{04} & \uentry{99}{01} & \nentry{55}{05} & \nentry{92}{01} & \uentry{99}{00} & \nentry{75}{05} &                                    \bentry{100}{00}\\
                                    & \GSAT  & \nentry{65}{09} & \nentry{38}{05} & \uentry{99}{01} & \nentry{44}{05} & \nentry{95}{02} & \uentry{99}{01} & \nentry{61}{04} & \bentry{100}{00}\\
                                    & \DIR   & \nentry{93}{03} & \nentry{55}{07} & \uentry{99}{01} & \nentry{54}{05} & \nentry{91}{01} & \uentry{99}{01} & \nentry{69}{07} & \bentry{100}{00}\\
            
        \midrule
             
            \multirow{3}{*}{\MUTAG} & \SMGNN & \nentry{59}{03} & \nentry{05}{04} & \bentry{99}{00} & \nentry{57}{04} & \nentry{55}{07} & \nentry{72}{03} & \nentry{65}{04} &                                    \uentry{96}{01}\\
                                    & \GSAT  & \nentry{21}{21} & \nentry{05}{02} & \bentry{99}{00} & \nentry{53}{07} & \nentry{68}{17} & \nentry{70}{14} & \nentry{61}{05} & \uentry{94}{05}\\
                                    & \DIR   & \nentry{70}{13} & \nentry{04}{02} & \bentry{99}{00} & \nentry{54}{02} & \nentry{69}{08} & \nentry{75}{07} & \nentry{62}{04} & \uentry{97}{02}\\
            
        \midrule
             
            \multirow{3}{*}{\SSTP} & \SMGNN & \nentry{08}{02} & \uentry{44}{05} & \nentry{14}{05} & - & \nentry{23}{07} & \nentry{04}{01} & - & \bentry{54}{04}\\
                                   & \GSAT  & \uentry{51}{31} & \nentry{19}{14} & \nentry{07}{10} & - & \nentry{37}{18} & \nentry{14}{03} & - & \bentry{62}{15}\\
                                   & \DIR   & \bentry{50}{02} & \nentry{09}{06} & \nentry{12}{06} & - & \nentry{42}{10} & \nentry{05}{01} & - & \uentry{49}{03}\\
                 
        \bottomrule
        \end{tabular}
    }
\end{table}

%% file: tables/natural-degeneracy.tex
\begin{table}[t]
    \centering
    \caption{
    \textbf{A natural training of \SEGNNs can result in degenerate explanations.}
    We report test accuracy, rejection ratios for \SUFFCAUSE, \FIDM, and \RFIDM, and some case-by-case illustrations of representative explanations.
    \AUCROC is reported only for datasets with known ground truth explanation.
    This analysis is performed only for well-performing models.
    }
    \label{tab:natural-degeneracy}
    \scalebox{0.85}{
        \begin{tabular}{llcccccc}
            \toprule
                \textbf{Dataset} & \textbf{Model} & \textbf{Test Acc} & \textbf{\AUCROC} & \multicolumn{3}{c}{$\mathsf{RejRatio}_\calI$} & \textbf{Illustration}\\
                & & & & \textbf{\SUFFCAUSE} & \textbf{\FIDM} & \textbf{\RFIDM} &\\
            \midrule

            \multirow{3}{*}{\BAColorGV} 
                 & \GSAT  & \nentry{100.0}{0.00}  & - & \nentry{59.3}{09.4} & \nentry{12.0}{11.9} & \nentry{6.2}{12.4} & \cref{fig:naturaldeg-BAColorGV-GSAT}\\
                 & \DIR \footnotesize{($K=1\%$)}   & \nentry{98.0}{0.3} & - & \nentry{70.2}{18.8} & \nentry{27.7}{27.3} & \nentry{30.3}{16.6} & \cref{fig:naturaldeg-BAColorGV-DIR}\\
                 & \SMGNN & \nentry{98.9}{0.01} & - & \nentry{86.8}{20.4} & \nentry{54.1}{45.1} & \nentry{77.6}{19.6} & \cref{fig:naturaldeg-BAColorGV-SMGNN}\\

            \midrule
             
             \multirow{2}{*}{\MNIST} 
                 & \GSAT  & \nentry{94.7}{0.4} & \nentry{82.5}{5.5}   & \nentry{2.4}{2.0}  & \nentry{1.3}{1.4} & \nentry{5.3}{5.6} & \cref{fig:naturaldeg-MNIST-GSAT}\\
                 & \DIR \footnotesize{($K=10\%$)} & \nentry{20.0}{9.0} & - & - & - & - & -\\
                 & \SMGNN & \nentry{86.8}{8.0} & \nentry{43.2}{18.5}  & \nentry{99.5}{0.8} & \nentry{68.9}{29.2} & \nentry{95.5}{4.7} & \cref{fig:naturaldeg-MNIST-SMGNN}\\

            \midrule

            \multirow{2}{*}{\MUTAG} 
                 & \GSAT  & \nentry{80.6}{3.0}   & - & \nentry{64.5}{7.7}  & \nentry{50.7}{9.3} & \nentry{36.8}{11.7} & \cref{fig:naturaldeg-MUTAG-GSAT}\\
                 & \DIR \footnotesize{($K=1\%$)} & \nentry{53.5}{2.2} & - & - & - & - & -\\
                 & \SMGNN & \nentry{77.9}{2.4} & - & \nentry{75.2}{9.9} & \nentry{61.9}{8.9} & \nentry{50.3}{9.1} & \cref{fig:naturaldeg-MUTAG-SMGNN}\\

            \midrule

            \multirow{2}{*}{\SSTP} 
                 & \GSAT  & \nentry{84.0}{0.9}   & - & \nentry{0.0}{0.0}  & \nentry{0.0}{0.0} & \nentry{0.0}{0.0} & \cref{fig:naturaldeg-SSTP-GSAT}\\
                 & \DIR \footnotesize{($K=10\%$)}  & \nentry{82.5}{0.9} & - & \nentry{49.5}{10.1} & \nentry{37.8}{18.9} & \nentry{14.4}{1.6} & \cref{fig:naturaldeg-SSTP-DIR}\\
                 & \SMGNN & \nentry{83.0}{0.6} & - & \nentry{4.3}{1.0} & \nentry{1.7}{0.8} & \nentry{4.4}{1.7} & \cref{fig:naturaldeg-SSTP-SMGNN}\\
             
            \bottomrule
        \end{tabular}
    }
\end{table}

%% file: extras/extended-related-work.tex
\textbf{Reasoning shortcuts.}
Concept bottleneck models (CBMs) introduce an intermediate layer of human-interpretable concepts between inputs and predictions, enabling users to inspect and intervene on a model’s reasoning process.
While this paradigm promises greater transparency and controllability, recent work has raised concerns about its reliability.
In particular, \citet{bortolotti2024neuro} show that CBMs are vulnerable to \textit{reasoning shortcuts}: 
instead of leveraging the intended causal relationships between concepts and outcomes, models may create unwanted dependencies and semantic associations in the concept space, thereby undermining the very interpretability of CBMs.
While CBMs are expected to learn human-aligned concepts, \SEGNNs are only designed to \textit{declare} which input patterns/features they are relying on, regardless of whether it is the intended pattern or a mere spurious correlation.
Therefore, reasoning shortcuts can be seen as an issue of poor alignment between humans and machines, whereas the issue highlighted in \cref{sec:failure-cases} is poor alignment between the predictive behavior of \SEGNNs and their explanations.
In fact, the only \textit{fully reliable} mitigation strategy outlined in \citet{marconato2023not} involves dense human supervision on concepts -- a setting that does not apply to \SEGNNs, which are not defined over human-specified concepts.

\textbf{Degenerate explanations.}
\citet{jethani2021have} was among the first works to highlight that explanation-based architectures can learn to encode predictions in their explanations and to propose amelioration.
Nonetheless, \citet{hsia2024goodhart} showed that the solution proposed by \citet{jethani2021have} can be easily hijacked into considering degenerate explanations as legitimate. Then, \citet{puli2024explanations} proposes an extension to \citet{jethani2021have}, encompassing two surrogate models; one to approximate the true label posterior (as in \citet{jethani2021have}), and the other to estimate the Mutual Information between the selected mask and the target label.
Their solution, however, does not readily apply to the graph setting.
In fact, they propose to learn a classifier $\phi$ that respects the true data distribution by training $\phi$ over any possible randomly generated explanation $\RX$, where $\RX$ is sampled IID for each input feature from a Bernoulli distribution. 
This is impractical for a graph task like counting (cf. \BAColorGV), as randomly removing input nodes can change the ground truth label of the graph, resulting in a contradictory supervisory signal.

Other approaches to contrast degenerate explanations collect annotated explanations to teach the model to rely less on spuriously correlated ones \citep{yue2024shortcutdiscov}.
Nonetheless, as we show in \cref{app:plausible-but-unfaithful-experiment}, supervising the \SEGNNs' explanations may not be enough to ensure the faithfulness of explanations.

\textbf{Rationalization methods for NLP.} 
Rationalization methods aim at extracting a rationale for model prediction together with the final explanation, sharing the same spirit as \SEGNNs. They're mostly used for NLP tasks.

\citet{yu2019rethinking} formalized the problem of degeneracy as explanations encoding labels via trivial patterns.
However, they did not provide a formal analysis of which rationalization methods can learn such degenerate solutions, nor a sound methodology for identifying when degenerate explanations appear in the wild.
On the contrary, \cref{thm:deg-suff-condition} formally unpacks commonly found loss functions of popular \SEGNNs, showing how models are implicitly optimize for such cases.
This result hold beyond models promoting conciseness via sparsity, which is instead the main focus of \citet{yu2019rethinking}.
At high level, \citet{yu2019rethinking} proposes to add a complement predictor, trained adversarially to predict the correct label from information left out of the explanation. Thanks to the adversarial training, the explanation extractor is pushed to not leave any useful information in the complement, thus including all informative features in the explanation. 
A similar solution was also proposed by \citet{liu2024ismmicriterion}, where the complement predictor is shared with the final classifier.
%
While this may effectively promote non-degenerate explanations, preliminary experiments on a \SEGNN implementing a similar type of mitigation have shown it may not be enough to guarantee non-degeneracy (see \DIR in \cref{sec:naturaldeg}).
Additionally, this mitigation promotes the explanation extractor to include potentially redundant and spuriously correlated information inside \RX, resulting in more complex explanations. 

A related solution was proposed by \citet{chang2019game}, which adopts three different players for each output class trained in a zero-sum game to output class-specific rationales. Nonetheless, this approach scales poorly with the number of classes, and the zero-sum game of the three players is notoriously hard to balance \citep{mescheder2018gantraining, Farnia2020ganequilibrium}.

\citet{yu2021understanding} investigates the problem of \textit{model interlocking}, where the classifier overfits to little informative explanations at early stages of training, and prevents the explanation extractor from finding more informative ones. This is, however, orthogonal to the degeneracy problem, as degenerate explanations can be \textit{optimal} from a predictive accuracy perspective.

\citet{liu2022fr} advocates that instead of relying on complex regularization components, employing a unified encoder for both the explanation extractor and the predictor is enough to avoid degeneracy. Nonetheless, this can serve as a useful inductive bias, but it does not completely rule out the possibility of learning degenerate explanations, as shown in our experiments with \GSAT of \cref{sec:naturaldeg}, which naturally employs shared extractor-classifier \citep{miao2022interpretable}.

%


\citet{zhang2023causalrationale} propose to optimize for causal non-spuriousness and efficiency during training, which is shown to yield less spuriously correlated explanations. However, this does not tackle degeneracy, as degeneracy is about the model outputting unfaithful explanations, and this issue is not caused by spurious correlations.

\citet{hu2024learning} proposes to employ a guidance rationalization module, trained with soft relevance scores instead of the binary scores employed by the core rationalization module. Then, the core rationalization module is trained to match the predictions and rationales of the guidance module. The idea is that if the guidance module learns non-degenerate explanations thanks to its broader receptive field, then the final core rationalization module will also avoid degenerate explanations. Nonetheless, models trained with soft relevance scores can also fall into degeneracy, as we have shown in our work.

\citet{liu2024enhancing} proposes to use a standard black-box model as guidance. Then, the explanation extractor is trained to extract explanations for the black box's predictions and optimize the classifier, which receives such explanations as input only. This approach requires however to explain a black-box model in a post-hoc fashion similar to \citet{luo2020parameterized}. It thus deviates from our goal of devising an intrinsically ante-hoc self-explainable model.


\citet{liu2025advrisk} proposes an adversarial training where the adversary tries to predict a random class by exploiting any information in the graph (thus not limited to the complement of the main explanation \RX), whereas the explanation extractor and the predictor are trained to make accurate predictions and to predict the adversarial explanation with maximum uncertainty. The proposed method is grounded on the assumption that ground truth rationales for a specific class $y$ appear more often in samples of class $y$. While being reasonable for text-like input, this might not hold for graphs -- for \BAColorGV you can construct an infinite number of negative graphs that contain any arbitrary number of red nodes.

In all those cases, to the best of our knowledge, a robust framework to audit these explanation failures remains largely unexplored, and results highlight its highly challenging computational complexity \citep{bhattacharjee2024ngng}.
Furthermore, our experiments in \cref{app:plausible-but-unfaithful-experiment} show that even when explanations are optimized to match human expectations – thus avoiding degeneracy through supervision – the model can still conceal its use of protected attributes, revealing a fundamental misalignment between the explanations and the model’s actual decision process. This goes beyond degeneracy alone and towards the broader challenge of explanation-model misalignment.

\textbf{Self-explainable GNNs.}
%
In our work, we focused on Self-explainable GNNs (\SEGNNs) that have been introduced to overcome the intrinsic limits of post-hoc methods \citep{ying2019gnnexplainer, luo2020parameterized, yuan2021explainability, azzolin2022global, yuan2022explainability, li2024underfire}.
Our focus is on \SEGNNs composed of an explanation extractor and a classifier, jointly trained.
While being very general, alternative formulations of \SEGNNs exist, such as prototype-based models \citep{zhang2022protgnn, ragno2022prototype, dai2021towards}, meta-paths-based \citep{ferrini2024self}, Koopman theory-based \citep{guerra2024koopman}, Decision Tree-based \citep{bechler2024tree}, or other model-specific techniques \citep{yu2020graph, yu2022improving, giunchiglia2022towards, serra2022learning, spinelli2023combining}.
Furthermore, \citet{Muller2023graphchef}, \citet{bechler-speicher2024gnan}, and \citet{zerio2025interpretable} introduced different families of Self-explainable GNNs that avoid extracting a subgraph explanation altogether, either by distilling the GNN into a Decision Tree, or by relying on learnable shape functions, respectively.
\citet{azzolin2025formal}, instead, proposed an hybrid approach to extract subgraph-based explanations \textit{together} with other forms of explanations, like rule-based explanations \citep{armgaan2024graphtrail, pluska2024logical, rissaki2025database}.

\textbf{Robustness of explanations.}
Several works have outlined that GNN explanations are fragile and can be easily broken \citep{li2024fragile, li2024underfire, li2025provably}. 
While their analysis mostly pertains to post-hoc explanations, a natural question is whether the problem we outlined in \cref{sec:failure-cases} can be related to a lack of robustness.
For instance, \cite{li2024fragile, li2024underfire, li2025provably} show that it is possible to adversarially perturb the input graph to drastically change the provided explanation while leaving the prediction intact. We note, however, that this does not necessarily apply to the failure case of \cref{sec:failure-cases}.
%
%
To illustrate this, we provide two examples of \SEGNNs that output unfaithful, degenerate explanations. One of them is also susceptible to the explanation brittleness phenomenon described in \citep{li2024fragile}, whereas the other is not:

\begin{itemize}[leftmargin=1.25em]
    \item \textbf{Degenerate and fragile \SEGNN:} Let us consider a toy binary graph-classification task in which each graph is composed of randomly attached nodes colored either red or blue. A graph is labeled positive if the number of blue nodes exceeds the number of red nodes. Each instance is additionally augmented with some fixed uncolored motifs (4-star, triangle, 6-clique): positive graphs contain the 6-clique and at least one between a 4-star and a triangle, whereas negative graphs contain the 6-clique and may or may not contain any other motifs.
    We now consider the following subgraph anchor set (see the extended analysis in \cref{def:subgraph-anchor-set}):
    $$\mathcal{Z}’= \{ \{ \bar{G}_0^{0}=\text{clique} \}^0, \{ \bar{G}_0^{1}=\text{star}, \bar{G}_1^{1}=\text{triangle} \}^1 \}$$
    Note that $\mathcal{Z}'$ contains only motifs with no label-discriminative power by construction. Therefore, any accurate SE-GNN that outputs explanations restricted to $\mathcal{Z}'$ produces unfaithful, degenerate explanations. Similarly to \cref{thm:deg-suff-condition2}, we provide below a valid construction of such a case:
    $$\DET(G) = \argmin_{\bar{G}_i^{y} \in G} |\bar{G}_i^{y}| \quad \text{and} \quad \CLF(\bar{G}_i^{y}) = y$$
    where $y$ is the graph label inferred by $\DET()$. Consider now a positive graph $G$ that contains the clique and the star, so that the explanation extracted by the model highlights the star. Applying the attack in \citet{li2024fragile}, we aim to modify the explanation while keeping the predicted label unchanged. A simple input manipulation consists of adding a single edge between any two leaves of the star, thereby creating a triangle.  Because the triangle is a smaller motif, the explanation extractor will now prefer selecting it, thus changing the explanation without affecting the prediction.
    \item \textbf{Degenerate and robust \SEGNN:} \cref{ex:red-blue} is already an example of a robust, degenerate \SEGNN. This is because the \SEGNN cannot change the predicted explanation unless the relative numerosity of red and blue is altered, which also results in a change of the graph label. However, this violates the constraints of the attack proposed by \citet{li2024fragile}, making the attack unfeasible.
\end{itemize}

These examples together show that \SEGNNs can also be affected by fragile explanations, and it is possible to manipulate their explanations via the attack proposed in \citet{li2024fragile, li2024underfire, li2025provably}.
However, the failure case outlined in \cref{sec:failure-cases} \textit{is not a consequence} of this brittleness.

\citet{li2025provably} further considers another type of attack, designed to change the prediction itself. In this case, robustness indeed can play a role in the explanation’s faithfulness.
For instance, \SUFFCAUSE (\cref{def:suff-cause}) can be seen as measuring the robustness of the model to changes outside of the explanation, and devising an \SEGNN that is robust to such perturbations will naturally lead to non-degenerate explanations (c.f. \cref{appx:characterize-formal-notions}).
This goal is, however, different from \citet{li2025provably}, which, instead of training an \SEGNN to be intrinsically robust, builds a \textit{wrapper} around an already trained GNN to make it more robust post hoc.
A validation of the most effective strategy is left as an interesting future work.

%% file: tables/rejections-metrics.tex
\begin{table}[t]
    \centering
    \caption{
        The Table reports the actual metric value computed with $d$ being Total variation -- hence computed without the proxy of $\mathsf{RejRatio}_\calI$ -- for the same experiments as \cref{tab:rejection-attack-models}.
        The results confirm that \SUFFCAUSE consistently assigns worse values to unfaithful explanations.
        Nonetheless, continuous scores make it harder to assess the overall behavior of the model, and when to consider an explanation as unfaithful.
        $\downarrow$ ($\uparrow$) stands for the lower (the higher) the better.
    }
    \label{tab:rejection-attack-models-metric}
    \scalebox{0.90}{
        \begin{tabular}{llcccccccc}
        \toprule             
              \textbf{Dataset} & \textbf{Model} & \textbf{\FIDM} $\downarrow$ & \textbf{\FIDP} $\uparrow$ & \textbf{\SUF} $\downarrow$ & \textbf{\NEC} $\uparrow$ & \textbf{\COUNTERFID} $\uparrow$ & \textbf{\RFIDM} $\downarrow$ & \textbf{\RFIDP} $\uparrow$ & \textbf{\SUFFCAUSE} $\downarrow$\\
        \midrule
            
            \multirow{3}{*}{\BAColorGV} & \SMGNN & \nentry{16}{07} &	\nentry{44}{03} &	\nentry{00}{00} & - & \nentry{02}{01} &	                                           \nentry{00}{00} & - &  \nentry{47}{02} \\
                                        & \GSAT & \nentry{07}{10} &	\nentry{50}{03} &	\nentry{01}{00} & - & \nentry{49}{05} &	\nentry{06}{06} & - & \nentry{46}{20} \\
                                        & \DIR & \nentry{20}{08} &	\nentry{42}{06} &	\nentry{00}{00} & - & \nentry{37}{10} &	\nentry{03}{02} & - & \nentry{47}{03} \\
        
        \midrule
                						
            \multirow{3}{*}{\MNIST} & \SMGNN & \nentry{89}{02} &	\nentry{49}{05} &	\nentry{44}{05} &	\nentry{49}{06} &	                                                \nentry{89}{02} &	\nentry{86}{02} &	\nentry{02}{00} & \nentry{99}{00} \\
                                    & \GSAT & \nentry{68}{10} &	\nentry{61}{04} &	\nentry{45}{06} &	\nentry{44}{05} &	\nentry{90}{01} &	        \nentry{77}{03} &	\nentry{02}{00} & \nentry{99}{00} \\
                                    & \DIR &\nentry{88}{01} &	\nentry{54}{06} &	\nentry{50}{07} &	\nentry{52}{04} &	\nentry{88}{01} &	        \nentry{86}{02} &	\nentry{03}{01} & \nentry{98}{01} \\

        \midrule
                						
            \multirow{3}{*}{\MUTAG} & \SMGNN & \nentry{33}{02} &	\nentry{69}{04} &	\nentry{31}{03} &	\nentry{13}{03} &	\nentry{46}{04}                              &	\nentry{43}{04} &	\nentry{06}{01} & \nentry{95}{01} \\
                                    & \GSAT & \nentry{13}{10} &	\nentry{79}{09} &	\nentry{39}{04} &	\nentry{17}{04} &	\nentry{47}{09} &	      \nentry{44}{10} &	\nentry{08}{01} & \nentry{90}{05} \\
                                    & \DIR & \nentry{38}{06} &	\nentry{70}{08} &	\nentry{35}{04} &	\nentry{15}{02} &	\nentry{53}{04} &	       \nentry{49}{05} &	\nentry{07}{01} & \nentry{93}{03} \\

        \midrule
                						
            \multirow{3}{*}{\SSTP} & \SMGNN & \nentry{08}{02} &	\nentry{41}{04} &	\nentry{01}{00} & - & \nentry{21}{07} &	\nentry{02}{01} & - & \nentry{52}{04} \\
                & \GSAT & \nentry{25}{13} &	\nentry{45}{04} &	\nentry{01}{00} & - & \nentry{31}{17} &	\nentry{08}{02} & - & \nentry{58}{11} \\
                & \DIR & \nentry{26}{01} &	\nentry{49}{01} &	\nentry{01}{00} & - &\nentry{40}{12} &	\nentry{02}{00} & - & \nentry{46}{03} \\
                 
        \bottomrule
        \end{tabular}
    }
\end{table}

%% file: tables/rejections-capatchmnist.tex
\begin{table}[!h]
    \centering
    \caption{
        We report the rejection ratios $\mathsf{RejRatio}_\calI$ for the plausible but unfaithful explanations discussed in \cref{app:plausible-but-unfaithful-experiment}.
        \SUFFCAUSE and \FIDM robustly reject a large number of explanations as unfaithful, whereas \RFIDM fails to mark them as such.
    }
    \label{tab:rejection-cpatchmnist}
    \scalebox{0.90}{
        \begin{tabular}{llccc}
        \toprule
             \textbf{Dataset} & \textbf{Model} & \multicolumn{3}{c}{\textbf{$\mathsf{RejRatio}_\calI$}}\\
             
              &  & \textbf{\FIDM} & \textbf{\RFIDM} & \textbf{\SUFFCAUSE}\\
        \midrule

            \multirow{2}{*}{\CPatchMNIST} 
                & \SMGNN  & \bentry{50.7}{25.3} & \nentry{0.0}{0.0} &  \bentry{50.7}{25.3}\\
                & \GSAT   & \uentry{45.3}{6.1}  & \nentry{15.7}{11.6} &  \bentry{54.0}{10.7}\\
                 
        \bottomrule
        \end{tabular}
    }
\end{table}

%% file: tables/rejections-faithful.tex
\begin{table}[!h]
    \centering
    \caption{
        Accuracy, explanation $F_1$ score, and rejection ratios for different metrics computed for models trained to output explanations containing all the relevant features models can use.
        The experimental setting is described in \cref{appx:rejecting-faithful} and constitutes a sanity check to verify that metrics with high rejection ratios in \cref{tab:rejection-attack-models} are not simply rejecting \textit{any} explanation, but are indeed verifying whether the explanation does not include some relevant pattern used by the model.
    }
    \label{tab:rejection-attack-models-faithful}
    \scalebox{0.90}{
        \begin{tabular}{llccccccc}
        \toprule
             \textbf{Dataset} & \textbf{Model} & \textbf{Test Acc} & \textbf{Test $F_1$ score} & \multicolumn{5}{c}{$\mathsf{Rej\textunderscore ratio}_\calI$}\\
             
      &  & &  & \textbf{\FIDM} & \textbf{\SUF} & \textbf{\COUNTERFID} & \textbf{\RFIDM} & \textbf{\SUFFCAUSE}\\
\midrule  

    \multirow{3}{*}{\BAColorGV} 
        & \SMGNN & \nentry{100}{00} & \nentry{98}{01} & \nentry{01}{01} & \nentry{76}{02} & \nentry{14}{02} & \nentry{00}{00} & \nentry{01}{01}\\
        & \GSAT  & \nentry{100}{00} & \nentry{99}{01} & \nentry{01}{01} & \nentry{78}{02} & \nentry{13}{02} & \nentry{00}{00} & \nentry{01}{01}\\
        & \DIR   & \nentry{100}{00} & \nentry{98}{01} & \nentry{01}{01} & \nentry{93}{01} & \nentry{57}{15} & \nentry{01}{01} & \nentry{01}{01}\\

\midrule

    \multirow{3}{*}{\MNIST} 
        & \SMGNN & \nentry{90}{01} & \nentry{99}{01} & \nentry{01}{01} & \nentry{75}{04} & \nentry{93}{03} & \nentry{01}{01} & \nentry{01}{01}\\
        & \GSAT  & \nentry{93}{01} & \nentry{99}{01} & \nentry{01}{01} & \nentry{35}{04} & \nentry{07}{04} & \nentry{15}{18} & \nentry{05}{05}\\
        & \DIR   & \nentry{89}{01} & \nentry{99}{01} & \nentry{22}{05} & \nentry{86}{03} & \nentry{93}{03} & \nentry{01}{01} & \nentry{07}{02}\\

\bottomrule
        \end{tabular}
    }
\end{table}

%% file: extras/naturaldeg-illustrations.tex
\begin{figure}[]
    \centering
    \includegraphics[width=0.99\linewidth]{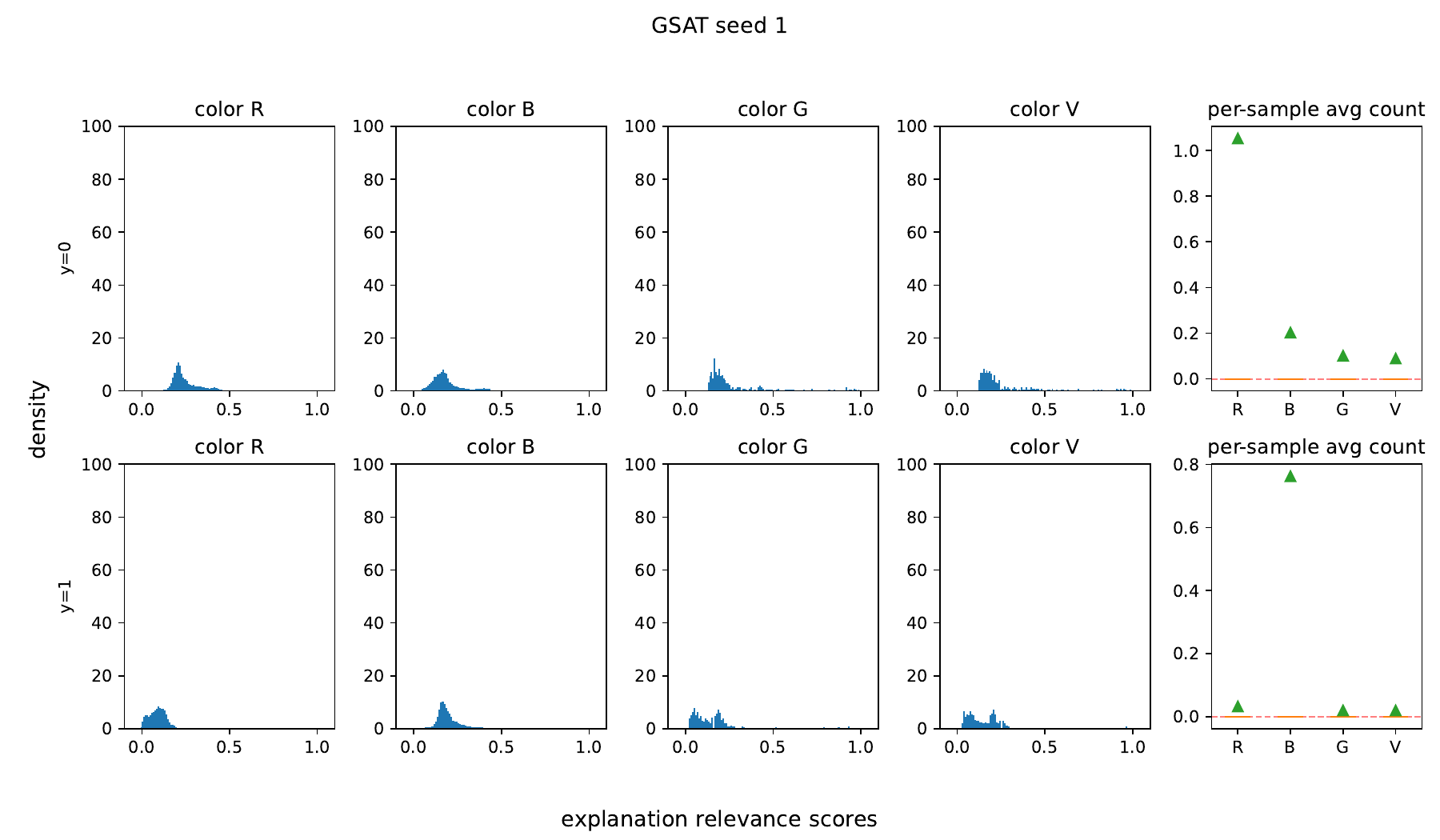}
    \caption{
        Diagram representing the distribution of explanation relevance score for seed 1 of \GSAT trained on \BAColorGV according to the setup described in \cref{sec:naturaldeg}.
        The four pairs of plots on the left report the distribution of relevance scores, divided by class label and node color (Red, Blue, Green, and Violet), while the last pair on the right reports the boxplot (colored bars) and average count (green triangle) of each color present in the explanation of each sample. 
        Nodes included in the explanation are chosen as those with a score greater than $0.5$.
        Overall, \GSAT roughly picks a single red node for $y=0$ and a single blue node for $y=1$, corresponding to the smallest label-preserving explanations.
        When computing rejection ratios for faithfulness metrics, we respectively get a rejection ratio of $58.7\%$ for \SUFFCAUSE, $0.9\%$ for \FIDM, and $0\%$ for \RFIDM.
        %
        Although red and blue nodes seem a legitimate explanation for this task, highlighting only one of them is nonetheless still concealing the fact that the model needs to look at all of them to make a correct prediction.
        Therefore, the low rejection ratios of \FIDM and \RFIDM are considered to be failure cases, as they flagged explanations not containing every element the model is actually using to infer the prediction as faithful.
        \SUFFCAUSE, instead, correctly rejects a considerable number of explanations.
    }
    \label{fig:naturaldeg-BAColorGV-GSAT}
\end{figure}

\begin{figure}[]
    \centering
    \includegraphics[width=0.99\linewidth]{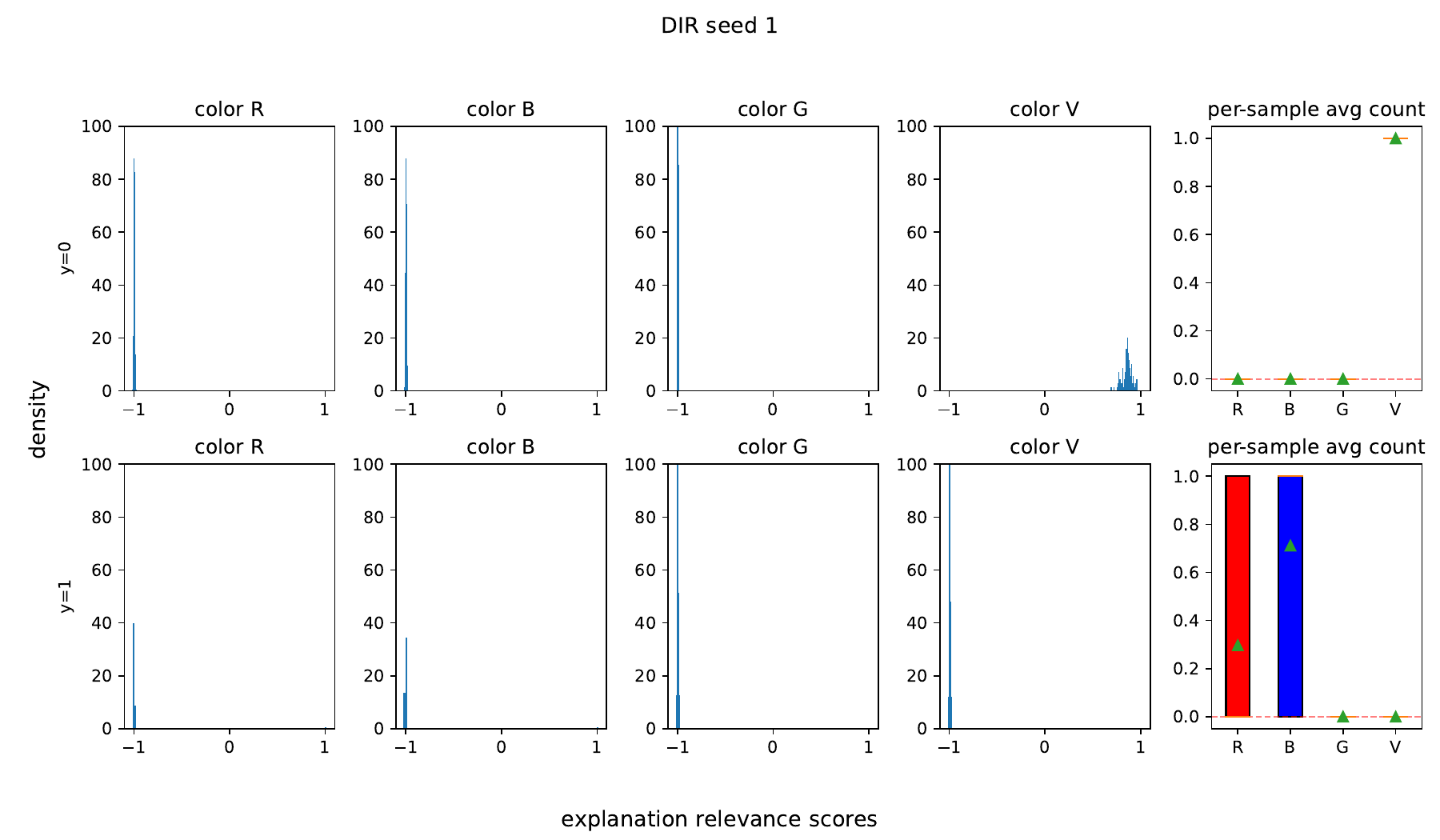}
    \caption{
        Diagram representing the distribution of explanation relevance score for seed 1 of \DIR trained on \BAColorGV according to the setup described in \cref{sec:naturaldeg}.
        The four pairs of plots on the left report the distribution of relevance scores, divided by class label and node color (Red, Blue, Green, and Violet), while the last pair on the right reports the boxplot (colored bars) and average count (green triangle) of each color present in the explanation of each sample. 
        Nodes included in the explanation are chosen as those having the top-$1\%$ relevance scores. 
        Nodes not belonging to the top-$1\%$ are assigned a score of $-1$ in the plot.
        Overall, even with natural training, \DIR picks the violet node as the explanation for $y=0$ -- similarly to the malicious explanations defined in \cref{sec:segnns-can-be-manipulated} -- whereas for $y=1$ the model is roughly selecting a single blue node.
        When computing rejection ratios for faithfulness metrics, we respectively get a rejection ratio of $47\%$ for \SUFFCAUSE, $0\%$ for \FIDM, and $22\%$ for \RFIDM.
        Those values highlight a catastrophic failure of \FIDM, which is reflected by the high standard deviation in \cref{tab:natural-degeneracy}.
    }
    \label{fig:naturaldeg-BAColorGV-DIR}
\end{figure}

\begin{figure}[]
    \centering
    \includegraphics[width=0.99\linewidth]{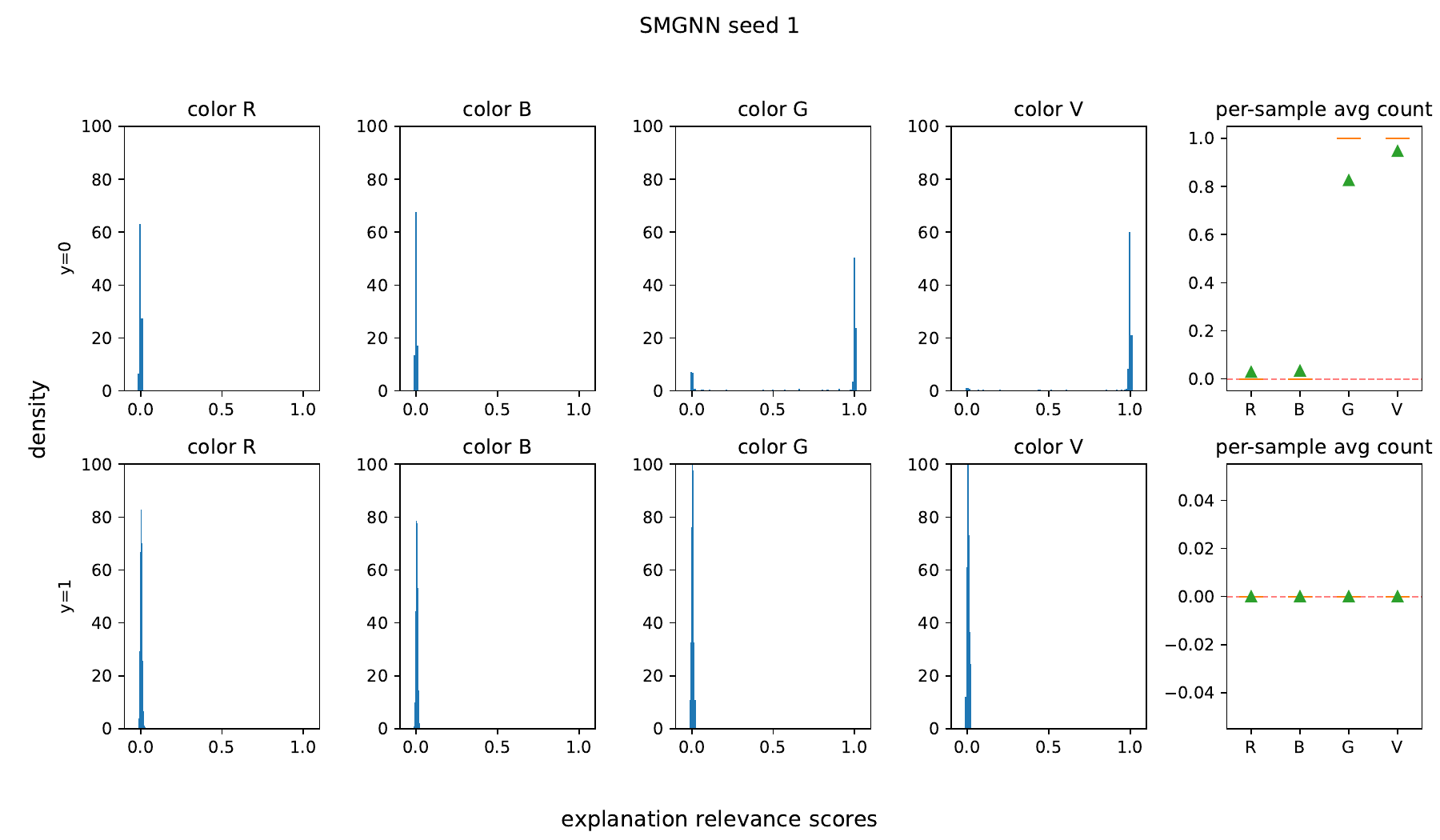}
    \caption{
        Diagram representing the distribution of explanation relevance score for seed 1 of \SMGNN trained on \BAColorGV according to the setup described in \cref{sec:naturaldeg}.
        The four pairs of plots on the left report the distribution of relevance scores, divided by class label and node color (Red, Blue, Green, and Violet), while the last pair on the right reports the boxplot (colored bars) and average count (green triangle) of each color present in the explanation of each sample. 
        Nodes included in the explanation are chosen as those with a score greater than $0.5$.
        Overall, even with natural training, \SMGNN picks green and violet nodes as the explanation for $y=0$ -- similarly to the malicious explanations defined in \cref{sec:segnns-can-be-manipulated} -- whereas for $y=1$ the model is assigning zero relevance to every node.
        We then compute rejection ratios only for the negative class, getting a value of $99\%$ for \SUFFCAUSE, $0\%$ for \FIDM, and $98\%$ for \RFIDM.
        Those values highlight a catastrophic failure of \FIDM, which is reflected by the high standard deviation in \cref{tab:natural-degeneracy}.
    }
    \label{fig:naturaldeg-BAColorGV-SMGNN}
\end{figure}

\begin{figure}[]
    \centering
    \begin{subfigure}{0.45\linewidth}
        \centering
        \includegraphics[width=\linewidth]{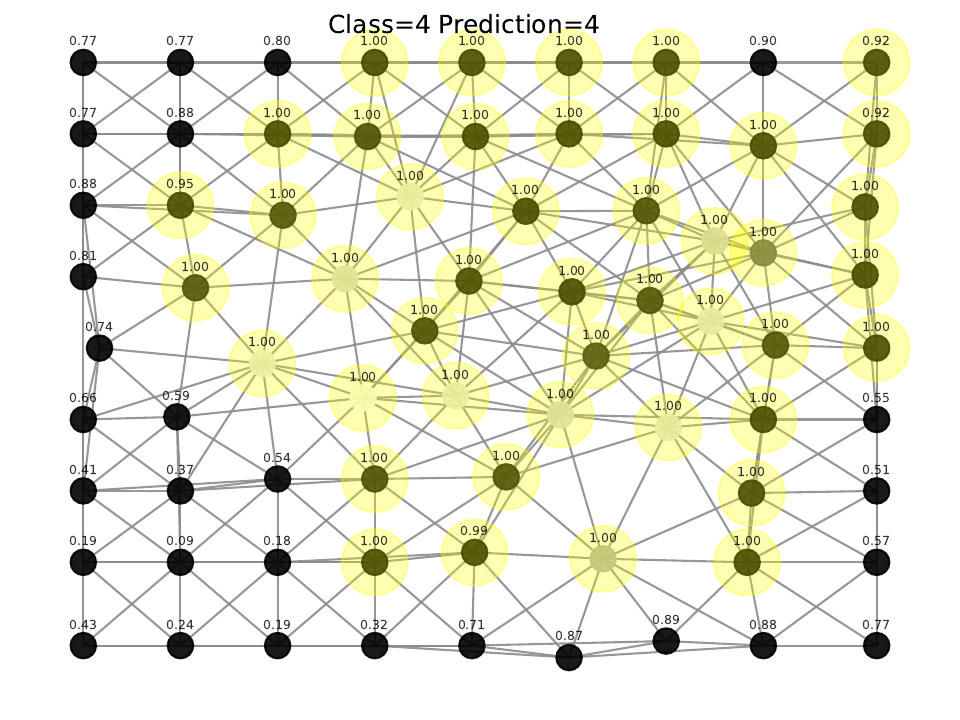}
        \caption{$\SUFFCAUSE=0$, $\FIDM=0$, $\RFIDM=0$}
    \end{subfigure}
    \begin{subfigure}{0.45\linewidth}
        \centering
        \includegraphics[width=\linewidth]{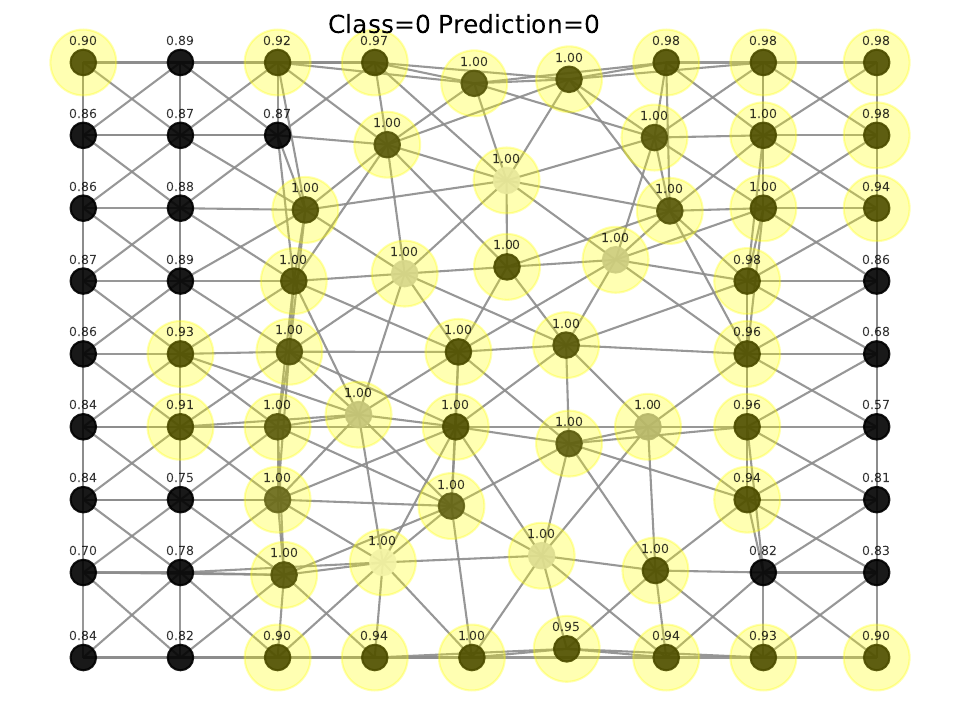}
        \caption{$\SUFFCAUSE=0$, $\FIDM=0$, $\RFIDM=0$}
    \end{subfigure}
    \begin{subfigure}{0.45\linewidth}
        \centering
        \includegraphics[width=\linewidth]{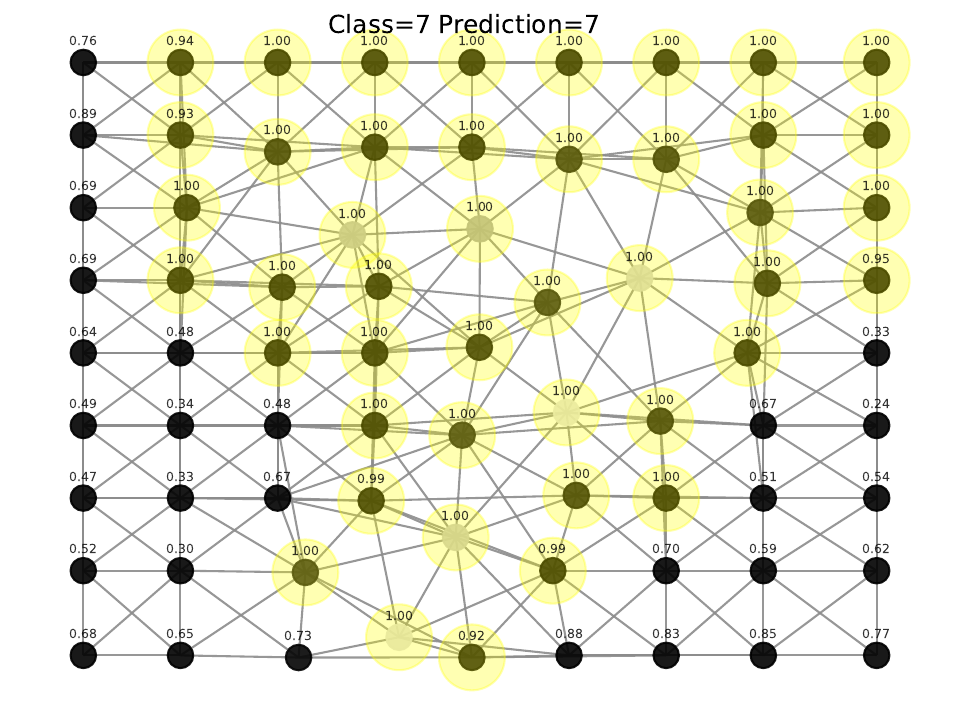}
        \caption{$\SUFFCAUSE=0$, $\FIDM=0$, $\RFIDM=0$}
    \end{subfigure}
    \begin{subfigure}{0.45\linewidth}
        \centering
        \includegraphics[width=\linewidth]{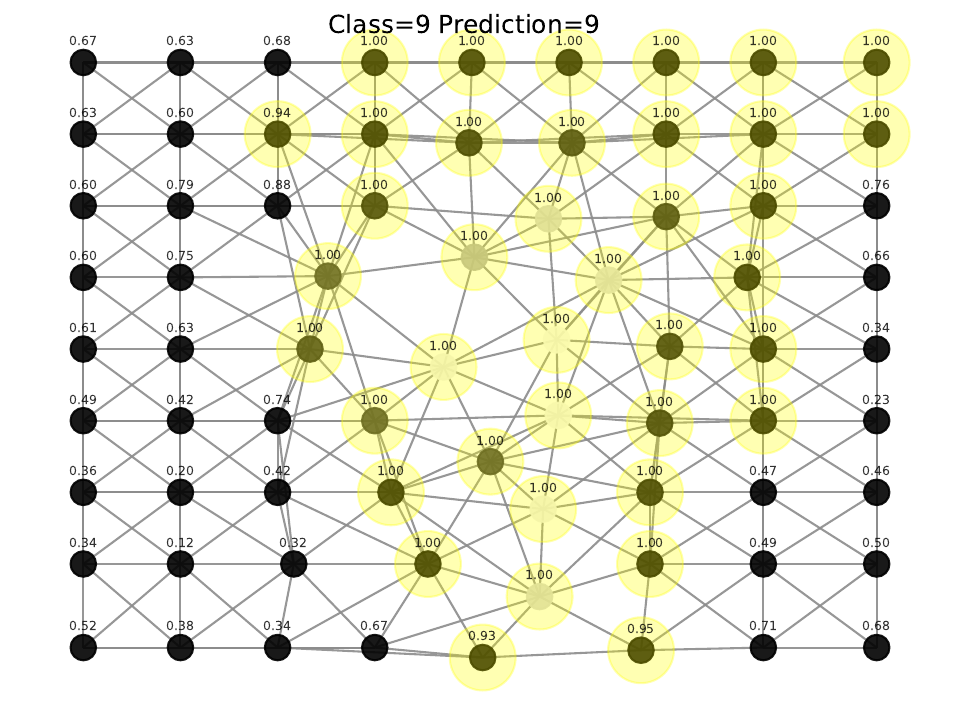}
        \caption{$\SUFFCAUSE=1$, $\FIDM=0$, $\RFIDM=1$}
    \end{subfigure}
    \begin{subfigure}{0.45\linewidth}
        \centering
        \includegraphics[width=\linewidth]{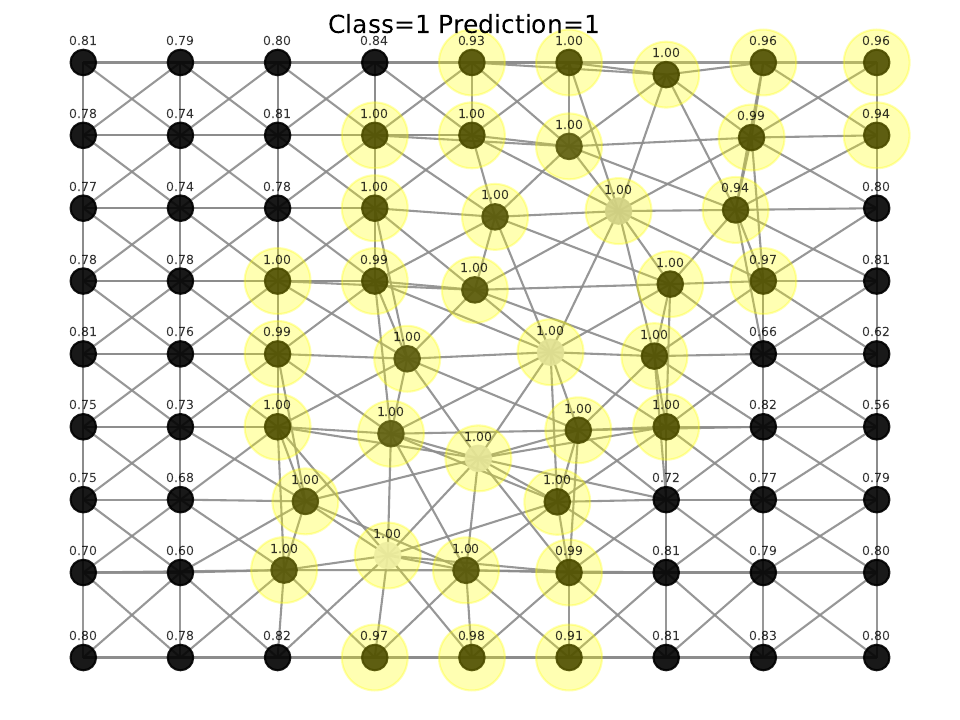}
        \caption{$\SUFFCAUSE=0$, $\FIDM=0$, $\RFIDM=1$}
    \end{subfigure}
    \begin{subfigure}{0.45\linewidth}
        \centering
        \includegraphics[width=\linewidth]{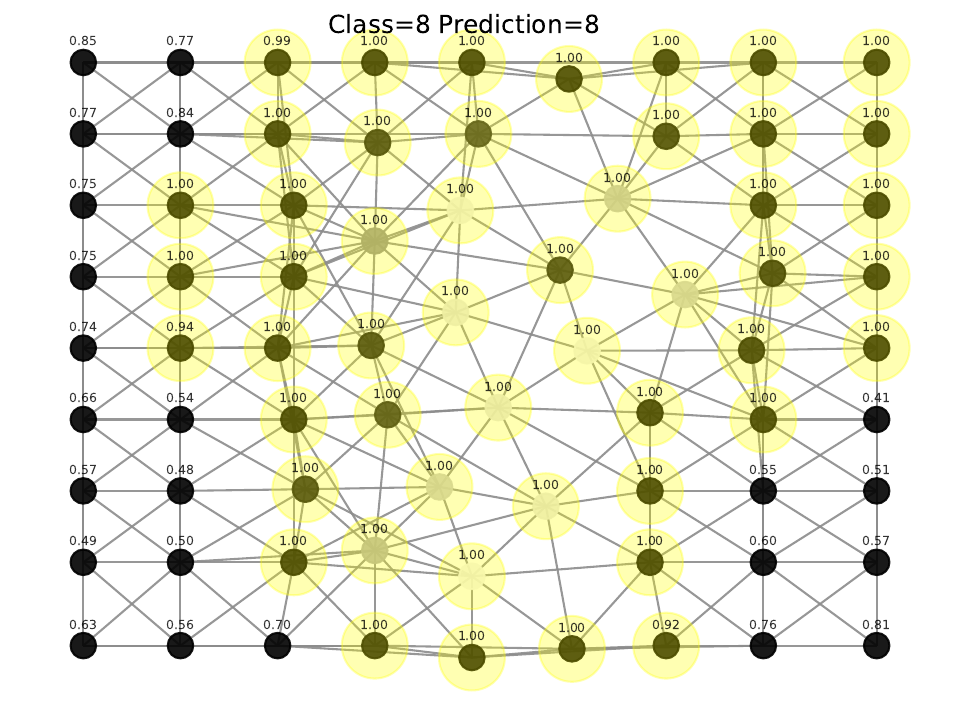}
        \caption{$\SUFFCAUSE=0$, $\FIDM=0$, $\RFIDM=0$}
    \end{subfigure}
    \caption{
        Examples of explanations for seed 4 of \GSAT trained on \MNIST according to the setup described in \cref{sec:naturaldeg}.
        Nodes included in the explanation are chosen as those with a score greater than $0.9$.
        Overall, \GSAT extracts explanations highlighting the whole digit with the addition of some background pixels.
        Therefore, we expect explanations to contain most of the information the model can use to make correct predictions.
        In fact, the rejection ratios reported below each example confirm that, in most cases, little to no relevant information is left in the complement.
        In particular, we report the value $1$ whenever each metric rejects the explanation, and $0$ when not.
    }
    \label{fig:naturaldeg-MNIST-GSAT}
\end{figure}

\begin{figure}[]
    \centering
    \begin{subfigure}{0.45\linewidth}
        \centering
        \includegraphics[width=\linewidth]{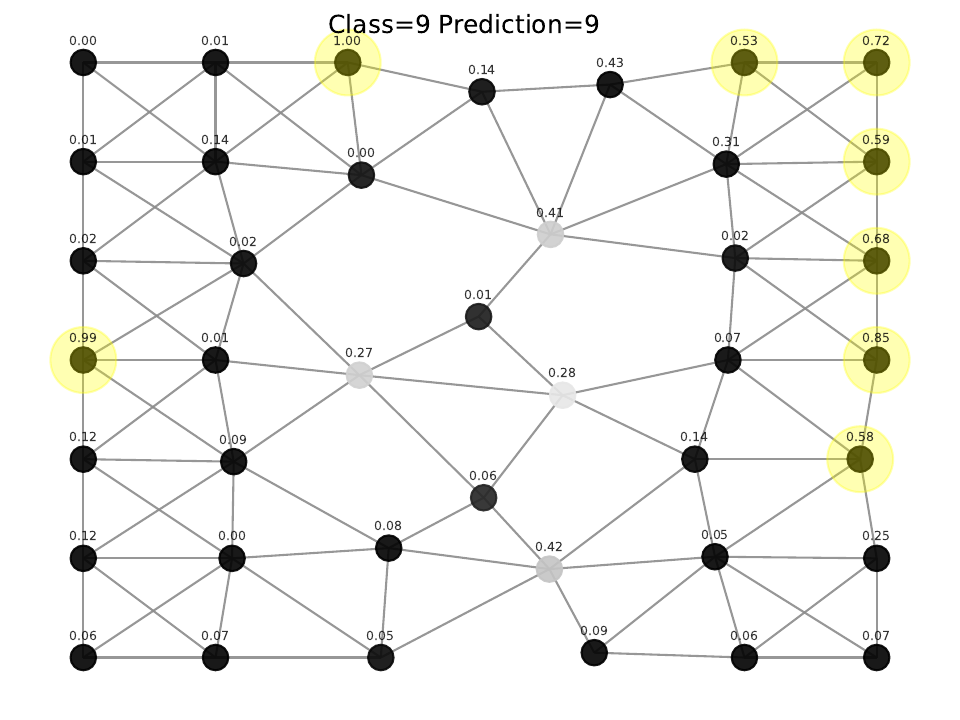}
        \caption{$\SUFFCAUSE=1$, $\FIDM=1$, $\RFIDM=1$}
    \end{subfigure}
    \begin{subfigure}{0.45\linewidth}
        \centering
        \includegraphics[width=\linewidth]{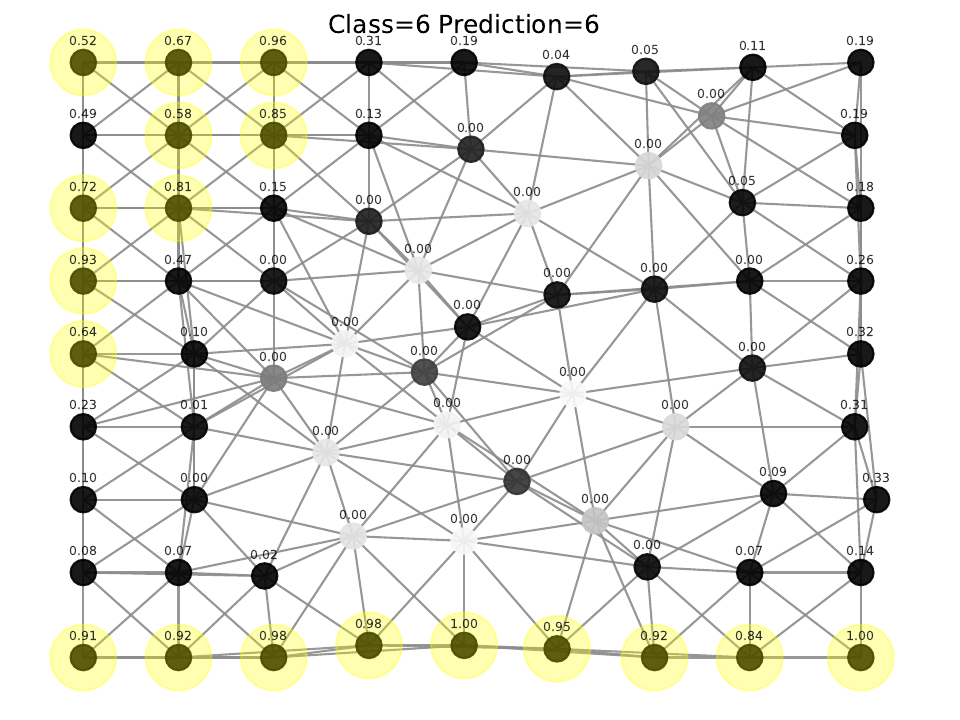}
        \caption{$\SUFFCAUSE=1$, $\FIDM=0$, $\RFIDM=1$}
    \end{subfigure}
    \begin{subfigure}{0.45\linewidth}
        \centering
        \includegraphics[width=\linewidth]{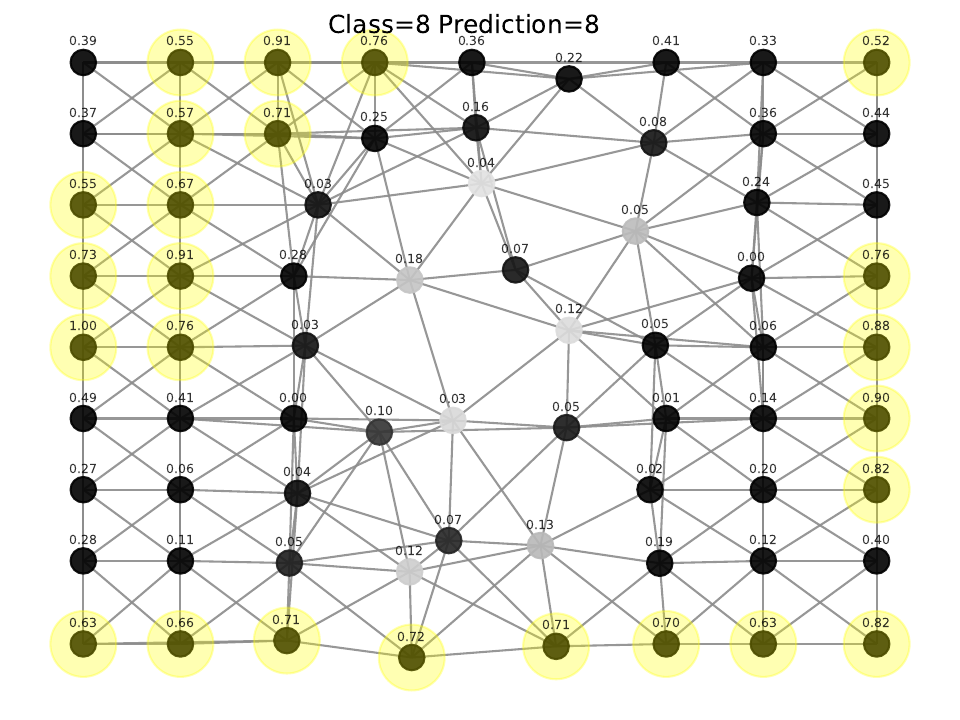}
        \caption{$\SUFFCAUSE=1$, $\FIDM=1$, $\RFIDM=1$}
    \end{subfigure}
    \begin{subfigure}{0.45\linewidth}
        \centering
        \includegraphics[width=\linewidth]{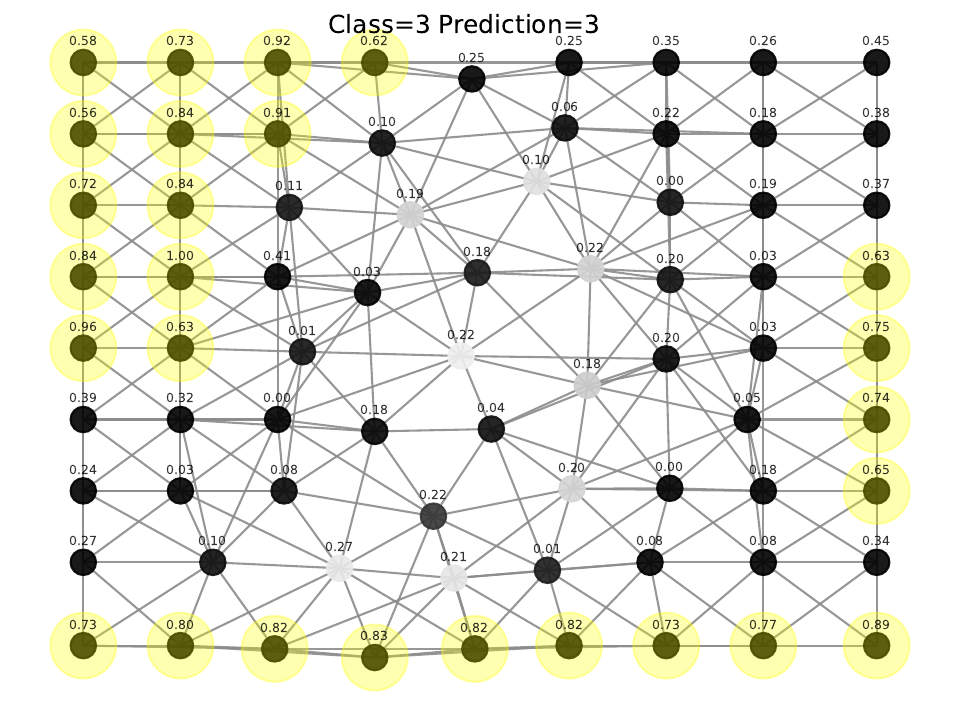}
        \caption{$\SUFFCAUSE=1$, $\FIDM=1$, $\RFIDM=1$}
    \end{subfigure}
    \begin{subfigure}{0.45\linewidth}
        \centering
        \includegraphics[width=\linewidth]{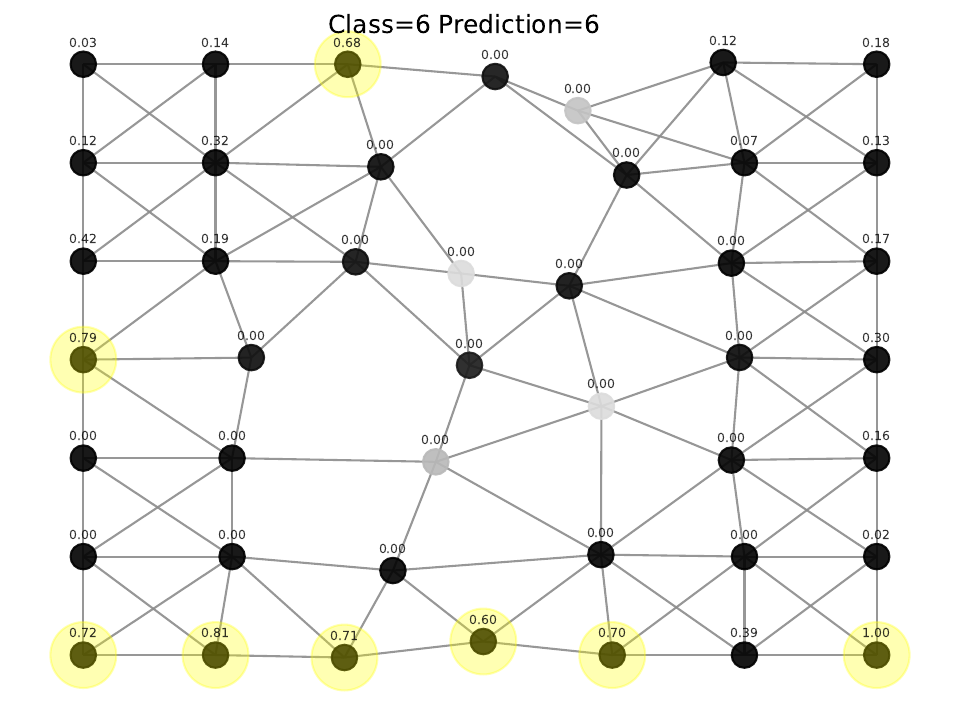}
        \caption{$\SUFFCAUSE=1$, $\FIDM=0$, $\RFIDM=1$}
    \end{subfigure}
    \begin{subfigure}{0.45\linewidth}
        \centering
        \includegraphics[width=\linewidth]{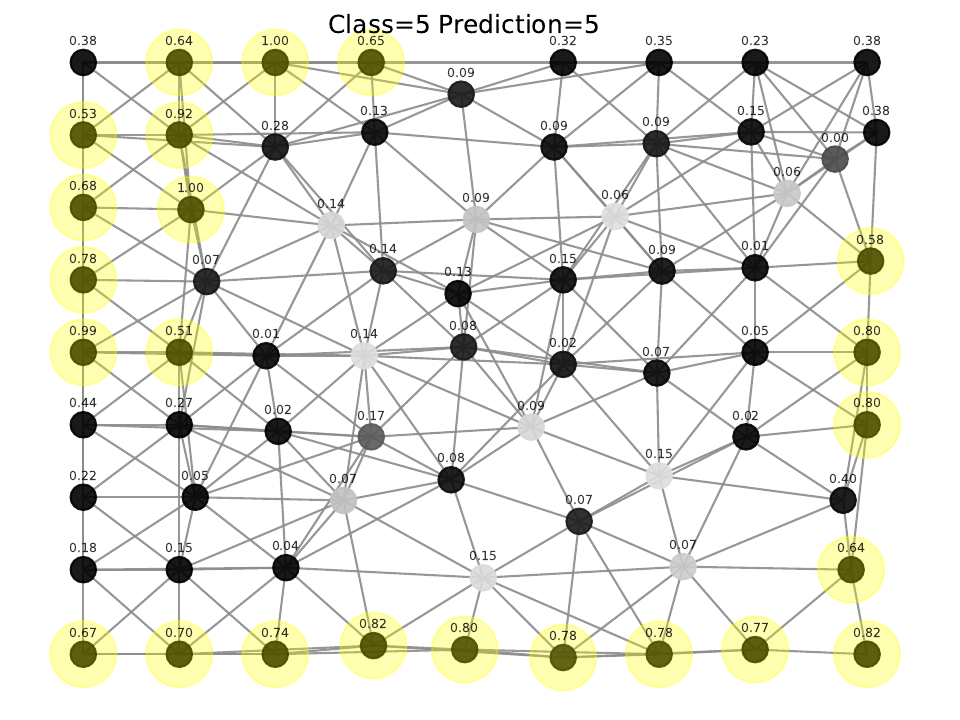}
        \caption{$\SUFFCAUSE=1$, $\FIDM=1$, $\RFIDM=1$}
    \end{subfigure}
    \caption{
        Examples of explanations for seed 2 of \SMGNN trained on \MNIST according to the setup described in \cref{sec:naturaldeg}.
        Node relevance scores are min-max normalized in the range $[0,1]$ to ensure a meaningful thresholding across seeds,  and the nodes included in the explanation are chosen as those with a score greater than $0.5$.
        Overall, \SMGNN extracts explanations highlighting background pixels, hinting to the fact that the model could be extracting Degenerate explanations.
        In fact, the rejection ratios computed for each example confirm that explanations are marked as unfaithful in most cases.
        The only exception is \FIDM, which marks them as faithful in $2$ cases.
        Recall that a rejection ratio of $1$ in the plots above means the explanation is rejected, hence considered unfaithful by the metric.
    }
    \label{fig:naturaldeg-MNIST-SMGNN}
\end{figure}

\begin{figure}[]
    \centering
    \begin{subfigure}{0.45\linewidth}
        \centering
        \includegraphics[width=\linewidth]{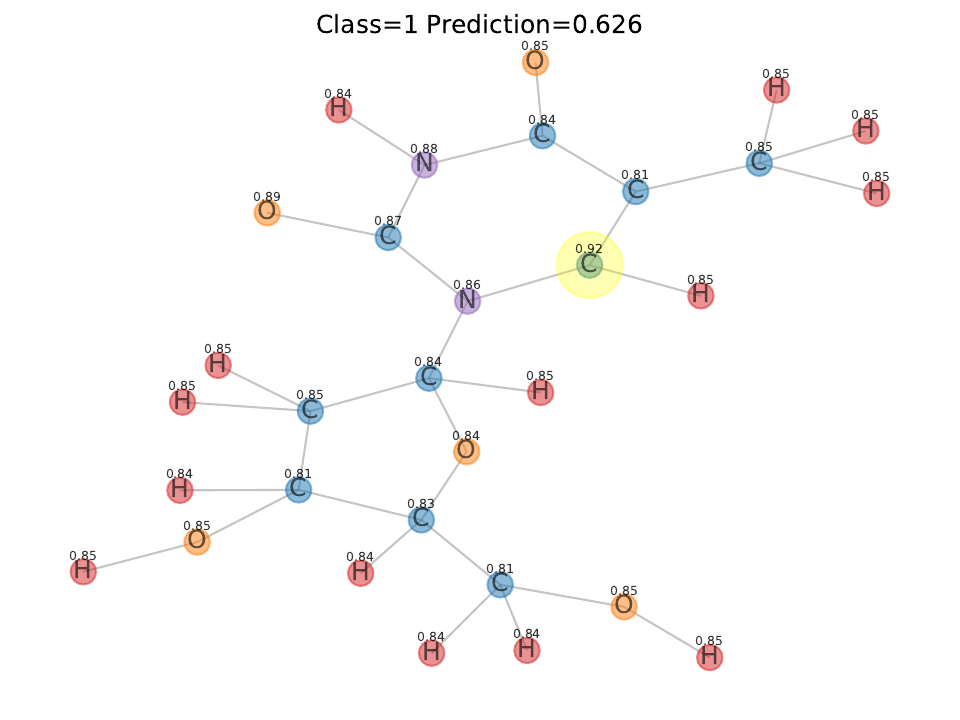}
        \caption{$\SUFFCAUSE=1$, $\FIDM=1$, $\RFIDM=1$}
    \end{subfigure}
    \begin{subfigure}{0.45\linewidth}
        \centering
        \includegraphics[width=\linewidth]{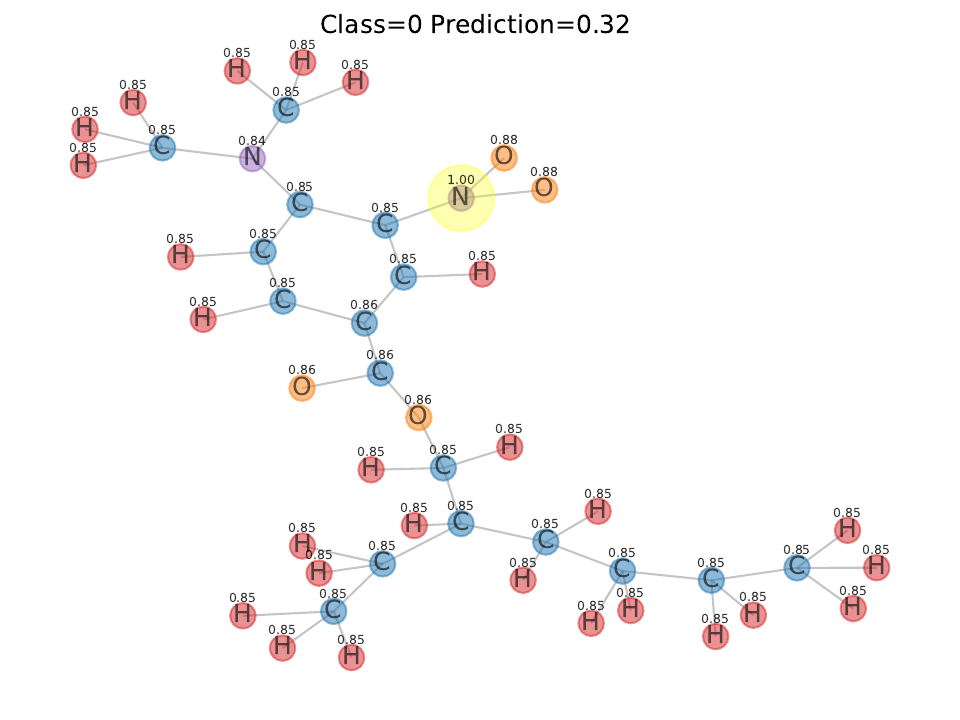}
        \caption{$\SUFFCAUSE=1$, $\FIDM=1$, $\RFIDM=1$}
    \end{subfigure}
    \begin{subfigure}{0.45\linewidth}
        \centering
        \includegraphics[width=\linewidth]{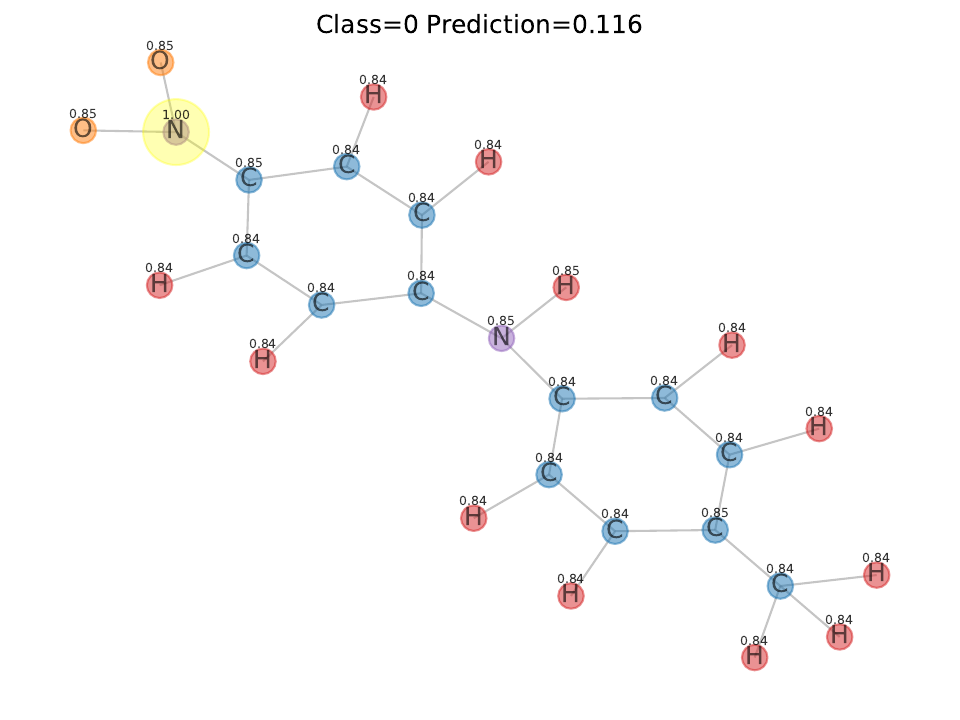}
        \caption{$\SUFFCAUSE=1$, $\FIDM=1$, $\RFIDM=1$}
    \end{subfigure}
    \begin{subfigure}{0.45\linewidth}
        \centering
        \includegraphics[width=\linewidth]{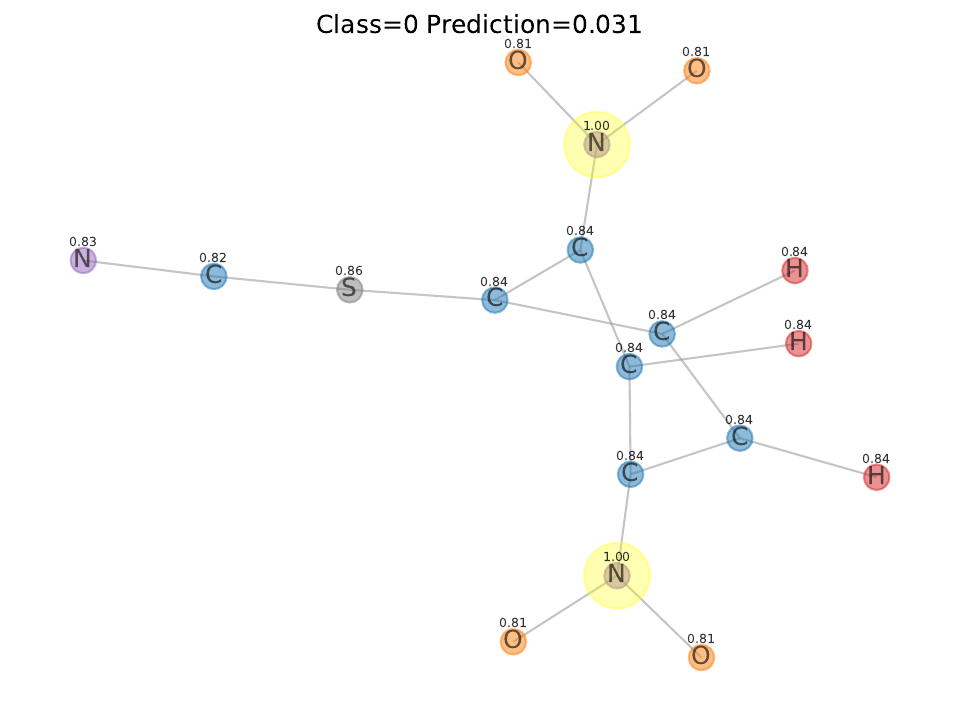}
        \caption{$\SUFFCAUSE=1$, $\FIDM=1$, $\RFIDM=1$}
    \end{subfigure}
    \begin{subfigure}{0.45\linewidth}
        \centering
        \includegraphics[width=\linewidth]{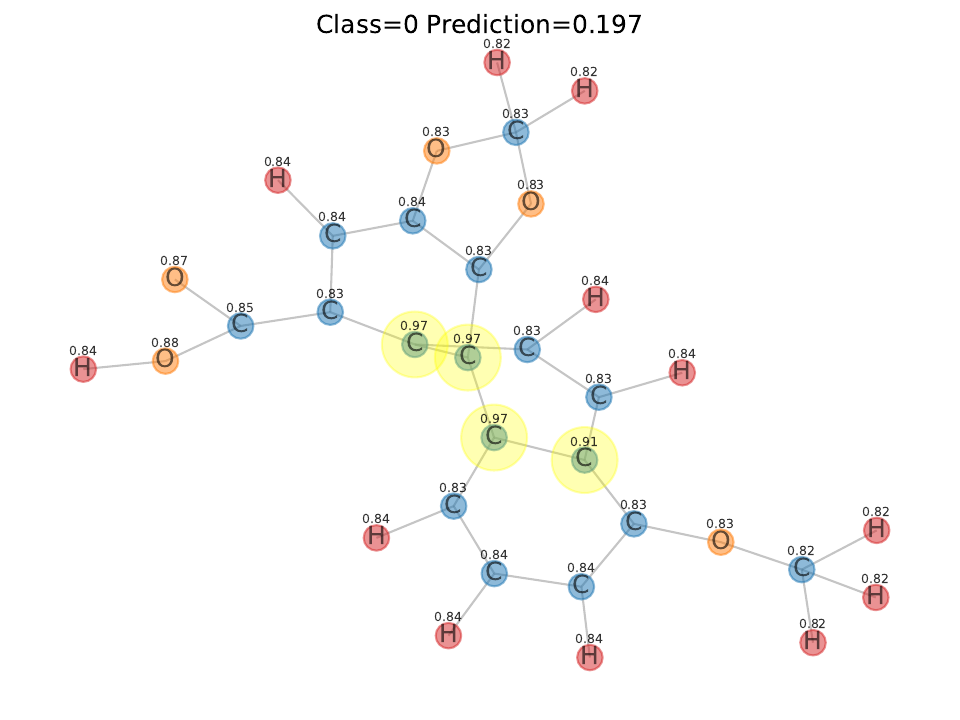}
        \caption{$\SUFFCAUSE=1$, $\FIDM=1$, $\RFIDM=1$}
    \end{subfigure}
    \begin{subfigure}{0.45\linewidth}
        \centering
        \includegraphics[width=\linewidth]{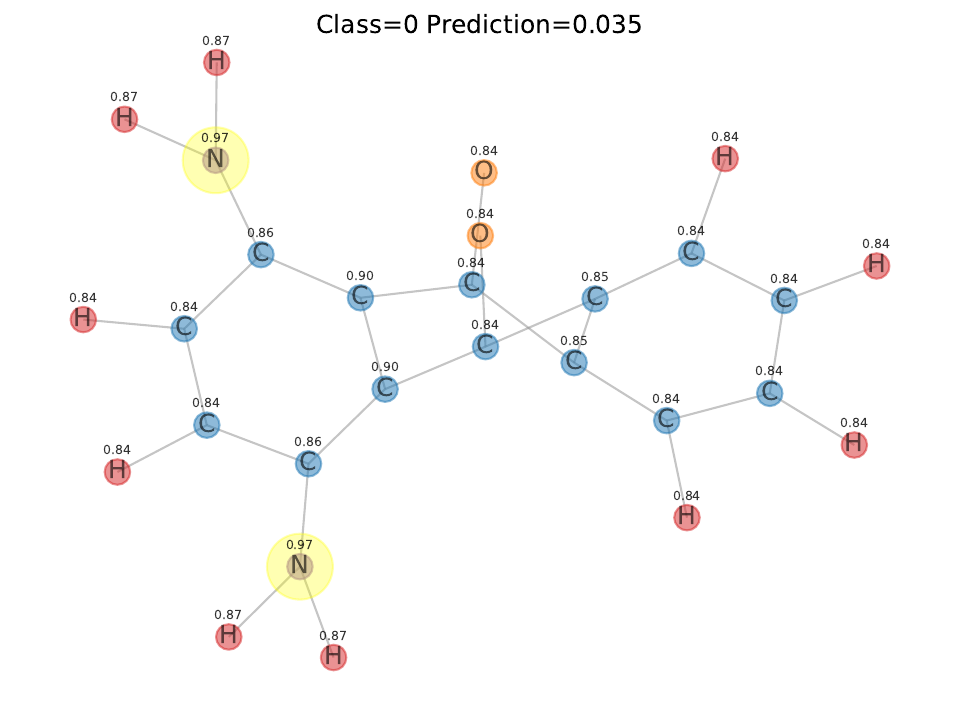}
        \caption{$\SUFFCAUSE=1$, $\FIDM=1$, $\RFIDM=1$}
    \end{subfigure}
    \caption{
        Examples of explanations for seed 1 of \GSAT trained on \MUTAG according to the setup described in \cref{sec:naturaldeg}.
        Nodes included in the explanation are chosen as those with a score greater than $0.9$.
        Overall, although it is hard to judge whether explanations are faithful or not by visual inspection in this case -- as we do not have full knowledge about the task and scores are not maximally sparse -- we can still make the following observations:
        First, explanations tend to highlight atoms that are known to be uninformative of the target label when taken alone, like C and N (cf. \cref{fig:hist-MUTAG});
        Second, rejection ratios for all metrics are non-negligible (see \cref{tab:natural-degeneracy}), outlining a substantial agreement in judging these explanations as unfaithful.
        Recall that a rejection ratio of $1$ in the plots above means the explanation is rejected, hence considered unfaithful by the metric.
    }
    \label{fig:naturaldeg-MUTAG-GSAT}
\end{figure}

\begin{figure}[]
    \centering
    \begin{subfigure}{0.45\linewidth}
        \centering
        \includegraphics[width=\linewidth]{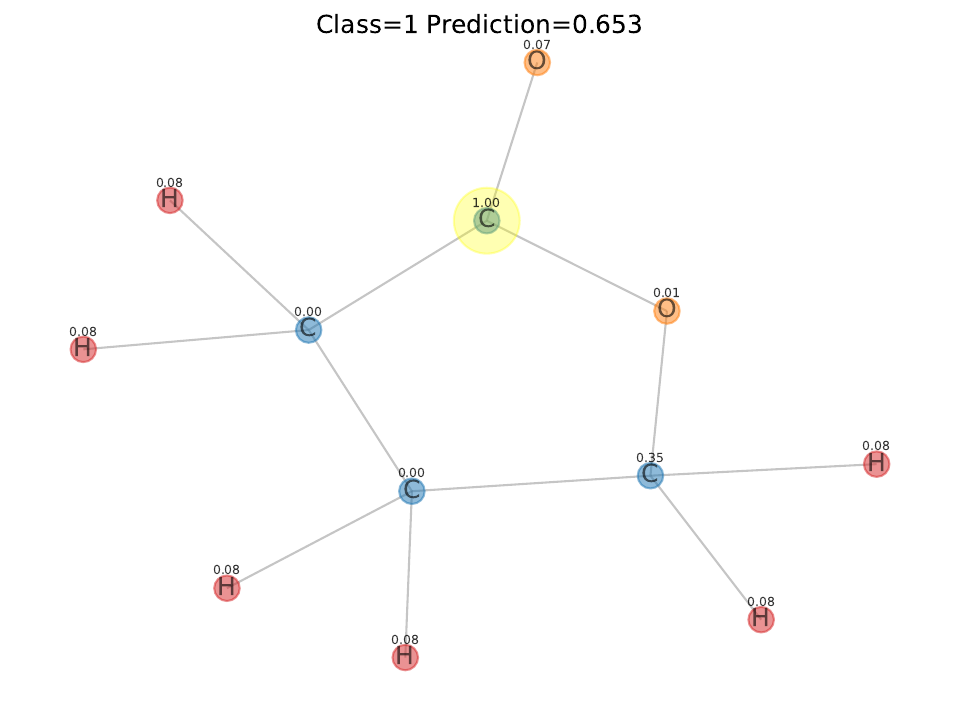}
        \caption{$\SUFFCAUSE=1$, $\FIDM=1$, $\RFIDM=1$}
        \label{subfig:MUTAG-SMGNN-a}
    \end{subfigure}
    \begin{subfigure}{0.45\linewidth}
        \centering
        \includegraphics[width=\linewidth]{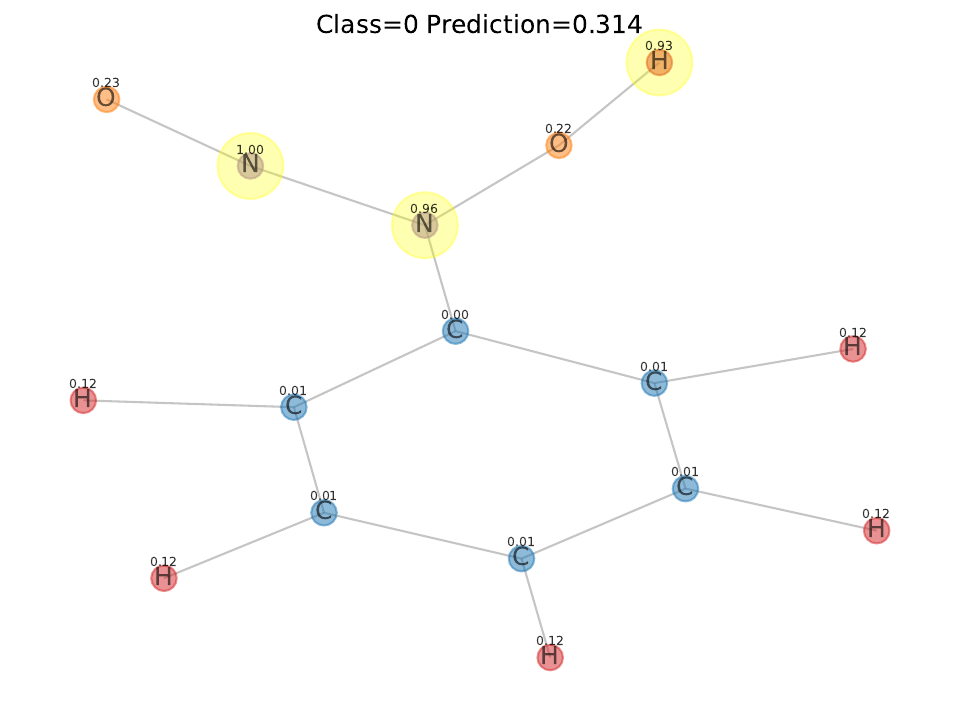}
        \caption{$\SUFFCAUSE=1$, $\FIDM=1$, $\RFIDM=1$}
    \end{subfigure}
    \begin{subfigure}{0.45\linewidth}
        \centering
        \includegraphics[width=\linewidth]{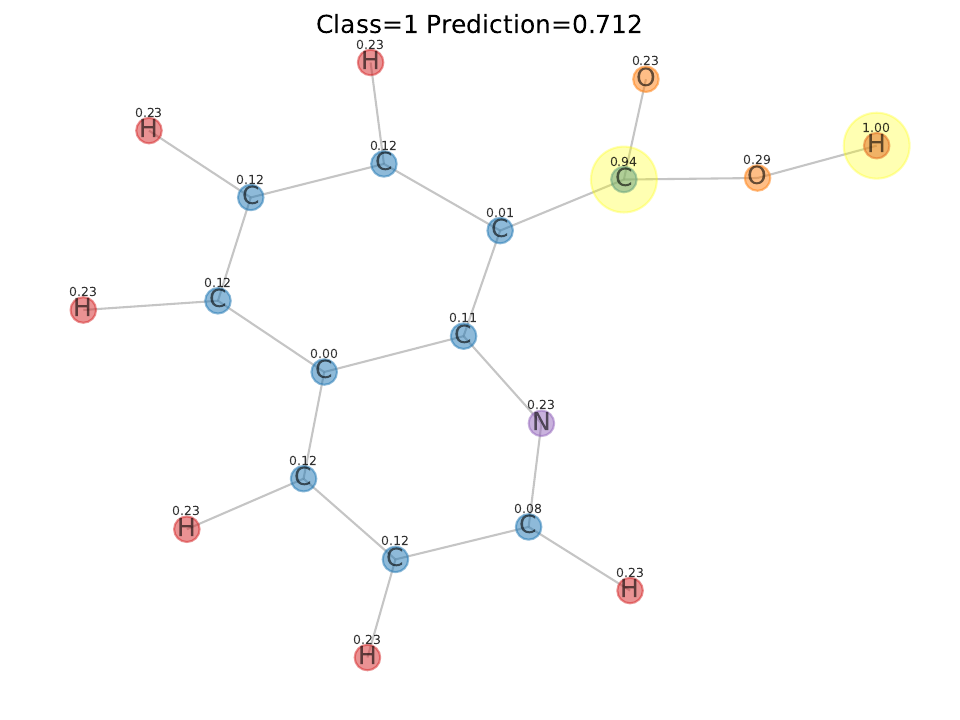}
        \caption{$\SUFFCAUSE=1$, $\FIDM=1$, $\RFIDM=1$}
    \end{subfigure}
    \begin{subfigure}{0.45\linewidth}
        \centering
        \includegraphics[width=\linewidth]{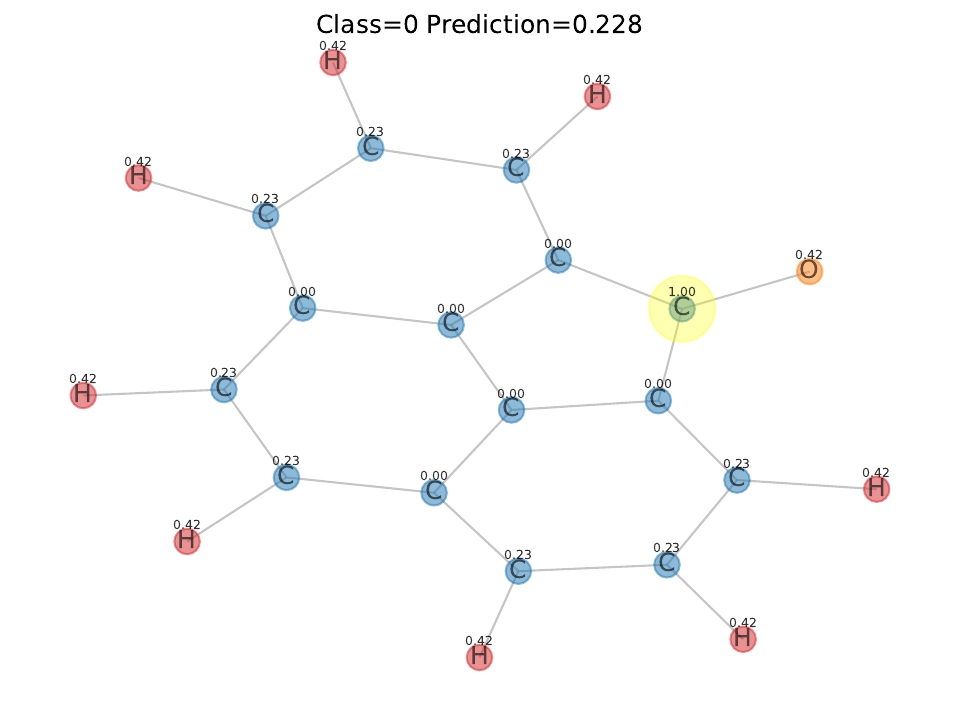}
        \caption{$\SUFFCAUSE=1$, $\FIDM=1$, $\RFIDM=1$}
        \label{subfig:MUTAG-SMGNN-d}
    \end{subfigure}
    \begin{subfigure}{0.45\linewidth}
        \centering
        \includegraphics[width=\linewidth]{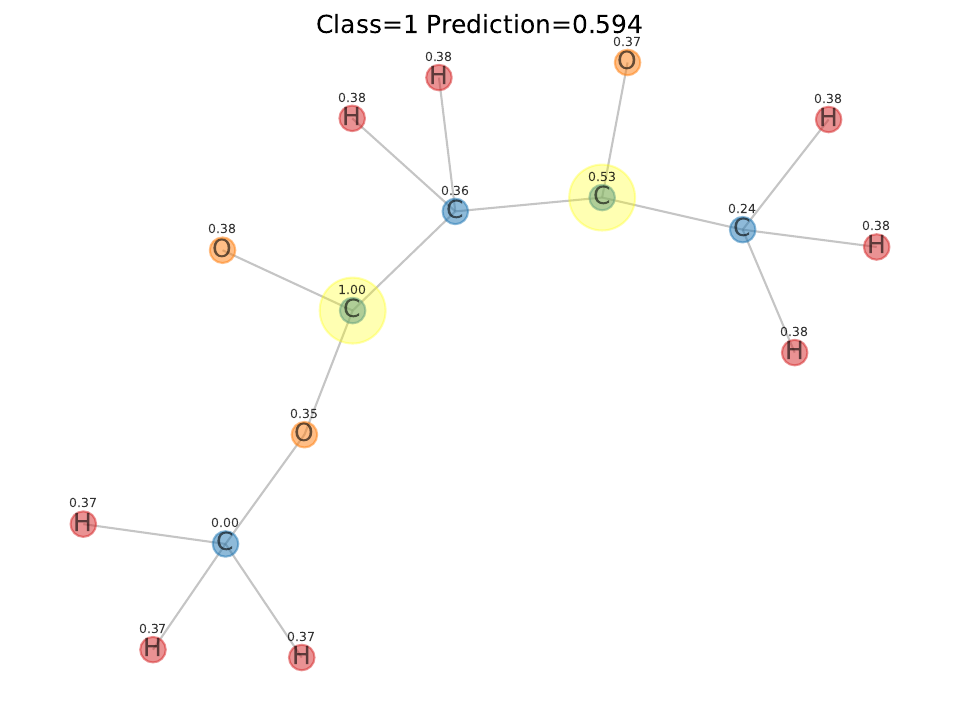}
        \caption{$\SUFFCAUSE=1$, $\FIDM=0$, $\RFIDM=1$}
    \end{subfigure}
    \begin{subfigure}{0.45\linewidth}
        \centering
        \includegraphics[width=\linewidth]{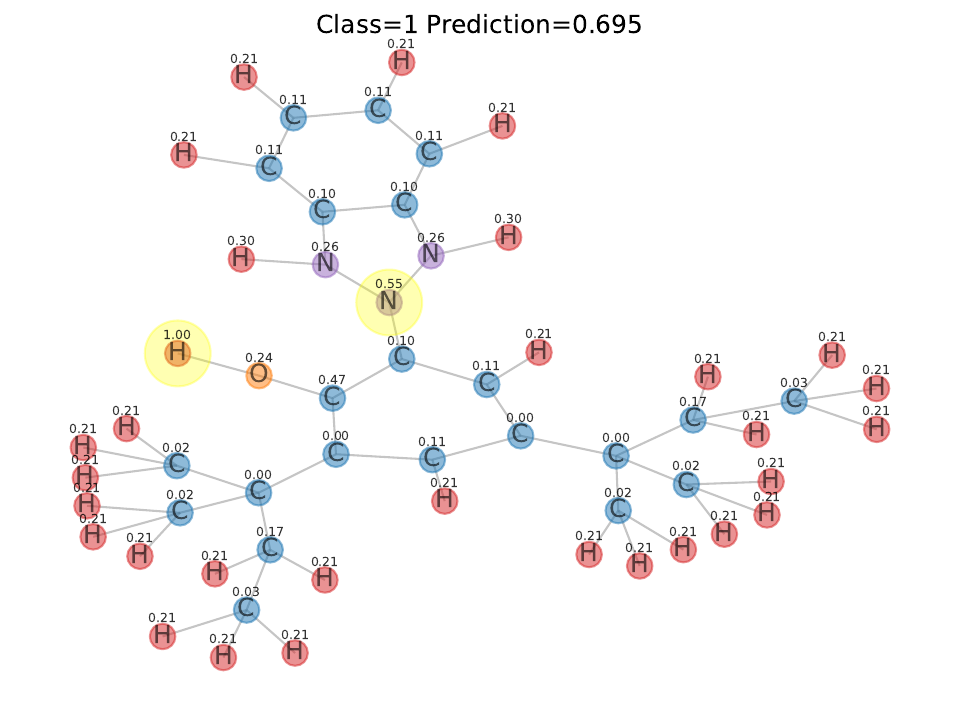}
        \caption{$\SUFFCAUSE=1$, $\FIDM=1$, $\RFIDM=0$}
    \end{subfigure}
    \caption{
        Examples of explanations for seed 3 of \SMGNN trained on \MUTAG according to the setup described in \cref{sec:naturaldeg}.
        Node relevance scores are min-max normalized in the range $[0,1]$ to ensure a meaningful thresholding across seeds,  and the nodes included in the explanation are chosen as those with a score greater than $0.5$.
        Overall, although it is hard to judge whether explanations are faithful or not by visual inspection in this case -- as we do not have full knowledge about the task and scores are not maximally sparse -- we can still make the following observations:
        First, explanations tend to highlight atoms that are known to be uninformative of the target label when taken alone, like C and H (cf. \cref{fig:hist-MUTAG});
        Second, the same explanation can appear for different classes, like (\subref{subfig:MUTAG-SMGNN-a}) and (\subref{subfig:MUTAG-SMGNN-d});
        Third, rejection ratios for all metrics are substantially high (see \cref{tab:natural-degeneracy}), outlining agreement in judging these explanations as unfaithful.
        Recall that a rejection ratio of $1$ in the plots above means the explanation is rejected, hence considered unfaithful by the metric.
    }
    \label{fig:naturaldeg-MUTAG-SMGNN}
\end{figure}

\begin{figure}[]
    \centering
    \begin{subfigure}{0.45\linewidth}
        \centering
        \includegraphics[width=\linewidth]{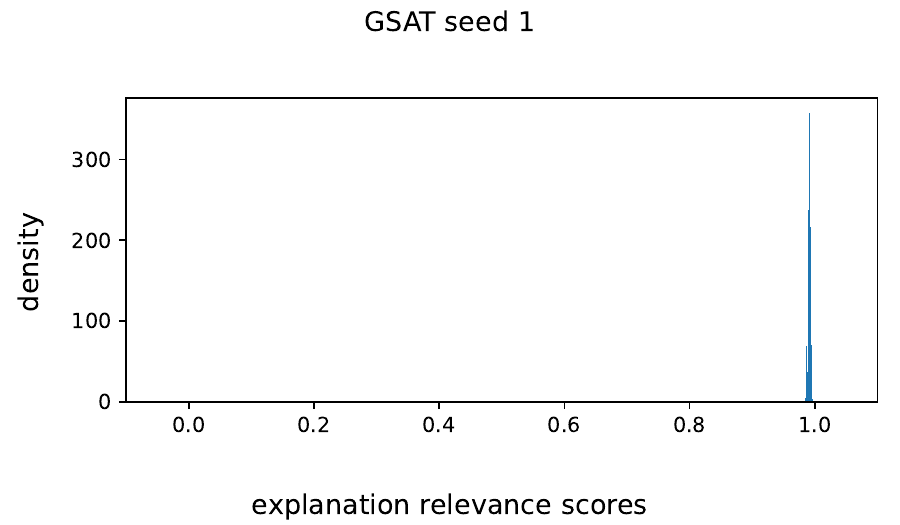}
    \end{subfigure}
    \begin{subfigure}{0.45\linewidth}
        \centering
        \includegraphics[width=\linewidth]{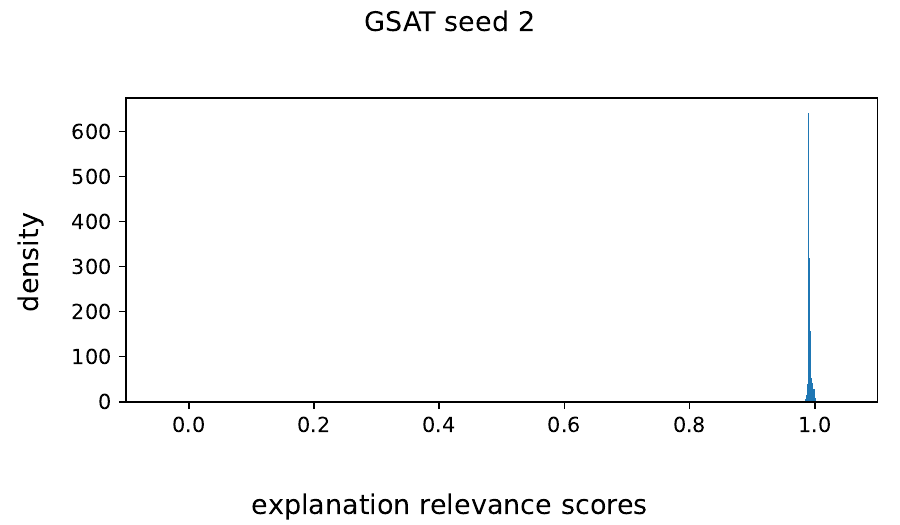}
    \end{subfigure}
    \begin{subfigure}{0.45\linewidth}
        \centering
        \includegraphics[width=\linewidth]{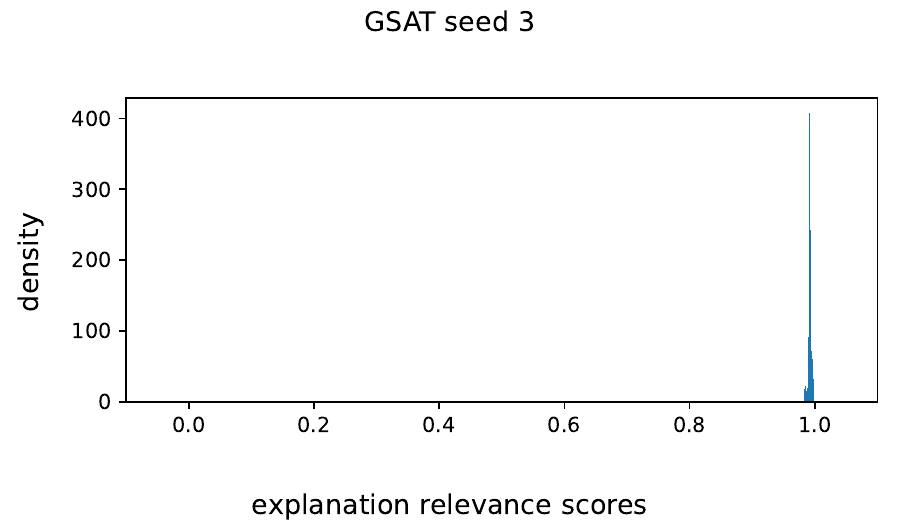}
    \end{subfigure}
    \begin{subfigure}{0.45\linewidth}
        \centering
        \includegraphics[width=\linewidth]{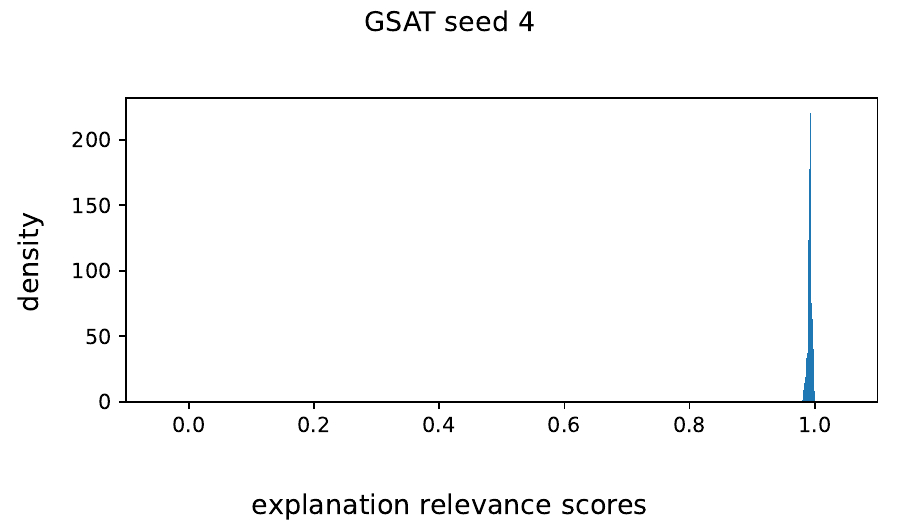}
    \end{subfigure}
    \begin{subfigure}{0.45\linewidth}
        \centering
        \includegraphics[width=\linewidth]{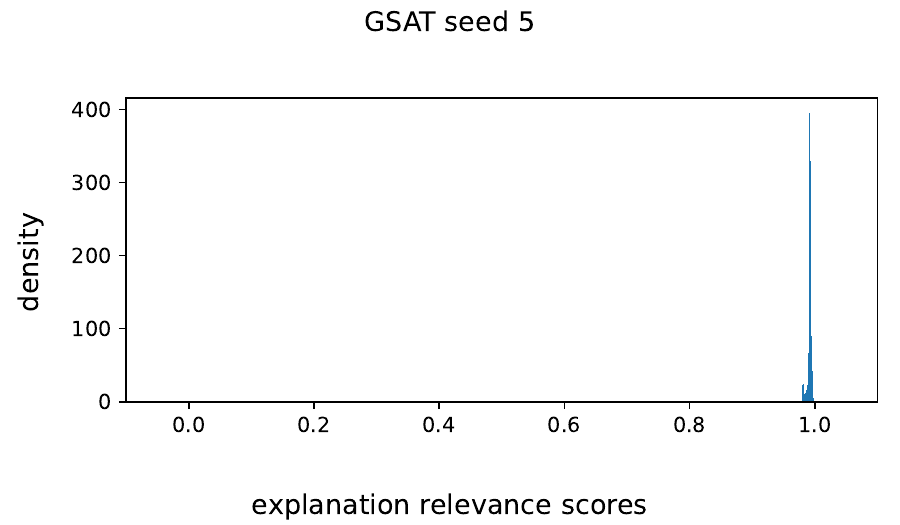}
    \end{subfigure}
    \caption{
        Histograms of explanations scores for different random seeds of \GSAT trained on \SSTP according to the setup described in \cref{sec:naturaldeg}.
        Overall, relevance scores are all squashed around $1$, meaning the model weights equally all nodes.
        Therefore, by thresholding explanations at $0.9$ results in explanations covering the full graph, which makes the explanation trivially \textit{sufficient}. 
        Hence, metrics achieve a rejection ratio of $0\%$.
    }
    \label{fig:naturaldeg-SSTP-GSAT}
\end{figure}

\begin{figure}[]
    \centering
    \begin{subfigure}{0.99\linewidth}
        \centering
        \includegraphics[width=\linewidth]{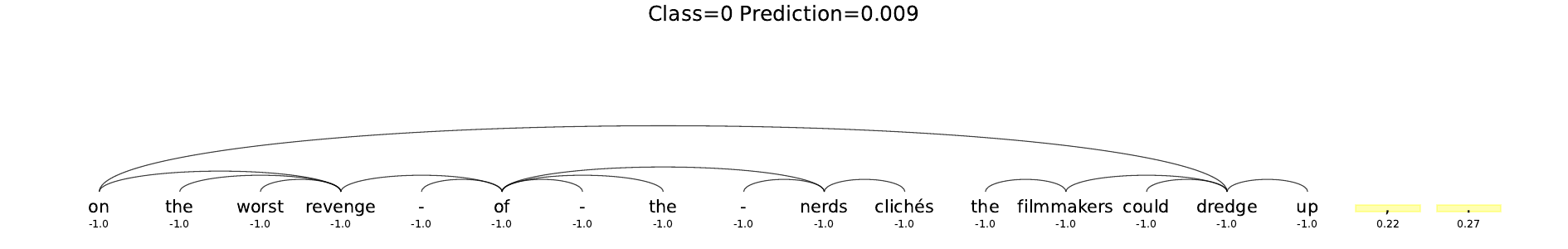}
        \caption{$\SUFFCAUSE=1$, $\FIDM=0$, $\RFIDM=0$}
    \end{subfigure}
    \begin{subfigure}{0.99\linewidth}
        \centering
        \includegraphics[width=\linewidth]{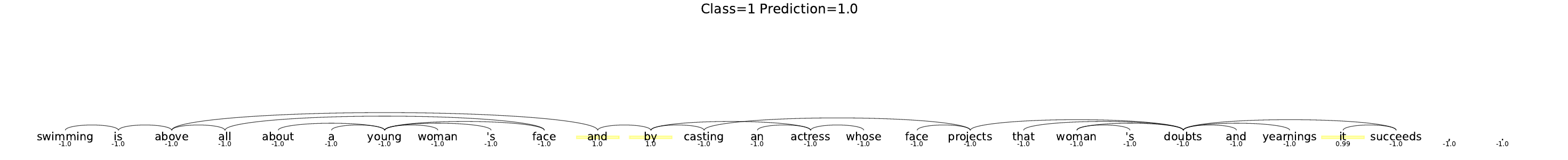}
        \caption{$\SUFFCAUSE=0$, $\FIDM=0$, $\RFIDM=0$}
    \end{subfigure}
    \begin{subfigure}{0.90\linewidth}
        \centering
        \includegraphics[width=\linewidth]{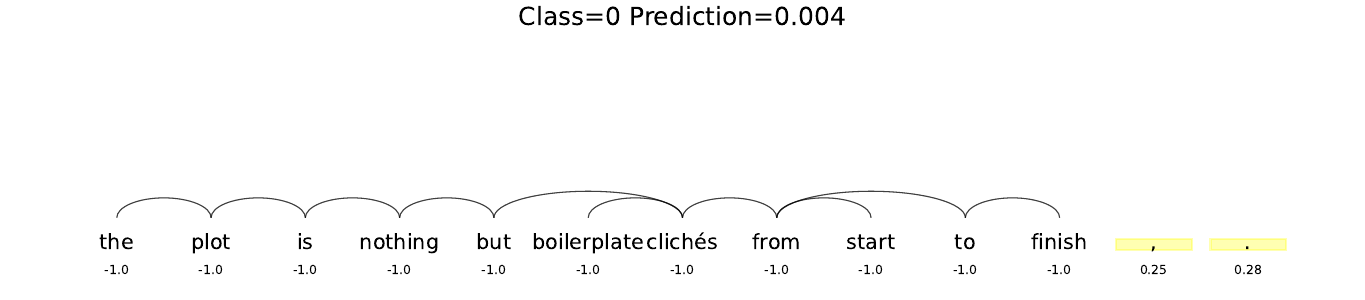}
        \caption{$\SUFFCAUSE=1$, $\FIDM=0$, $\RFIDM=0$}
    \end{subfigure}
    \begin{subfigure}{0.99\linewidth}
        \centering
        \includegraphics[width=\linewidth]{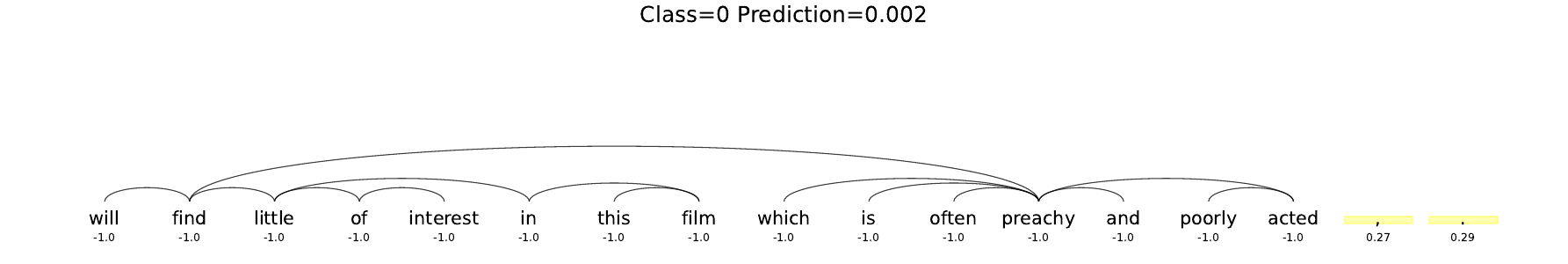}
        \caption{$\SUFFCAUSE=0$, $\FIDM=0$, $\RFIDM=0$}
    \end{subfigure}
    \begin{subfigure}{0.90\linewidth}
        \centering
        \includegraphics[width=\linewidth]{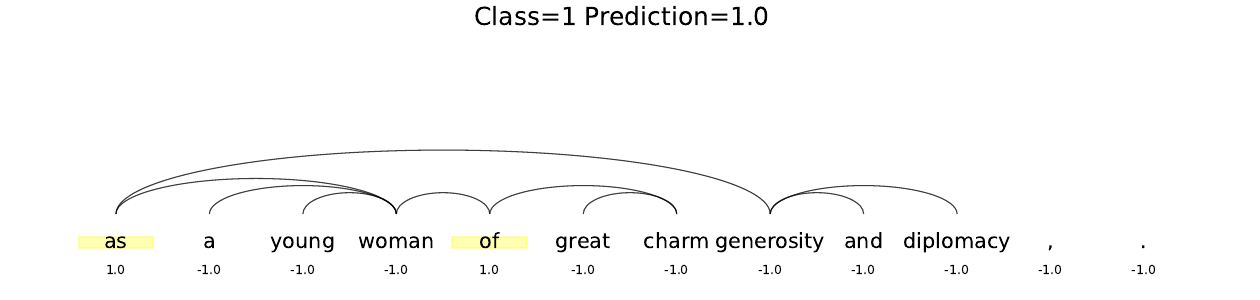}
        \caption{$\SUFFCAUSE=0$, $\FIDM=0$, $\RFIDM=0$}
    \end{subfigure}
    \begin{subfigure}{0.85\linewidth}
        \centering
        \includegraphics[width=\linewidth]{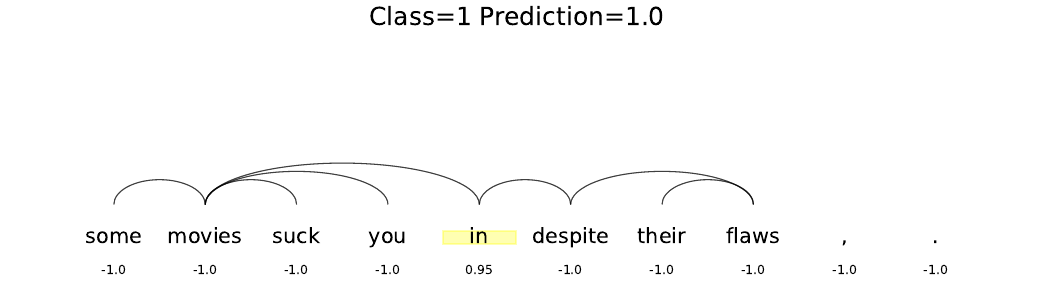}
        \caption{$\SUFFCAUSE=1$, $\FIDM=1$, $\RFIDM=1$}
    \end{subfigure}
    \caption{
        Examples of explanations for seed 4 of \DIR trained on \SSTP according to the setup described in \cref{sec:naturaldeg}.
        Nodes included in the explanation are chosen as those with the top $10\%$ relevance scores.
        Nodes left out of the explanation are given the default score of $-1$.
        Overall, for $y=0$ the explanations consistently highlight "," and ".", similarly to the malicious explanations of \cref{sec:segnns-can-be-manipulated}.
        For $y=1$, instead, explanations often highlight stop words.
        In addition, as shown by the rejection ratios reported below each example, \SUFFCAUSE, \FIDM, and \RFIDM can disagree on which explanation to reject. In particular, \SUFFCAUSE correctly rejects explanations highlighting "," and "." in two cases out of three, whereas the other metrics do not reject any of them.
        Recall that a rejection ratio of $1$ in the plots above means the explanation is rejected, hence considered unfaithful by the metric.
    }
    \label{fig:naturaldeg-SSTP-DIR}
\end{figure}

\begin{figure}[]
    \centering
    \begin{subfigure}{0.99\linewidth}
        \centering
        \includegraphics[width=\linewidth]{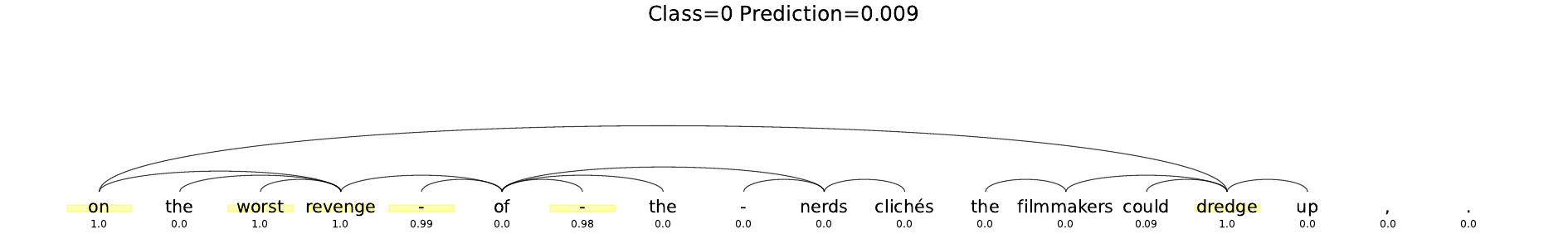}
        \caption{$\SUFFCAUSE=0$, $\FIDM=0$, $\RFIDM=0$}
    \end{subfigure}
    \begin{subfigure}{0.99\linewidth}
        \centering
        \includegraphics[width=\linewidth]{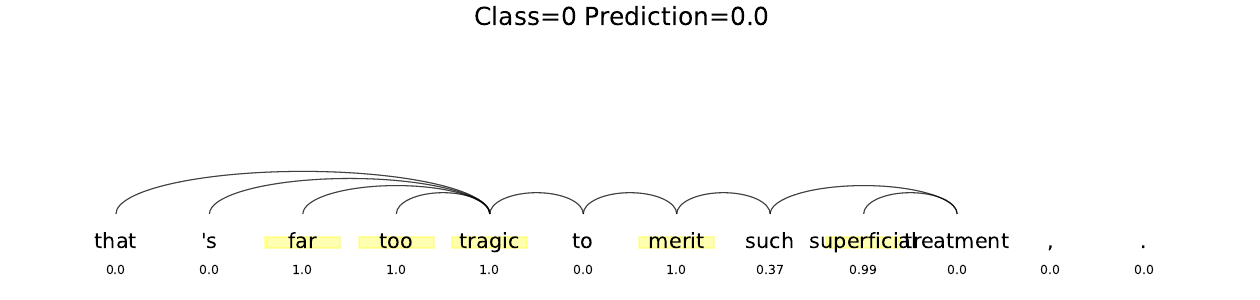}
        \caption{$\SUFFCAUSE=0$, $\FIDM=0$, $\RFIDM=0$}
    \end{subfigure}
    \begin{subfigure}{0.99\linewidth}
        \centering
        \includegraphics[width=\linewidth]{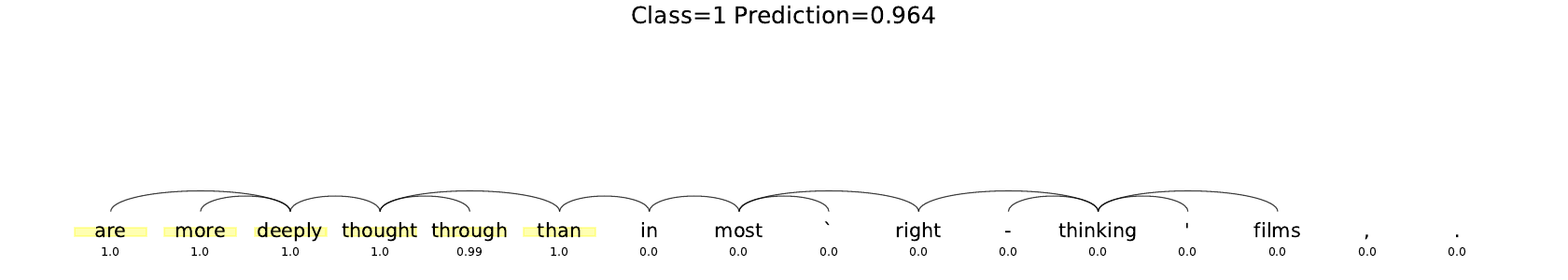}
        \caption{$\SUFFCAUSE=0$, $\FIDM=0$, $\RFIDM=0$}
    \end{subfigure}
    \begin{subfigure}{0.99\linewidth}
        \centering
        \includegraphics[width=\linewidth]{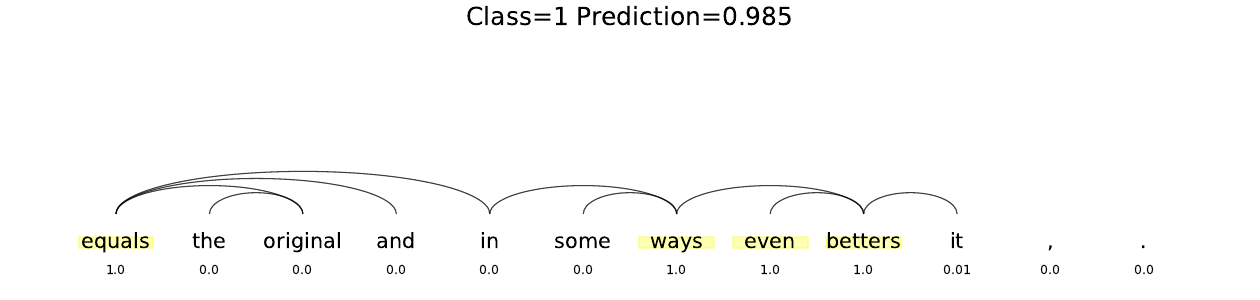}
        \caption{$\SUFFCAUSE=0$, $\FIDM=0$, $\RFIDM=0$}
    \end{subfigure}
    \begin{subfigure}{0.99\linewidth}
        \centering
        \includegraphics[width=\linewidth]{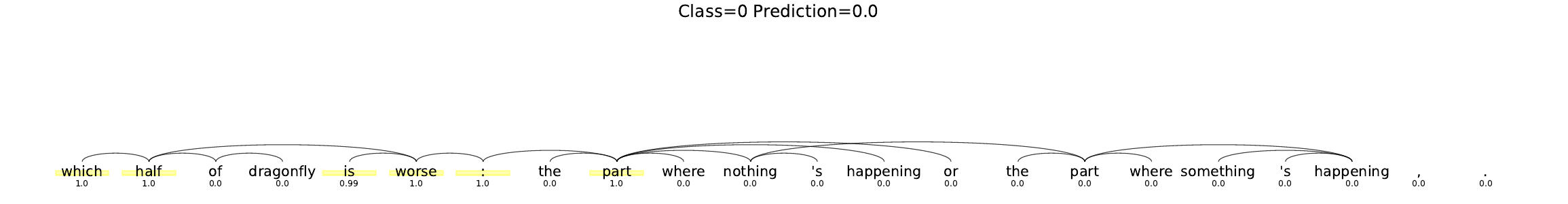}
        \caption{$\SUFFCAUSE=0$, $\FIDM=0$, $\RFIDM=0$}
    \end{subfigure}
    \begin{subfigure}{0.99\linewidth}
        \centering
        \includegraphics[width=\linewidth]{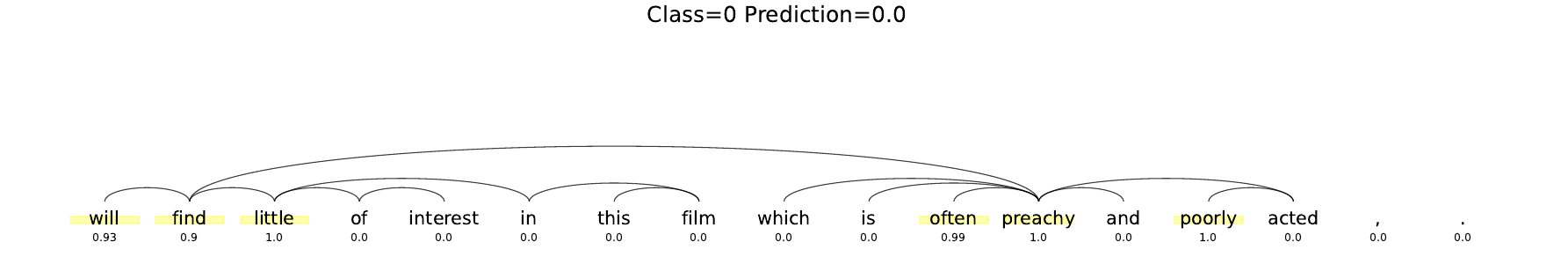}
        \caption{$\SUFFCAUSE=0$, $\FIDM=0$, $\RFIDM=0$}
    \end{subfigure}
    \caption{
        Examples of explanations for seed 1 of \SMGNN trained on \SSTP according to the setup described in \cref{sec:naturaldeg}.
        Nodes included in the explanation are chosen as those with a score greater than $0.5$.
        Overall, all explanations highlight words with high emotional content, hinting at the polarity of the sentence.
        Whilst this is not an indicator of faithful explanations, it is reasonable to at least expect that the model could have used this information to infer the final prediction.
        Then, the fact that all the tested metrics agree that those explanations are not to be rejected hints that they can be considered as faithful.
        Recall that a rejection ratio of $1$ in the plots above means the explanation is rejected, hence considered unfaithful by the metric.
    }
    \label{fig:naturaldeg-SSTP-SMGNN}
\end{figure}